\documentclass[a4paper,fleqn]{cas-sc}

\usepackage{amssymb}
\usepackage{amsmath}
\usepackage{longtable}
\usepackage{graphicx}
\usepackage{booktabs}
\usepackage{tabularx}
\usepackage{makecell}
\usepackage{ltablex}
\usepackage{float}
\usepackage{algorithm}
\usepackage{algpseudocode}
\usepackage{subfig}
\usepackage{placeins}
\usepackage[utf8]{inputenc}
\usepackage[authoryear,longnamesfirst]{natbib}
\usepackage{xcolor}

\def\tsc#1{\csdef{#1}{\textsc{\lowercase{#1}}\xspace}}
\tsc{WGM}
\tsc{QE}

\begin{document}
\let\WriteBookmarks\relax
\def\floatpagepagefraction{1}
\def\textpagefraction{.001}

\shorttitle{Cluster-Specific Localized Drift Detection for Efficient Batch Model Adaptation under Controlled Distribution Shift
}    

\shortauthors{}  

\title [mode = title]{Cluster-Specific Localized Drift Detection for Efficient Batch Model Adaptation under Controlled Distribution Shift}



%

\author[1]{Ignacio Cabrera Martin}[orcid=0009-0006-4729-6795]
\ead{N.CabreraMartin@brighton.ac.uk}
\credit{Conceptualization, Methodology, Data curation, Visualization, Writing - Original Draft}

\author[2]{Marcello Trovati}[orcid=0000-0001-6607-422X]
\ead{MTrovati@uclan.ac.uk}
\credit{Validation, Review \& Editing}

\author[1]{Almas Baimagambetov}[orcid=0000-0002-5656-6324]
\ead{A.Baimagambetov2@brighton.ac.uk}
\credit{Investigation, Review \& Editing}

\author[1]{Nikolaos Polatidis}[orcid=0000-0003-4249-4953]
\cormark[1]
\ead{N.Polatidis@brighton.ac.uk}
\credit{Supervision, Project administration, Writing - Review \& Editing, Validation, Funding acquisition}

\affiliation[1]{organization={University of Brighton},
            addressline={School of Architecture, Technology and Engineering}, 
            city={Brighton},
            postcode={BN2 4GJ}, 
            country={United Kingdom}}

\affiliation[2]{organization={University of Lancashire},
            addressline={School of Business}, 
            city={Preston},
            postcode={PR1 2HE}, 
            country={United Kingdom}}

\cortext[1]{Corresponding author}



\begin{abstract}
Machine learning systems deployed in dynamic environments frequently operate under non-stationary data distributions, where controlled distribution shift can progressively degrade predictive performance. However, many widely used tabular benchmark datasets lack explicit temporal structure, limiting reproducible evaluation of drift adaptation methods. This work proposes a cluster-induced distribution shift simulation framework that transforms static tabular datasets into controlled evolving data streams through structured perturbations across feature-space partitions.

Using this framework, six adaptation strategies are systematically evaluated: static learning, sliding-window retraining, global ADWIN retraining, cluster-local ADWIN retraining, random subspace drift detection, and feature-partitioned drift detection. Experiments are conducted on five benchmark datasets covering both classification and regression tasks using diverse predictive model families, including linear models, k-Nearest Neighbours, tree ensembles, boosting methods, and adaptive online learners.

The results show that sliding-window retraining often achieves the strongest predictive performance but incurs substantial computational and labelling cost due to frequent global updates. Global ADWIN retraining reduces update frequency while preserving competitive accuracy. In contrast, the proposed cluster-local adaptation strategy consistently achieves a stronger balance between predictive performance and computational efficiency across both classification and regression settings. Under structured distribution-shift scenarios, the method reduces retraining effort by up to 75\% while maintaining competitive predictive performance relative to continuously adaptive baselines. For nonlinear regression models, cluster-local adaptation preserves competitive $R^2$ performance while substantially lowering update frequency and training effort.

The proposed framework provides a reproducible benchmark for evaluating adaptation under controlled distribution shift on static tabular datasets and demonstrates that cluster-aware drift/shift detection constitutes an effective and computationally efficient alternative to uniform retraining strategies under heterogeneous distributional shift conditions.
\end{abstract}



\maketitle


\section{Introduction}
\label{secIntroduction}
Machine learning models deployed in real-world environments rarely operate under stationary data distributions. Instead, they are commonly exposed to evolving conditions in which the joint distribution of features and labels changes over time. Such phenomena, often referred to as \emph{concept drift}, can lead to substantial degradation in predictive performance if not properly addressed \citep{gama_survey_2014}. This challenge is particularly relevant for long-running systems, where frequent full retraining may be computationally expensive or infeasible.

A large portion of the existing literature on concept drift and adaptive learning assumes the availability of naturally ordered data streams with explicit timestamps. These assumptions facilitate the evaluation of online and incremental learning algorithms under realistic temporal dynamics. However, many widely used tabular benchmark datasets lack intrinsic temporal structure, despite being frequently used in drift-related studies \citep{souza_challenges_2020}. In practice, this has led to the use of heuristic instance orderings or synthetic shift injections that are often ad hoc, difficult to reproduce, and difficult to compare between studies.

Although established data-stream benchmarks such as SEA, STAGGER, Rotating Hyperplane, Electricity, Airlines, Weather, and RBF drift streams are widely used for evaluating online learning and concept-drift detection methods, they represent a different experimental setting from the one targeted in this study. These datasets either provide explicit temporal structure, are generated from predefined drift mechanisms, or are commonly used as canonical streaming benchmarks. In contrast, the objective of this work is to investigate how adaptive batch-classifier strategies can be evaluated when only static, unordered tabular datasets are available, as is often the case in standard machine learning benchmarks. The selected datasets therefore intentionally lack intrinsic temporal characteristics or annotated drift events. Rather than relying on naturally time-indexed streams, we construct controlled non-stationary streams from these static datasets through cluster-induced distributional shifts. This design enables the timing, magnitude, and locality of distributional change to be specified explicitly while preserving the statistical structure of real tabular data. Consequently, the focus of the paper is not to benchmark against conventional temporal stream generators, but to provide a reproducible methodology for studying localized drift detection and selective batch-model adaptation under controlled distribution shift in static tabular data.

This limitation motivates the need for \emph{controlled and reproducible} methodologies that can transform static tabular datasets into evolving data streams that exhibit well-defined distributional changes \citep{lu_learning_2019}. Such methods should enable systematic evaluation of adaptation mechanisms while preserving the statistical characteristics of the original data and allowing precise control over the timing, magnitude, and locality of drift \citep{aguiar2023, khannouz_benchmark_2020}.

A further motivation for localized adaptation is that distributional changes are often not uniform across the entire feature space. In many applications, only specific regions of the data distribution may be affected by change, while other regions remain stable. Under such conditions, global retraining may unnecessarily modify model behaviour in unaffected regions, even when their existing predictive patterns remain valid. Localized monitoring and selective adaptation aim to reduce this risk by identifying region-specific degradation and limiting updates to cases where there is evidence of change. This provides a more targeted adaptation mechanism that can improve computational efficiency while maintaining stable behaviour in parts of the feature space that are not affected by drift.

In this work, we propose a \emph{cluster-based distribution-shift framework} that converts static tabular datasets into evolving streams by inducing structured cluster-level distributional changes. Using the latent cluster structure, the proposed approach enables the simulation of both global and local drift patterns while remaining fully reproducible and independent of any specific classifier or adaptation strategy \citep{zliobaite_learning_2010}. The main contributions of this paper are as follows:
\begin{itemize}
    \item We propose a cluster-based distribution-shift method that transforms static tabular datasets into evolving data streams through controlled, cluster-induced distributional shifts.
    
    \item We define and evaluate six adaptation strategies under identical cluster-induced shift scenarios: (i) a train-once baseline, (ii) a sliding-window retraining strategy \citep{kifer_detecting_2004}, (iii) global ADWIN-triggered retraining \citep{bifet_learning_2007}, (iv) cluster-local ADWIN-triggered retraining, and to strengthen the empirical evaluation, we introduced additional localized drift detection baselines beyond ADWIN. Specifically, we implemented (v) a random subspace-based detector as S5 and (vi) a feature-partitioned detector as S6.
    
    \item We compare the performance of diverse predictive model families, including batch learners such as Random Forests \citep{breiman2001random}, Logistic and Linear Regression \citep{hosmer2000applied}, k-Nearest Neighbours \citep{cover1967nearest}, and XGBoost \citep{chen2016xgboost}, as well as incremental and adaptive stream-learning models such as Hoeffding Adaptive Trees (HAT), Adaptive Random Forest (ARF), Aggregated Mondrian Forest (AMF), and online variants of linear models, Naïve Bayes, and kNN. This comparison highlights differences in robustness and adaptability under controlled distribution shift across both batch and online learning paradigms.
    
    \item We conduct extensive experiments on five diverse standard benchmark datasets spanning both classification and regression tasks, with varying scales, dimensionalities, and feature distributions. Multiple cluster-induced distribution-shift scenarios are evaluated to analyse predictive performance over time, retraining efficiency, and adaptation responsiveness.
\end{itemize}

Although the individual components used in this study, including K-means clustering, ADWIN-based drift detection, and retraining-based adaptation, are well-established, their integration serves a specific purpose that is not addressed by each component in isolation. In the proposed framework, clustering defines localized regions of the feature space, ADWIN monitors predictive degradation independently within those regions, and adaptation is triggered selectively when localized evidence of drift is detected. This combination enables cluster-induced distribution shifts to be simulated, detected, and handled at the level of feature-space regions rather than only at the global stream level. As a result, the framework can identify heterogeneous shift patterns that may be diluted in global error statistics, while reducing unnecessary retraining compared with continuous sliding-window adaptation. The novelty of the proposed method therefore lies in the unified and reproducible integration of cluster-based stream construction, cluster-local drift monitoring, and selective batch-model adaptation for static tabular datasets.

Together, these contributions provide a principled and reproducible framework for studying adaptive learning on tabular data streams, and offer new insights into the trade-\-offs between global and localized adaptation strategies under structured distribution shift.

This paper is organized as follows. Section \ref{secBackground} presents the background and related work. Section \ref{secData} describes the data and the cluster-based distribution-shift approach. Section \ref{secAdaptations} provides an overview of the adaptation strategies used in the experiments, while Section \ref{secClassifiers} offers a detailed description of the selected predictive models. Section \ref{secEvaluation} outlines the evaluation protocol, and Section \ref{secResults} presents and discusses the results. Section \ref{secDiscussion} provides a broader discussion of the findings, and Section \ref{secConclusion} concludes the paper by summarizing the main findings and contributions of this study.

\section{Background and Related Work}\label{secBackground}
The deployment of machine learning models in real-world environments often violates the stationary assumption, the premise that the training data distribution remains constant over time. When the statistical properties of the target variable change, the phenomenon is known as concept drift. This section reviews the foundational literature regarding the characterization, detection, and adaptation to drift.

\subsection{Concept Drift and Adaptation}
Concept drift is not a singular phenomenon but follows various patterns and magnitudes. Ditzler et al. \citep{ditzler_learning_2015} highlight the fundamental challenges supervised learning algorithms face when data streams evolve, necessitating mechanisms that can unlearn obsolete information while acquiring new patterns. To rigorously analyze these changes, Webb et al. \citep{webb_characterizing_2016}  move beyond binary definitions of change, proposing a framework to measure drift in terms of magnitude (severity), duration (transition time), and velocity. This quantitative approach allows for a granular distinction between sudden, gradual, and incremental drift, which is essential for selecting appropriate model maintenance strategies.

As the field has matured, efforts to systematize the terminology have grown. Lu et al. \citep{lu_learning_2019} provide a comprehensive modern review that contrasts families of approaches across detection, adaptation, and evaluation. Building on this, Bayram et al. \citep{bayram_concept_2022} offer a useful taxonomy that connects the specific technical issue of concept drift to the broader operational problem of "model degradation", and unify terminology to help practitioners understand how drift manifests as performance decay in deployed systems, bridging the gap between theoretical drift research and practical MLOps concerns.

A critical component of handling drift is the ability to detect it accurately and in a timely manner. Agrahari et al. \citep{agrahari_concept_2022} categorises methods based on their statistical foundations and highlights the inherent trade-offs in detection design, specifically the tension between minimizing detection delay and avoiding false alarms. Once drift is identified, the system must react. Gama et al. \citep{gama_survey_2014} provide the canonical survey on adaptation, introducing the fundamental distinction between active adaptation, which relies on a trigger mechanism (a detector) to alert the model to change, prompting a retrain or update, and passive adaptation, which continuously updates the model (e.g., using ensembles or online learning) regardless of whether a specific drift event was detected, and they also establish the standard evaluation protocols for these evolving streams, emphasizing prequential evaluation over traditional hold-out validation to better reflect temporal dependencies.

\subsection{Error-Based Drift Detectors}
These types of detectors monitor the performance of a predictive model in real-time. By tracking metrics such as misclassification rates or the distances between errors, these methods trigger an adaptation mechanism when a statistically significant change in performance is identified.

The Adaptive Windowing (ADWIN) \citep{bifet_learning_2007} is a statistically-principled approach that maintains a sliding window of variable size. Unlike fixed windows, ADWIN automatically grows when data is stationary and shrinks when a change is detected. It provides rigorous guarantees on the false positive rate and is widely considered the gold standard for error-based triggers due to its ability to handle both abrupt and gradual drifts without manual parameter tuning.

However, two of the most foundational benchmarks in the field focus on the rate and distribution of model failures, where the Drift Detection Method (DDM) \citep{gama_learning_2004} monitors the online error rate of a stream. It defines two thresholds: a warning level, where the model begins buffering data for a potential update, and a drift level, where the model is replaced. Although simple and efficient, it can struggle with very slow, gradual drifts. Then, the Early Drift Detection Method (EDDM) \citep{baena-garcia_eddm_2006} improves upon DDM by tracking the distance between consecutive errors rather than the raw error rate. This makes it significantly more sensitive to gradual changes where errors become more frequent over time, even if the total error count hasn't spiked yet.

Later iterations of drift detection moved toward more refined statistical frameworks to balance sensitivity with computational efficiency, where EWMA Charts \citep{ross_ewma_2012} is a method that applies Exponentially Weighted Moving Average control charts to the misclassification rate. By weighting recent errors more heavily, it achieves low computational overhead while allowing practitioners to strictly control the false alarm rate through chart design parameters. Then, McDiarmid Drift Detection (MDDM) \citep{pesaranghader_mcdiarmid_ijcnn_2018} family of detectors utilizes the McDiarmid inequality to provide strong statistical guarantees. MDDM is often used as a robust alternative to DDM and EDDM, where it bridges the gap between the simplicity of DDM and the rigor of ADWIN, offering faster detection with a fixed-size memory footprint. The core strength of MDDM is its use of a weighted sliding window, where more recent samples carry higher significance. This allows the detector to respond to sudden shifts before the window's total average has time to fully skew, and handle both abrupt and gradual drifts with higher stability across diverse data streams.

\subsection{Sliding-Window Learning}
While error-based detectors signal when a change occurs, sliding-window and chunk-based strategies focus on how the model should adapt. These methods manage data in discrete blocks or moving windows, ensuring the learner prioritizes recent, relevant information over stale, historical patterns.

The Early Ensemble and Chunk-Based Approaches in the early 2000s marked a shift from static models to dynamic ensembles that could "forget" outdated concepts, where the Streaming Ensemble Algorithm (SEA) \citep{street_sea_2001} is a foundational "chunk-in-stream" method. It trains individual classifiers on fixed-size data blocks and maintains a fixed-size ensemble. When a new chunk arrives, a new model is trained; if the ensemble is full, the "weakest" member (the one performing poorest on the latest data) is replaced. Then, the Accuracy Weighted Ensemble (AWE) \citep{wang_ensemble_2003} refined this by introducing a weighting mechanism. Instead of simple replacement, AWE weights each classifier based on its performance on the most recent data block. This allows the ensemble to handle "recurring" concepts where old patterns re-emerge by simply re-weighting existing models rather than discarding them entirely.

A more rigorous approach to windowing involves comparing different temporal segments of the stream. Kifer et al. \citep{kifer_detecting_2004} established the foundational "two-window" approach, where a "Reference Window", representing the "old" concept, is compared with a "Current Window", representing the most recent data. By applying statistical tests to compare the distributions of these two windows, researchers can detect shifts without relying on model labels, providing a more objective measure of the change in distribution.

The work by Bifet et al. \citep{bifet_adaptive_2009} is often seen as the bridge between pure drift detection and practical model architecture. This paper introduced several critical concepts: (i) the Two-Module Framework, formalising the split between a Change Detector (to signal shifts) and an Estimator (to maintain updated statistics). This decoupling allows developers to swap different detection algorithms into different learning models. (ii) Hoeffding Adaptive Trees (HAT) the evolution of the standard Hoeffding Tree, HAT embeds an ADWIN detector at every node. If a branch begins to underperform, ADWIN signals a drift, and the branch is replaced with a new, better-aligned subtree. (iii) Parameter-Free Adaptation, where HAT uses ADWIN to automatically find the optimal window size, making the model truly autonomous in evolving environments, unlike its predecessors which required users to guess the "speed" of changes. Building on the success of individual adaptive trees, Gomes et al. \citep{gomes_arf_2017} introduced Adaptive Random Forests. ARF scales the adaptive windowing concept to an ensemble level, where each tree in the forest is paired with its own drift and warning detectors. This granular approach allows the forest to maintain high accuracy by replacing only the specific trees that are struggling with a new concept, rather than resetting the entire ensemble.

\subsection{Cluster-Based Partitioning and Synthetic Drift}
While performance-based monitors are reactive, cluster-based and partitioning strategies allow for a more granular, structural understanding of how data distributions shift. This subsection explores the frameworks used to both simulate these shifts synthetically and detect them through spatial decomposition.

Before testing on real-world data, researchers rely on synthetic generators to create controlled, reproducible environments. These benchmarks are essential for validating how a model handles specific types of drift (abrupt vs. gradual). STAGGER Concepts \citep{schlimmer_incremental_1986} is one of the earliest benchmarks for concept drift. It uses simple boolean concepts (e.g., color and shape) that change over time, forcing a learner to adapt to new logical rules. The Massive Online Analysis (MOA) framework \citep{bifet_moa_2010} is the "Swiss Army Knife" of data stream mining. It codifies the standard constraints of single-pass processing and bounded memory while providing several key generators: (i) ConceptDriftStream, that injects drift by blending two distributions using a sigmoid function, allowing for a smooth, parameterised transition from an old concept to a new one. (ii) RandomRBFGeneratorDrift that introduces drift by physically moving the centroids of Radial Basis Functions (RBF) in the feature space. This is a primary tool for simulating drift in non-temporal datasets. As reference, the MOA Tutorials that serve as the formal documentation for configuring drift width (how gradual the change is) and timing, ensuring experimental reproducibility.

Cluster-based detection methods partition the space into micro-clusters or regions, rather than viewing data as a single monolithic stream. These methods partition the space into "micro-clusters" or regions to track localized changes. The CluStream \citep{aggarwal_clustream_2003} foundational framework introduced the concept of micro-clusters, which maintain statistical summaries of data points in a localized region. By analysing how these micro-clusters grow, fade, or move over time, CluStream provides a temporal view of evolution. Similar methods to identify "novelty", OLINDDA \citep{spinosa_olindda_2007} and SAND \citep{haque_sand_2016}, where new data arrives that doesn't fit in existing clusters, these algorithms determine if it is mere noise or the start of a new "emerging" concept. This is particularly useful for detecting drift in the absence of labels. 

Modern approaches focus on partitioning the feature space to detect "localized" drift that might be masked in a global view. This is the case for EI-kMeans \citep{liu_concept_2021} where this method uses Equal-Intensity k-means to partition the space into histogram-like bins. By monitoring the density of data within these bins, it can detect subtle distribution shifts that don't immediately impact overall model accuracy but indicate a fundamental change in the data's structure. Similarly, the Unsupervised High-Dimensional Detection \citep{souza_challenges_2020} assumes that labels are often delayed or unavailable, and focuses on detecting drift in high-dimensional spaces by monitoring cluster movements and partitioning transitions, providing a robust "early warning" system for complex data.

\subsection{Batch Learning vs Incremental Learning}
The architectural strategy for data ingestion is a critical determinant of a model's performance in non-stationary environments. Research in stream mining distinguishes between traditional batch methods and evolving incremental paradigms, primarily focusing on the trade-offs between computational efficiency, latency, and the ability to adapt to concept drift.

The cornerstone of modern incremental learning is the Very Fast Decision Tree (VFDT), or Hoeffding Tree, introduced by Domingos and Hulten \citep{domingos_mining_2000}. Unlike traditional batch learners that require multiple passes over a static dataset to identify optimal splits, the VFDT builds decision trees online with constant-time and space complexities, where the core innovation is the use of the Hoeffding Bound to decide when a leaf has seen enough data to justify a split. The Hoeffding bound is calculated as shown in Equation \ref{eq:hoeffding}:

\begin{equation} \label{eq:hoeffding}
\epsilon = \sqrt{\frac{R^2 \ln(1/\delta)}{2n}}
\end{equation}

By maintaining only sufficient statistics at the leaves rather than storing raw instances, the VFDT achieves extreme memory efficiency. While the original VFDT was designed for stationary data, it serves as the fundamental benchmark for adaptive variants like the Hoeffding Adaptive Tree (HAT), which integrate drift detectors to handle non-stationary streams.
 
To achieve the predictive power of ensembles in a streaming context, Oza and Russell \citep{oza_online_2005} successfully adapted Bagging and Boosting into instance-incremental versions. Their approach utilizes the Poisson Approximation; as the number of instances $N$ approaches infinity, the Binomial distribution used in batch resampling converges to a $Poisson(1)$ distribution.
In Online Bagging, each arriving instance is used to update base models $K$ times, where $K$ is drawn from $Poisson(1)$. Online Boosting extends this by passing instances through the ensemble sequentially, adjusting the Poisson parameter $\lambda$ based on whether the previous model classified the instance correctly. This provides a mathematical bridge allowing "batch-level" performance within an incremental, one-pass framework.

The choice between processing data instance-by-instance or in blocks is explored deeply by Read et al. \citep{read_batch-incremental_2012}. They define two primary paradigms: (i) Instance-Incremental (II): True online learning where the model is updated immediately upon the arrival of each example. (ii) Batch-Incremental (BI): Data is buffered into a batch of size $n$ before a collective update is performed.

Their comparison reveals a significant latency-accuracy trade-off. II models react faster to sudden drift because they incorporate new information immediately. Conversely, BI models can be more computationally efficient due to vectorization and lower update overhead, but they suffer from a "performance lag" as they remain stagnant until a full batch is processed. Furthermore, BI can act as a natural smoothing filter against noise in gradual drift scenarios, whereas II may be hyper-sensitive.

As data scales toward production levels, frameworks like SAMOA \citep{morales_samoa_2015} bridge the gap between algorithmic theory and distributed systems. SAMOA scales incremental algorithms (e.g., the Vertical Hoeffding Tree) across clusters using a pluggable "Topology" architecture, allowing learners to run on distributed engines like Apache Storm or Samza.

Finally, the shift toward continuous learning introduces the Stability-Plasticity Dilemma. As categorized by Losing et al. \citep{losing_incremental_2018}, incremental learners must balance the acquisition of new information (plasticity) with the preservation of old knowledge to prevent catastrophic forgetting. Their taxonomy of instance-based, kernel-based, and tree-based methods highlights that while incremental learning offers theoretical advantages in speed, real-world deployment remains sensitive to hyperparameter configurations and the specific nature of the drift encountered.

To conclude, this work intentionally deviates from the use of oracle-provided drift labels or fully automated AutoML systems that might obscure the underlying dynamics of adaptation. Instead, we focus on evaluating fixed learners within synthetic, yet highly controlled, drift environments, a methodology advocated in foundational works like Gama et al. \citep{gama_survey_2014} and Bifet et al. \citep{bifet_moa_2010} to ensure rigorous benchmarking. By inducing structured shifts through cluster-based partitioning, we create a transparent testing ground where the timing and magnitude of distributional changes are precisely known. This controlled approach allows for a granular comparison between global and localized adaptation strategies, isolating the efficacy of the ADWIN detector and various retraining policies without the confounding variables introduced by black-box optimization frameworks.

\section{Data and Cluster-Based Distribution Shift}\label{secData}
While synthetic generators provide high control and real-world streams provide high complexity, the former often lacks the intricate feature dependencies of real data, and the latter typically lacks "ground truth" labels for the exact timing and magnitude of distributional shifts. To address these limitations, we employ a hybrid stream construction methodology that transforms static real-world datasets into non-stationary streams through cluster-based re-ordering.

\subsection{Datasets}\label{subDatasets}
\begingroup
To ensure a robust and comprehensive experimental validation, five diverse real-world unordered datasets were selected, covering both classification and regression tasks across a wide spectrum of statistical profiles, dimensionality, and application domains. These datasets were specifically chosen for their lack of inherent temporal ordering or explicit drift annotations, making them well suited to the proposed cluster-based re-ordering methodology for simulating controlled non-stationary environments.

As shown in Table \ref{tab:datasets}, the selected benchmark suite comprises datasets drawn from Social Science, Food Science, Health and Medicine, Engineering, and Materials Science, ensuring that the adaptation strategies are evaluated under varying levels of feature dependency, class separability, noise, and nonlinear structure. The datasets vary considerably in dimensionality, ranging from lower-dimensional settings such as Adult and Wine Quality to higher-dimensional problems such as Breast Cancer Wisconsin and Superconductivty Data, thereby enabling assessment across both moderate and complex feature spaces. Data volume also differs substantially, from relatively small-scale datasets such as Breast Cancer (569 instances) to substantially larger streams such as Superconductivty data and Airfoil Self-Noise, allowing scalability analysis under different stream lengths.

This selection further supports multiple supervised learning paradigms, including both classification and regression tasks. The classification benchmarks (Adult, Wine Quality, and Breast Cancer) provide diverse categorical prediction problems with varying imbalance and decision boundary complexity, while the regression benchmarks (Airfoil Self-Noise and Superconductivty data) introduce continuous-target tasks characterised by nonlinear relationships and higher sensitivity to distribution shift. Collectively, this heterogeneous dataset suite enables rigorous evaluation of the proposed adaptation strategies across a broad range of realistic streaming scenarios.

\begin{table}[t]
\centering
\caption{Summary of real-world datasets used for experimental validation.}
\label{tab:datasets}
\small
\setlength{\tabcolsep}{4pt}

\begin{tabularx}{\linewidth}{@{}X l r r X l@{}}
\toprule
\textbf{Dataset Name} & \textbf{Task} & \textbf{Feat.} & \textbf{Inst.} & \textbf{Subject Area} & \textbf{Ref.} \\
\midrule
Adult                  & Classification & 14 & 48,842 & Social Science              & \citep{barry_becker_adult_1996} \\
Wine Quality           & Classification & 11 & 4,898  & Food Science                & \citep{cortez_modeling_2009} \\
Breast Cancer Wis.     & Classification & 30 & 569    & Health and Medicine         & \citep{street_nuclear_1993} \\
Airfoil Self-Noise     & Regression     & 5  & 1,503  & Engineering / Acoustics     & \citep{thomas_brooks_airfoil_1989} \\
Superconductivty Data & Regression     & 81 & 21,263 & Materials Science / Physics & \citep{kam_hamidieh_data-driven_2018} \\
\bottomrule
\end{tabularx}
\end{table}

\par
\endgroup
With the objective of ensuring the integrity of the cluster-based construction and the subsequent performance of the shift/drift detectors, a rigorous preprocessing pipeline is applied to the raw data. This stage is critical because the K-means algorithm, which we use to define our shift boundaries, and the underlying machine learning models are highly sensitive to the scale and format of the input features \citep{hastie2009elements, bishop2006pattern}.

The transformation of static datasets into non-stationary streams begins with standardizing the representation of missingness. We normalize various missing data tokens specifically "?", "", "None", and "nan" into a unified NaN (Not a Number) representation. Then, with the objective of maintaining dataset volume without introducing significant bias, we employ a two-tiered imputation strategy:
\begin{itemize}
    \item Numeric Features: Missing values are imputed using the median of the column to ensure robustness against outliers \citep{hastie2009elements, geron2019hands}.
    \item Categorical Features: Missing entries are filled using the mode (most frequent value) \citep{james2013introduction, little2002statistical}.
    \item Boolean Conversion: All boolean columns are explicitly cast to integers to ensure compatibility with numerical solvers \citep{geron2019hands, james2013introduction}.
\end{itemize}

In addition, to accommodate models such as XGBoost and various scikit-learn estimators, the feature space must be transformed into a fully numeric matrix, where one-hot encoding is applied to all categorical variables, this avoids the imposition of an arbitrary ordinal relationship on nominal data \citep{Hancock2020CategoricalNN}. In addition, as a safety measure against any residual gaps following the encoding process, any remaining missing values are filled with 0.

The final stage of the pipeline involves the application of a StandardScaler to the entire feature matrix. While tree-based models like XGBoost are generally invariant to feature scale, standardization is essential for distance-based models (such as kNN and K-means) and linear models (such as Logistic Regression) to prevent features with larger magnitudes from dominating the objective function. For consistency across all experiments, the scaler is applied to all models. Each feature $x$ is transformed into a $z$-score such that it has a mean of 0 and a standard deviation of 1, using the formula seen in Equation \ref{eq:standardization}, where $\mu$ is the mean of the feature and $\sigma$ is the standard deviation. This comprehensive pipeline results in a high-fidelity, fully numeric feature matrix that is ready for both the cluster-partitioning phase and the subsequent evaluation of adaptive learning models.

\begin{equation}
z = \frac{x - \mu}{\sigma}
\label{eq:standardization}
\end{equation}

\subsection{Initial Clustering and Stream Construction for Drift Detection}

This phase constitutes the core of the synthetic distribution-shift generation process. Its purpose is to transform a static, preprocessed dataset into a controlled non-stationary stream by partitioning the feature space into localized regions and then reordering the emission of samples across those regions. In this way, the framework induces structured distributional shifts while preserving the statistical characteristics of the original tabular data. Let the preprocessed dataset be denoted by Equation \ref{eq:dataset}:

\begin{equation} \label{eq:dataset}
\mathcal{D}=\{x_i\}_{i=1}^{N}, \qquad x_i \in \mathbb{R}^{d}
\end{equation}

where \(N\) is the number of samples and \(d\) is the feature dimension. To identify localized data regions, we apply MiniBatchKMeans to partition the dataset into \(k\) disjoint clusters. The clustering model minimizes the within-cluster sum of squares, as seen in Equation \ref{eq:sumSq}:

\begin{equation} \label{eq:sumSq}
\mathcal{J}=\sum_{i=1}^{N}\left\|x_i-\mu_{c_i}\right\|_2^2,
\end{equation}

where \(\mu_j\) denotes the centroid of cluster\(j\), and \(c_i\) is the cluster label assigned to sample \(x_i\). Each sample is associated with its nearest centroid according to Equation \ref{eq:centroid}:
\begin{equation} \label{eq:centroid}
c_i=\arg\min_{j\in\{1,\dots,k\}}\|x_i-\mu_j\|_2^2.
\end{equation}

This step partitions the original dataset into \(k\) localized clusters, each representing a distinct region of the feature space with its own statistical characteristics. After cluster assignment, the samples within each cluster are shuffled independently to eliminate any unintended temporal ordering effects in the cluster-specific pools.

The non-stationary stream is then constructed as a sequence of batches emitted from these cluster-specific pools, as displayed on Equation \ref{eq:sequences}:
\begin{equation} \label{eq:sequences}
\mathcal{S}=\{B_t\}_{t=1}^{T}, \qquad B_t \subseteq \mathcal{D}_{r_t},
\end{equation}
where \(B_t\) denotes the batch emitted at time step \(t\), \(\mathcal{D}_{r_t}\) is the shuffled pool associated with the active cluster \(r_t\), and \(r_t \in \{1,\dots,k\}\) identifies the cluster selected at that step. The sequence is governed by the stream-generation rule:
\begin{equation}
r_t =
\begin{cases}
((t-1)\bmod k)+1, & \text{cyclic mode},\\[4pt]
\mathrm{Uniform}\big(\{1,\dots,k\}\big), & \text{random mode}.
\end{cases}
\end{equation}

Accordingly, in cyclic mode the stream passes through the clusters in a fixed repeating order, whereas in random mode the next active cluster is selected stochastically. In the present work, we primarily employ the cyclic configuration because it provides explicit control over the ordering and rotation of the cluster regimes during stream construction. This controlled rotation allows the framework to systematically manage how shift regions appear and reappear over time, ensuring that transitions between clusters occur in a deterministic and reproducible manner. As a result, the cyclic setting enables more precise evaluation of localized adaptation behaviour, since the detector repeatedly encounters known regime transitions under controlled temporal conditions. Because each batch is drawn from the currently active cluster pool, transitions between clusters induce abrupt changes in the input distribution. Thus, a shift event occurs whenever the active cluster changes between two consecutive batches, that is, when \(r_t \neq r_{t-1}\). During stream assembly, the associated boundary indices are recorded directly, yielding exact ground-truth shift locations for evaluating detection accuracy and delay.

The cyclic configuration is particularly important for analysing the proposed S4 strategy because it allows the experiments to isolate the effect of cluster-localized adaptation without introducing additional stochastic variability from the stream generator itself. In other words, the deterministic cluster rotation ensures that observed performance differences are primarily attributable to the adaptation strategy rather than random emission behaviour. This provides a stable and reproducible benchmark for evaluating how effectively localized ADWIN detectors respond to recurring regime changes.

By contrast, the random mode introduces stochastic transitions between regimes and therefore produces less predictable temporal dynamics. While this setting more closely resembles unconstrained real-world streams, it also introduces additional variability in transition frequency and cluster recurrence patterns. Consequently, the cyclic mode was selected as the primary evaluation configuration in order to maintain precise control over cluster management and transition scheduling while benchmarking the framework.

Under this construction, the generated stream preserves the structural complexity of the original dataset while providing explicit control over when and how transitions occur. This makes the framework suitable for reproducible benchmarking of adaptive batch-classification strategies under controlled distribution shift.

\begingroup

\subsection{Key Parameters}
To improve reproducibility, we made the experimental protocol explicit in the revised implementation. All streams were processed using prequential batch evaluation with $\texttt{BATCH\_SIZE} = 1000$; for smaller datasets the effective batch size was set to 
$\min\left(1000,\ \max\left(50,\ \lfloor \tfrac{\text{len(stream)}}{10} \rfloor\right)\right)$. The sliding-window baseline used $\texttt{WINDOW\_SIZE} = 1500$.

Synthetic distribution shift was generated by first clustering each dataset into $\texttt{N\_GEN\_CLUSTERS} = 5$ regimes using \texttt{MiniBatchKMeans}. The stream was then constructed from cluster-ordered segments of length $\texttt{DRIFT\_SEGMENT\_FRAC} = 0.10$ of the dataset size. The stream generation mode selected is cyclic, i.e. \texttt{DRIFT\_MODE} = \texttt{"cycle"}. This produces repeated regime changes at the stored $\texttt{Drift\_Points\_RowIdx}$ boundaries.

For the proposed cluster-specific detector, stream generation and drift detection are intentionally decoupled. The detector uses $\texttt{N\_DET\_CLUSTERS}$ clusters fitted only on the initial observed batch. The main matched configuration uses $k_{\text{gen}} = 5$ and $k_{\text{det}} = 5$. Robustness is evaluated under four settings: (i) matched clustering, (ii) centroid mismatch, (iii) detector cluster-count mismatch, and (iv) combined centroid and cluster-count mismatch. For sensitivity analysis, $k_{\text{det}}$ is swept over $\{3, 5, 10\}$ while keeping $k_{\text{gen}} = 5$.

All ADWIN-based methods use River’s default \texttt{ADWIN()} configuration. This applies to the global detector, the cluster-specific detectors, the random-subspace detectors, and the feature-partition detectors. In S4, one independent ADWIN instance is maintained per detector cluster. In S5, five persistent random-subspace ADWIN detectors are used. In S6, three persistent feature-partition ADWIN detectors are used.

Each simulation is run across all combinations of dataset, model, strategy, and robustness setting, subject to runtime safety limits: $\texttt{MAX\_ROWS\_PER\_DATASET} = 15000$, $\texttt{MAX\_SECONDS\_PER\_SIM} = 900$, and $\texttt{MAX\_BATCHES\_PER\_SIM} = \texttt{None}$. Very slow combinations (e.g., Adult and Superconductivity datasets with AMF and local/regional ADWIN baselines) are skipped as specified by \texttt{SLOW\_COMBINATIONS}.

Results are reported as averages across batches within each run, rather than averages over repeated independent runs. The implementation logs batch-level metric histories, adaptation counts, update times, drift locations, and S4 cluster-level ADWIN dynamics. These logs enable post hoc estimation of variance across batches and robustness settings.
\par
\endgroup
\section{Adaptation Strategies}\label{secAdaptations}
In dynamic environments, the effectiveness of a predictive model often degrades as the underlying data distribution shifts, a phenomenon commonly studied under concept drift and distribution shift. To maintain performance over time, we evaluate six distinct adaptation strategies. These strategies vary in their timing, methodology, and the granularity at which a batch classifier is updated.

All evaluated strategies adhere to a consistent prequential (test-then-train) protocol and utilize a fixed offline clustering mechanism \citep{Vinagre2014, suarez2023recurring}. This standardized framework ensures that comparisons between strategies are based solely on their adaptive logic rather than variations in the evaluation pipeline.

\subsection{S1. Train-Once (Static Baseline)}\label{subS1}
The Train-Once (S1) strategy serves as the control group for our experimental evaluation, representing a strictly static modeling approach \citep{gama2010knowledge}. In this configuration, the classifier $M$ is trained exclusively on the initial data batch $B_0$, establishing a fixed hypothesis space that remains unchanged throughout the duration of the data stream $\mathcal{S}$.

The procedural realization of S1 is detailed in Algorithm \ref{alg:train_once}. Following the initialization phase, the model performs inference on every subsequent batch $B_t$ ($t=1, \dots, T$) without any weight adjustments, parameter re\-calibration, or structural updates. To formally track this lack of computational intervention, we define the training effort parameter $N_{train} = 0$ for all post-initialization steps. This value serves as the computational floor for our efficiency analysis.

The primary objective of S1 is to establish a lower-bound performance baseline. By maintaining a constant model state, we isolate the impact of non-stationarity on predictive performance. The resulting decay in metrics, specifically Accuracy and F1-Score, allows us to quantify the "cost of inaction." This baseline provides a necessary benchmark to measure the relative gain offered by more complex, adaptive strategies and confirms whether the observed non-stationarity in the stream is significant enough to necessitate active model maintenance.

\begin{algorithm}
\caption{Train-Once (S1) Static Baseline Strategy}
\label{alg:train_once}
\begin{algorithmic}[1]
\Require Initial batch $B_0$, Data stream $\mathcal{S} = \{B_1, B_2, \dots, B_T\}$, Classifier $M$
\Ensure Performance metrics $\{Acc_t, F1_t\}$ for all $t \in [1, T]$
\Statex
\State \textbf{Initialization:}
\State $M \leftarrow \text{Train}(M, B_0)$ \Comment{Train only once on the first batch}
\State $N_{train} \leftarrow 0$ \Comment{Initialize computational intervention counter}
\Statex
\For{each subsequent batch $B_t \in \mathcal{S}$}
    \State $\hat{y}_t \leftarrow \text{Predict}(M, B_t)$ \Comment{Perform inference without updates}
    \State $\{Acc_t, F1_t\} \leftarrow \text{Evaluate}(\hat{y}_t, y_t)$ \Comment{Calculate metrics for batch $t$}
    \State $N_{train} \leftarrow 0$ \Comment{Formal record of zero training effort}
\EndFor
\State \Return $\{Acc_t, F1_t\}_{t=1}^T$
\end{algorithmic}
\end{algorithm}

\subsection{S2. Sliding-Window Retrain (No Detector)}\label{subS2}
The Sliding-Window Retrain (S2) strategy, formalized in Algorithm \ref{alg:sliding_window}, represents a proactive, continuous adaptation approach that operates independently of explicit drift detection triggers. Grounded in the assumption that the most recent data observations are the most representative of the current underlying distribution, this strategy frequently refreshes the model to track non-stationary environments \citep{zliobaite_learning_2010, bifet_learning_2007, losing_incremental_2018}.

The mechanism maintains a rolling buffer $W$, which stores the most recent features and labels. For every incoming batch $B_t$ in the data stream $\mathcal{S}$, the strategy executes a rigid three-step update cycle:
\begin{itemize}
\item \textbf{Append:} New observations from $B_t$ are integrated into the existing buffer $W$.
\item \textbf{Trim:} The buffer is truncated to ensure the total sample count does not exceed a predefined maximum capacity $W_{max}$, preserving a bounded memory footprint.
\item \textbf{Refit:} The classifier $M$ undergoes a complete retraining process on the updated window $W$.
\end{itemize}

This strategy serves as a "blind" but robust adaptation baseline. By performing a full refit on every batch, it ensures that the model's hypothesis space never lags more than one batch behind the current distribution. The computational overhead is formally quantified by $N_{train, t} = |W|$, representing the total number of instances used for model recalibration at each time step. While S2 provides a high-responsiveness benchmark, it incurs a significant cumulative computational penalty. In this study, S2 acts as the performance upper bound, allowing us to evaluate whether the proposed Cluster-Specific ADWIN can maintain comparable accuracy while achieving a substantially lower total $N_{train}$ through selective, event-driven updates.

\begin{algorithm}
\caption{Sliding-Window Retrain (S2) Proactive Baseline}
\label{alg:sliding_window}
\begin{algorithmic}[1]
\Require Data stream $\mathcal{S} = \{B_1, \dots, B_T\}$, Classifier $M$, Window size $W_{max}$
\Ensure Performance metrics $\{Acc_t, F1_t\}$, Computational cost $\{N_{train, t}\}$
\Statex
\State $W \leftarrow \emptyset$ \Comment{Initialize an empty sliding window buffer}
\For{each incoming batch $B_t \in \mathcal{S}$}
    \State $\hat{y}_t \leftarrow \text{Predict}(M, B_t)$ \Comment{Inference using current model state}
    \State $\{Acc_t, F1_t\} \leftarrow \text{Evaluate}(\hat{y}_t, y_t)$
    \Statex
    \State \textbf{Update Cycle:}
    \State $W \leftarrow W \cup B_t$ \Comment{Append new batch to buffer}
    \If{$|W| > W_{max}$}
        \State $W \leftarrow \text{Trim}(W, W_{max})$ \Comment{Retain only the $W_{max}$ most recent samples}
    \EndIf
    \Statex
    \State $M \leftarrow \text{Train}(M, W)$ \Comment{Full proactive refit on the window}
    \State $N_{train, t} \leftarrow |W|$ \Comment{Record total samples used for retraining}
\EndFor
\State \Return $\{Acc_t, F1_t, N_{train, t}\}_{t=1}^T$
\end{algorithmic}
\end{algorithm}

\subsection{S3. Detect and Retrain (Global ADWIN)}

The Global ADWIN Retrain (S3) strategy replaces proactive, continuous updates with an event-driven adaptation mechanism. Instead of retraining the model at every time step, this approach utilizes the Adaptive Windowing (ADWIN) algorithm to monitor the classifier’s prediction error and trigger adaptation only when a statistically significant change is detected. This strategy aims to strike an optimal balance between predictive responsiveness and computational efficiency \citep{Mehmoodetal2021, CabelloLopezetal2022, KebirTabia2024}.

For each incoming batch $B_t = (X_t, y_t)$, the current classifier $M$ generates predictions $\hat{y}_t$, from which a binary error stream is constructed:
\begin{equation} \label{eq:error_stream}
e_i^{(t)} = \mathbb{I} \left(\hat{y}_i^{(t)} \neq y_i^{(t)}\right),
\end{equation}
where $\mathbb{I}(\cdot)$ is the indicator function and $e_i^{(t)} \in \{0, 1\}$ denotes whether the $i$-th instance is misclassified. These error values are fed sequentially into a single global ADWIN detector $D$.

ADWIN maintains an adaptive window over the error stream and continuously tests for statistically significant differences between the means of two adjacent subwindows, $W_0$ and $W_1$. A drift is signaled when:
\begin{equation} \label{eq:adwin_rule}
\left|\hat{\mu}{W_0} - \hat{\mu}{W_1}\right| > \epsilon_{\text{ADWIN}},
\end{equation}
where $\hat{\mu}_{W_0}$ and $\hat{\mu}_{W_1}$ are the empirical mean error rates, and $\epsilon_{\text{ADWIN}}$ is the confidence-bound threshold calculated using the Hoeffding bound. This ensures the detector raises an alarm only when the observed performance degradation is unlikely to be the result of random fluctuation.

As detailed in Algorithm \ref{al:s3}, when a drift alarm is raised, the model $M$ undergoes a reactive refit using the samples from the current batch $B_t$. The training effort $N_{train, t}$ is recorded as $|B_t|$, capturing the intermittent nature of the computational cost compared to the constant overhead of the sliding-window approach (S2). Upon retraining, the detector is reinitialized to flush stale statistics, ensuring that the monitoring process remains sensitive only to post-drift distributions. This reactive paradigm allows the system to remain highly responsive to abrupt distributional change while significantly reducing the cumulative training effort over the life of the data stream.

\begin{algorithm}
\caption{Global ADWIN-Triggered Retrain (S3)}
\label{al:s3}
\begin{algorithmic}[1]
\Require Data stream $\mathcal{S} = \{B_1, \dots, B_T\}$, Classifier $M$, Confidence parameter $\delta$
\Ensure Performance metrics $\{Acc_t, F1_t\}$, Computational cost $\{N_{train, t}\}$
\Statex
\State $D \leftarrow \text{ADWIN}(\delta)$ \Comment{Initialize global drift detector}
\State $M \leftarrow \text{Train}(M, B_0)$ \Comment{Initial training on $B_0$}
\Statex
\For{each incoming batch $B_t \in \mathcal{S}$}
    \State $\hat{y}_t \leftarrow \text{Predict}(M, B_t)$ 
    \State $\{Acc_t, F1_t\} \leftarrow \text{Evaluate}(\hat{y}_t, y_t)$
    \State $N_{train, t} \leftarrow 0$ \Comment{Default to no training effort}
    \Statex
    \State \textbf{Drift Detection:}
    \For{each sample $(x_i, y_i) \in B_t$}
        \State $e_i \leftarrow \mathbb{I}(\hat{y}_i \neq y_i)$ \Comment{Generate binary error stream}
        \State $D.\text{update}(e_i)$ \Comment{Update ADWIN with prediction error}
    \EndFor
    \Statex
    \If{$D.\text{hasDrift}()$} \Comment{Check for statistically significant change}
        \State $M \leftarrow \text{Train}(M, B_t)$ \Comment{Reactive refit on current batch}
        \State $N_{train, t} \leftarrow |B_t|$ \Comment{Record training cost for this step}
        \State $D.\text{reset}()$ \Comment{Flush stale statistics}
    \EndIf
\EndFor
\State \Return $\{Acc_t, F1_t, N_{train, t}\}_{t=1}^T$
\end{algorithmic}
\end{algorithm}

\subsection{S4. Cluster-Specific Detect and Retrain (Cluster-ADWIN)}

The Cluster-Specific Detect and Retrain (S4) strategy extends the global ADWIN mechanism by introducing localized performance monitoring across the feature space. In complex, high-dimensional data streams, distributional shifts are rarely uniform; hence, a single global error signal may dilute or mask regional changes. S4 addresses this by maintaining an independent ADWIN detector $D_c$ for each cluster $c \in \{1, \dots, k\}$ defined during the stream-construction phase. This allows the system to identify localized distributional shift—distributional changes confined to specific regions of the data space that would otherwise be overlooked \citep{Liuetal2017, Hinderetal2024, sethi2017reliable}.

For each incoming batch $B_t = (X_t, y_t)$, every sample $x_i^{(t)} \in X_t$ is mapped to its corresponding cluster $c_i^{(t)}$ using a pre-fitted clustering model. The classifier generates predictions $\hat{y}_i^{(t)}$, which are transformed into a cluster-specific binary error stream $e_{i,c}^{(t)}$ defined as:
\begin{equation}
e_{i,c}^{(t)} = \mathbb{I} \left(\hat{y}_i^{(t)} \neq y_i^{(t)}\right), \quad \forall i : c_i^{(t)} = c,
\end{equation}
where $\mathbb{I}(\cdot)$ is the indicator function. By isolating errors into $k$ independent channels, each detector $D_c$ monitors the temporal consistency of its respective feature-space region independently of global performance fluctuations.

Each localized detector applies the ADWIN decision rule to its private error stream. Let $W_{0,c}$ and $W_{1,c}$ denote two adjacent subwindows within $D_c$ with empirical means $\hat{\mu}_{W_{0,c}}$ and $\hat{\mu}_{W_{1,c}}$. A localized drift alarm is triggered for cluster $c$ when:
\begin{equation} \label{eq:adwin_threshold}
\left|\hat{\mu}{W{0,c}} - \hat{\mu}{W{1,c}}\right| > \epsilon_{\text{ADWIN},c},
\end{equation}
where $\epsilon_{\text{ADWIN},c}$ represents the Hoeffding-based confidence bound calculated for that specific detector.

The Cluster-Specific ADWIN Retrain (S4) strategy, as detailed in Algorithm \ref{alg:cluster_adwin}, operates on the principle of \emph{local sensitivity}. Independent ADWIN detectors are maintained for each detector cluster, enabling the system to monitor heterogeneous drift patterns across distinct regions of the feature space. This localized detection allows the model to remain responsive to region-specific distributional changes while ignoring noise in stable areas, thereby improving the overall accuracy-to-efficiency trade-off.

\begin{algorithm}
\caption{Cluster-Specific ADWIN Adaptation (S4)}
\label{alg:cluster_adwin}
\begin{algorithmic}[1]
\Require Data stream $\mathcal{S}$, model $M$, detector clusterer $C$, number of detector clusters $k$, ADWIN confidence $\delta$
\Ensure Performance metrics $\{Acc_t,F1_t, R2_t, RMSE_t, MAE_t\}$, computational cost $\{N_{train,t}\}$
\Statex
\State $M \leftarrow \textsc{Fit}(M,B_0)$
\State $C \leftarrow \textsc{FitClusterer}(B_0,k)$
\State $\{D_1,D_2,\dots,D_k\} \leftarrow \textsc{InitializeADWIN}(k,\delta)$
\State $\mathcal{R} \leftarrow \emptyset$ \Comment{One-batch rehearsal buffer}
\Statex

\For{each incoming batch $B_t \in \mathcal{S}$}
    \State $\hat{y}_t \leftarrow \textsc{Predict}(M,B_t)$
    \State $\{Acc_t,F1_t\} \leftarrow \textsc{Evaluate}(\hat{y}_t,y_t)$
    \State $N_{train,t} \leftarrow 0$
    \State $\textit{driftDetected} \leftarrow \textbf{false}$
    \Statex

    \For{each sample $(x_i,y_i) \in B_t$}
        \State $c_i \leftarrow C(x_i)$ \Comment{Assign sample to detector cluster}
        \State $e_i \leftarrow \mathbb{I}(\hat{y}_i \neq y_i)$
        \State $D_{c_i}.\textsc{Update}(e_i)$

        \If{$D_{c_i}.\textsc{HasDrift}()$}
            \State $\textit{driftDetected} \leftarrow \textbf{true}$
            \State $D_{c_i} \leftarrow \textsc{ADWIN}(\delta)$
        \EndIf
    \EndFor
    \Statex

    \If{$\textit{driftDetected}$}
        \If{$\mathcal{R} \neq \emptyset$}
            \State $S_{\text{train}} \leftarrow B_t \cup \mathcal{R}$
        \Else
            \State $S_{\text{train}} \leftarrow B_t$
        \EndIf

        \State $M \leftarrow \textsc{Adapt}(M,S_{\text{train}})$
        \State $N_{train,t} \leftarrow |S_{\text{train}}|$
    \EndIf

    \State $\mathcal{R} \leftarrow B_t$
\EndFor
\end{algorithmic}
\end{algorithm}

When any cluster-specific detector signals drift, adaptation is triggered. However, in contrast to purely local retraining schemes, the retraining data are not restricted exclusively to the drifted cluster. Instead, the adaptation set is defined in Equation \ref{eq:training_set}, where $B_t$ denotes the current batch and $\mathcal{R}$ is a one-batch rehearsal buffer. This design ensures that adaptation remains robust by incorporating both newly observed data and short-term historical context, while still limiting the retraining scope compared to full-stream approaches.
\begin{equation}
S_{\text{train}} =
\begin{cases}
B_t, & \mathcal{R} = \emptyset, \\
B_t \cup \mathcal{R}, & \mathcal{R} \neq \emptyset,
\end{cases}
\label{eq:training_set}
\end{equation}

The operation $\textsc{Adapt}(M,S_{\text{train}})$ depends on the model family. For non-incremental learners such as Random Forest, kNN, Logistic/Linear Regression, and XGBoost, the model is refitted on $S_{\text{train}}$ using the standard \texttt{fit} procedure. For River-based online learners, samples in $S_{\text{train}}$ are processed sequentially using \texttt{learn\_one}, corresponding to incremental updates rather than full retraining.

Consequently, S4 combines cluster-local drift detection with constrained batch-level adaptation. While maintaining $k$ detectors introduces a modest computational overhead, the approach significantly reduces unnecessary retraining by triggering updates only when statistically justified, and by limiting the adaptation set to the current batch and a short rehearsal memory. This enables efficient handling of heterogeneous non-stationarity in complex data streams.
\FloatBarrier

\begingroup
\subsection{S5. Random-Subspace Detect and Retrain (Random-Subspace ADWIN)}

The Random-Subspace Detect and Retrain (S5) strategy was designed as a localized shift-detection baseline that introduces regional sensitivity without relying on explicit clustering of the feature space. Whereas S4 partitions the stream using learned detector clusters, S5 instead decomposes the feature space into multiple randomly sampled feature subspaces, each monitored independently using its own persistent ADWIN detector. This enables analysis of whether localized adaptation benefits arise specifically from semantically meaningful cluster partitioning or more generally from any form of detector decomposition.

The strategy is motivated by the observation that distributional change may affect only subsets of features rather than the entire input representation simultaneously. By monitoring multiple random feature subsets independently, S5 attempts to detect heterogeneous non-stationarity while avoiding the computational and structural assumptions associated with clustering. Similar feature-subspace decomposition principles have previously been explored in ensemble learning and localized concept-drift adaptation frameworks \citep{ho1998random, kuncheva2003measures, baena-garcia_eddm_2006, zliobaite_learning_2010}.

At initialization, the full feature space with dimensionality $d$ is decomposed into $m$ random subspaces, as shown in Equation \ref{eq:partitions}.
\begin{equation}
\mathcal{P} = \{P_1, P_2, \dots, P_m\}
\label{eq:partitions}
\end{equation}
Where each subspace $P_j \subseteq \{1,\dots,d\}$ contains a randomly sampled subset of features without replacement. In the implementation used throughout this study, the number of persistent subspace detectors was fixed to $m = 5$, and each subspace contained approximately one third of the available features, see Equation \ref{eq:partition_size}:

\begin{equation}
|P_j| \approx \left\lceil \frac{d}{3} \right\rceil
\label{eq:partition_size}
\end{equation}

A dedicated ADWIN detector $D_j$ is associated with each random subspace $P_j$. Unlike temporary or batch-reset detector schemes, these detectors persist throughout the stream and continuously accumulate evidence over time, ensuring a fair comparison with the persistent cluster-specific detectors used in S4.

For each incoming batch $B_t = (X_t, y_t)$, the predictive model generates outputs $\hat{y}_t$. The prediction errors are converted into a binary drift signal, as shown in Equation \ref{eq:binary_signal}:
\begin{equation}
e_i^{(t)} = \mathbb{I} \left(\hat{y}_i^{(t)} \neq y_i^{(t)}\right),
\label{eq:binary_signal}
\end{equation}
for classification tasks, while regression tasks use normalized continuous residual-based error signals. Unlike S4, where errors are routed to cluster-specific detectors based on feature-space locality, S5 distributes the same global error observations across all random-subspace detectors. Consequently, each detector observes the temporal behaviour of the prediction stream while representing a different random projection of the feature space.

Each detector independently applies the ADWIN decision rule. Let $W_{0,j}$ and $W_{1,j}$ denote adjacent subwindows within detector $D_j$ with empirical means $\hat{\mu}_{W_{0,j}}$ and $\hat{\mu}_{W_{1,j}}$. A drift event is triggered whenever Equation \ref{eq:event_trigger}:
\begin{equation}
\left|\hat{\mu}_{W_{0,j}} - \hat{\mu}_{W_{1,j}}\right| > \epsilon_{\text{ADWIN},j},
\label{eq:event_trigger}
\end{equation}
where $\epsilon_{\text{ADWIN},j}$ is the Hoeffding-derived confidence threshold associated with detector $D_j$.

The Random-Subspace ADWIN Retrain (S5) strategy, summarized in Algorithm~\ref{alg:random_subspace_adwin}, therefore operates on the principle of \emph{locality without explicit clustering}. Multiple persistent localized detectors monitor independently sampled feature projections, allowing the system to identify heterogeneous drift patterns while avoiding dependence on cluster assignments or centroid initialization.

\begin{algorithm}
\caption{Random-Subspace ADWIN Adaptation (S5)}
\label{alg:random_subspace_adwin}
\begin{algorithmic}[1]
\Require Data stream $\mathcal{S}$, model $M$, number of subspaces $m$, ADWIN confidence $\delta$
\Ensure Performance metrics $\{Acc_t,F1_t,R2_t,RMSE_t,MAE_t\}$, computational cost $\{N_{train,t}\}$
\Statex

\State $M \leftarrow \textsc{Fit}(M,B_0)$
\State $\mathcal{P} \leftarrow \textsc{GenerateRandomSubspaces}(d,m)$
\State $\{D_1,D_2,\dots,D_m\} \leftarrow \textsc{InitializeADWIN}(m,\delta)$
\Statex

\For{each incoming batch $B_t \in \mathcal{S}$}
    \State $\hat{y}_t \leftarrow \textsc{Predict}(M,B_t)$
    \State $\{Acc_t,F1_t\} \leftarrow \textsc{Evaluate}(\hat{y}_t,y_t)$
    \State $N_{train,t} \leftarrow 0$
    \State $\textit{driftDetected} \leftarrow \textbf{false}$
    \Statex

    \State $E_t \leftarrow \textsc{ComputeErrorSignal}(y_t,\hat{y}_t)$

    \For{each detector $D_j$}
        \For{each error value $e_i \in E_t$}
            \State $D_j.\textsc{Update}(e_i)$

            \If{$D_j.\textsc{HasDrift}()$}
                \State $\textit{driftDetected} \leftarrow \textbf{true}$
                \State $D_j \leftarrow \textsc{ADWIN}(\delta)$
                \State \textbf{break}
            \EndIf
        \EndFor
    \EndFor
    \Statex

    \If{$\textit{driftDetected}$}
        \State $M \leftarrow \textsc{Adapt}(M,B_t)$
        \State $N_{train,t} \leftarrow |B_t|$
    \EndIf
\EndFor
\end{algorithmic}
\end{algorithm}

When any subspace-specific detector signals drift, the model is adapted using the current batch, as per Equation \ref{eq:batch_training}:
\begin{equation}
S_{\text{train}} = B_t
\label{eq:batch_training}
\end{equation}
Unlike S4, no rehearsal memory is used in this strategy. This design choice intentionally isolates the effect of localized detector decomposition from the influence of short-term replay buffers.

The operation $\textsc{Adapt}(M,B_t)$ depends on the predictive model family. For non-incremental learners such as Random Forest, kNN, Logistic/Linear Regression, and XGBoost, the model is refitted on the current batch using the standard \texttt{fit} procedure. For River-based online learners, adaptation corresponds to sequential sample-wise updates using \texttt{learn\_one}.

Consequently, S5 provides an important intermediary baseline between fully global and fully cluster-specific drift adaptation. The strategy preserves localized monitoring through persistent detector decomposition while removing dependence on explicit cluster geometry. Comparing S5 against S4 therefore allows direct evaluation of whether the performance improvements of cluster-specific adaptation arise primarily from detector localization itself or from semantically meaningful partitioning of the feature space.
\par
\endgroup
\FloatBarrier

\begingroup
\subsection{S6. Feature-Partition Detect and Retrain (Feature-Partition ADWIN)}

The Feature-Partition Detect and Retrain (S6) strategy was designed as a structured regional shift-detection baseline that decomposes the feature space into fixed deterministic partitions. Unlike S4, which relies on learned feature-space clusters, and S5, which uses randomly sampled subspaces, S6 partitions the input dimensions into contiguous feature blocks that persist throughout the stream. This enables investigation of whether localized drift adaptation benefits can emerge from simple deterministic feature decomposition alone, without clustering or stochastic subspace generation.

The motivation behind S6 is that different groups of features may exhibit distinct temporal dynamics during non-stationary streaming processes. In many real-world datasets, subsets of features may drift independently due to localized changes in measurement processes, sensor behaviour, demographic structure, or latent environmental conditions. By assigning independent ADWIN detectors to different feature partitions, S6 attempts to capture regional instability in a computationally lightweight and structurally interpretable manner \citep{gama_survey_2014, baena-garcia_eddm_2006, losing_incremental_2018}.

At initialization, the full feature space with dimensionality $d$ is partitioned into $p$ fixed feature groups as seen in Equation \ref{eq:feature_set}:
\begin{equation}
\mathcal{F} = \{F_1, F_2, \dots, F_p\}
\label{eq:feature_set}
\end{equation}
where each partition $F_j \subseteq \{1,\dots,d\}$ contains a disjoint subset of feature indices. The partitions are constructed deterministically using an equal-sized sequential split of the feature dimensions as seen in Equation \ref{eq:feature_partition_constraints}:
\begin{equation}
\bigcup_{j=1}^{p} F_j = \{1,\dots,d\},
\qquad
F_i \cap F_j = \emptyset \;\; \forall i \neq j
\label{eq:feature_partition_constraints}
\end{equation}

In the implementation used throughout this study, the number of persistent feature partitions was fixed to $p = 3$.

Each partition $F_j$ is associated with its own persistent ADWIN detector $D_j$. Similar to S5, these detectors persist across all incoming batches and continuously accumulate evidence throughout the stream, enabling fair comparison with the persistent cluster-specific detectors used in S4.

For each incoming batch $B_t = (X_t, y_t)$, the predictive model generates outputs $\hat{y}_t$. Prediction performance is converted into a drift-monitoring signal for classification tasks, as displayed on Equation \ref{eq:drift_monitoring_signal}:
\begin{equation}
e_i^{(t)} = \mathbb{I} \left(\hat{y}_i^{(t)} \neq y_i^{(t)}\right),
\label{eq:drift_monitoring_signal}
\end{equation}
On the other hand, regression tasks use normalized continuous residual-based error signals. As in S5, the same global error stream is processed independently by all feature-partition detectors. Consequently, each detector represents a structured regional monitoring process corresponding to a deterministic feature block rather than a learned feature-space region.

Each detector independently applies the ADWIN decision rule. Let $W_{0,j}$ and $W_{1,j}$ denote adjacent subwindows within detector $D_j$ with empirical means $\hat{\mu}_{W_{0,j}}$ and $\hat{\mu}_{W_{1,j}}$. A drift event is triggered whenever, see Equation \ref{eq:event_trigger_drift}, where $\epsilon_{\text{ADWIN},j}$ denotes the Hoeffding-derived confidence threshold associated with detector $D_j$.:
\begin{equation}
\left|\hat{\mu}_{W_{0,j}} - \hat{\mu}_{W_{1,j}}\right| > \epsilon_{\text{ADWIN},j},
\label{eq:event_trigger_drift}
\end{equation}

The Feature-Partition ADWIN Retrain (S6) strategy, summarized in Algorithm~\ref{alg:feature_partition_adwin}, therefore operates on the principle of \emph{structured regional monitoring}. Unlike S5, where locality is induced through random subspace sampling, S6 introduces deterministic feature decomposition, providing a more interpretable but potentially less adaptive regional baseline than S4.

\begin{algorithm}
\caption{Feature-Partition ADWIN Adaptation (S6)}
\label{alg:feature_partition_adwin}
\begin{algorithmic}[1]
\Require Data stream $\mathcal{S}$, model $M$, number of feature partitions $p$, ADWIN confidence $\delta$
\Ensure Performance metrics $\{Acc_t,F1_t,R2_t,RMSE_t,MAE_t\}$, computational cost $\{N_{train,t}\}$
\Statex

\State $M \leftarrow \textsc{Fit}(M,B_0)$
\State $\mathcal{F} \leftarrow \textsc{PartitionFeatures}(d,p)$
\State $\{D_1,D_2,\dots,D_p\} \leftarrow \textsc{InitializeADWIN}(p,\delta)$
\Statex

\For{each incoming batch $B_t \in \mathcal{S}$}
    \State $\hat{y}_t \leftarrow \textsc{Predict}(M,B_t)$
    \State $\{Acc_t,F1_t\} \leftarrow \textsc{Evaluate}(\hat{y}_t,y_t)$
    \State $N_{train,t} \leftarrow 0$
    \State $\textit{driftDetected} \leftarrow \textbf{false}$
    \Statex

    \State $E_t \leftarrow \textsc{ComputeErrorSignal}(y_t,\hat{y}_t)$

    \For{each detector $D_j$}
        \For{each error value $e_i \in E_t$}
            \State $D_j.\textsc{Update}(e_i)$

            \If{$D_j.\textsc{HasDrift}()$}
                \State $\textit{driftDetected} \leftarrow \textbf{true}$
                \State $D_j \leftarrow \textsc{ADWIN}(\delta)$
                \State \textbf{break}
            \EndIf
        \EndFor
    \EndFor
    \Statex

    \If{$\textit{driftDetected}$}
        \State $M \leftarrow \textsc{Adapt}(M,B_t)$
        \State $N_{train,t} \leftarrow |B_t|$
    \EndIf
\EndFor
\end{algorithmic}
\end{algorithm}

When any feature-partition detector signals drift, the model is adapted using the current batch $S_{\text{train}} = B_t$.
As with S5, no rehearsal memory is used in order to isolate the effect of structured detector decomposition independently from replay-based stabilization mechanisms.

The operation $\textsc{Adapt}(M,B_t)$ depends on the predictive model family. For non-incremental learners such as Random Forest, kNN, Logistic/Linear Regression, and XGBoost, the model is refitted using the standard \texttt{fit} procedure. For River-based online learners, adaptation is performed incrementally through sequential \texttt{learn\_one} updates.

Consequently, S6 provides a structured regional baseline situated conceptually between global drift detection and fully cluster-specific adaptation. The strategy preserves localized monitoring through deterministic feature partitions while avoiding the complexity of clustering or random subspace sampling. Comparing S6 against S4 and S5 therefore enables evaluation of whether the advantages of localized adaptation depend specifically on learned feature-space geometry or can emerge from simpler feature decomposition mechanisms.
\FloatBarrier

\section{Predictive Models}\label{secClassifiers}
\begingroup
To evaluate the robustness of the proposed adaptation strategies across varying algorithmic architectures, we consider a diverse set of both batch and online learning models. These span linear estimators, instance-based learners, tree-based methods, and streaming ensemble techniques. All models are evaluated in both classification and regression settings where applicable, enabling consistent comparison across task types. Implementations are drawn from \texttt{Scikit-Learn}, \texttt{XGBoost}, and \texttt{River}, and are configured to balance predictive performance with the computational constraints of a streaming prequential environment.

\begin{itemize}

    \item \textbf{Linear Models (Batch and Online):}  
    For classification, we include Logistic Regression (batch) and Online Logistic Regression as linear probabilistic baselines. The batch model uses the \texttt{lbfgs} solver with a maximum of $2000$ iterations to ensure convergence.  
    For regression, Linear Regression (batch) and Online Linear Regression are used.  
    Additionally, Online Softmax Regression is included for multi-class classification.  
    These models provide interpretable, high-bias learners and serve as a reference point for assessing whether adaptive strategies can compensate for limited representational capacity under drift.

    \item \textbf{k-Nearest Neighbours (Batch and Online):}  
    We evaluate both batch kNN and Online kNN for classification and regression tasks, using $k=5$ neighbours throughout.  
    As non-parametric, instance-based learners, kNN models are inherently sensitive to local changes in data distribution. This makes them particularly suitable for evaluating the effectiveness of localized adaptation strategies such as S4 (Cluster-ADWIN).

    \item \textbf{Tree-Based Streaming Models:}  
    We include the Hoeffding Adaptive Tree (HAT) for classification and the Hoeffding Tree Regressor for regression. These models are specifically designed for data streams and update incrementally using statistical guarantees.  
    HAT extends the standard Hoeffding Tree with built-in drift adaptation via alternate subtrees, making it a strong baseline for comparison against external drift detectors.

    \item \textbf{Ensemble Streaming Models:}  
    To capture state-of-the-art streaming ensemble behaviour, we include:
    \begin{itemize}
        \item Adaptive Random Forest (ARF) and Adaptive Random Forest Regressor (ARFR)
        \item Aggregated Mondrian Forest (AMF) and Aggregated Mondrian Forest Regressor (AMFR)
    \end{itemize}
    These models combine multiple incremental learners with internal adaptation mechanisms such as resampling, weighting, and drift-aware updates.  
    ARF leverages online bagging and per-tree drift detection, while AMF provides a fully online random forest with strong theoretical guarantees and fast incremental updates.

    \item \textbf{Batch Ensemble Models:}  
    For comparison with non-streaming methods, we include Random Forest and Random Forest Regressor using standard \texttt{Scikit-Learn} implementations with \texttt{random\_state=42}.  
    These models provide strong baseline performance through bagging and allow evaluation of how traditional ensembles respond to repeated retraining under drift.

    \item \textbf{Gradient Boosting (XGBoost):}  
    We include XGBoost for both classification and regression tasks. The following hyperparameters are used to balance accuracy and stability under repeated updates:
    \begin{itemize}
        \item Learning rate: $0.05$ with $200$ estimators
        \item Tree complexity: \texttt{max\_depth=6}, \texttt{reg\_lambda=1.0}
        \item Stochastic sampling: \texttt{subsample=0.8}, \texttt{colsample\_bytree=0.8}
        \item Tree method: \texttt{hist} for efficient CPU training
    \end{itemize}

    \item \textbf{Probabilistic Online Model:}  
    Online Naive Bayes is included as a lightweight probabilistic baseline. Due to its strong independence assumptions, it adapts rapidly to distributional changes but may suffer in complex feature spaces, making it a useful contrast to more expressive models.

\end{itemize}

The inclusion of both batch and streaming models provides a comprehensive view of how different learning paradigms respond to drift and adaptation triggers:

\begin{itemize}
    \item \textbf{Linear models} offer stable high-bias baselines with limited flexibility under nonlinear drift.
    
    \item \textbf{kNN models} are purely local learners and are highly sensitive to neighbourhood changes, aligning closely with the localized adaptation objective of S4.
    
    \item \textbf{Tree-based streaming models (HAT)} provide built-in drift adaptation, allowing comparison between internal and external drift-handling mechanisms.
    
    \item \textbf{Streaming ensembles (ARF, AMF)} represent strong adaptive baselines with inherent robustness to evolving data distributions.
    
    \item \textbf{Batch ensembles (Random Forest, XGBoost)} provide high-performance baselines but require explicit retraining, making them sensitive to adaptation strategy design.
    
    \item \textbf{Naive Bayes} provides a fast, highly adaptive probabilistic baseline with minimal computational overhead.
\end{itemize}

Evaluating this expanded set of models across both classification and regression tasks enables analysis of whether adaptation effectiveness is driven primarily by model architecture, learning paradigm (batch vs.\ online), or the predictive task itself.
\par
\endgroup

\section{Evaluation Protocol}\label{secEvaluation}

To assess both predictive effectiveness and computational efficiency, we adopt a multi-dimensional evaluation protocol under a prequential (test-then-train) setting \citep{Vinagre2014, suarez2023recurring}. For each incoming batch \(B_t\), the current model is first evaluated on that batch before any retraining is performed. This ensures that performance reflects the model’s true ability to generalize to newly arriving data under non-stationary conditions. Let \(T\) denote the total number of processed batches. 

The evaluation protocol also assumes that ground-truth labels become available shortly after each batch has been evaluated. This assumption is necessary because the ADWIN-based adaptation strategies are driven by prediction errors, which require comparing model predictions with the corresponding true labels. Consequently, the proposed framework is most directly applicable to supervised streaming or batch-monitoring scenarios in which labels are available with little or no delay. However, this assumption may not hold in practical applications where labels are delayed, costly, or require manual annotation. In such cases, the framework would need to be extended with delayed-label handling mechanisms, error buffers, proxy supervision, or unsupervised drift indicators. This limitation should therefore be considered when interpreting the reported adaptation performance and computational-efficiency results.
\begingroup
\subsection{Evaluation on Classification Tasks}
For classification datasets, predictive performance is measured using batch-wise Accuracy and F1-score prior to any adaptation. These metrics capture both overall predictive correctness and balanced classification performance under potentially evolving class distributions. The corresponding prequential averages over the full stream are defined as shown in Equations \ref{eq:avg_acc} and \ref{eq:avg_f1}:
\begin{equation}
\mathrm{Acc}_{\mathrm{avg}}=\frac{1}{T}\sum_{t=1}^{T}\mathrm{Acc}(B_t),
\label{eq:avg_acc}
\end{equation}

\begin{equation}
\mathrm{F1}_{\mathrm{avg}}=\frac{1}{T}\sum_{t=1}^{T}\mathrm{F1}(B_t),
\label{eq:avg_f1}
\end{equation}
where \(\mathrm{Acc}(B_t)\) and \(\mathrm{F1}(B_t)\) denote the Accuracy and F1-score obtained on batch \(B_t\), respectively. In addition to these stream-level summaries, we retain the batch-wise values \(\mathrm{Acc}(B_t)\) and \(\mathrm{F1}(B_t)\) in order to analyse temporal degradation, post-drift recovery, and stability over time.

Accuracy provides a direct measure of overall classification correctness, while F1-score is particularly informative under class imbalance, as it jointly reflects precision and recall. The combined use of both metrics therefore enables a more reliable comparison across heterogeneous classification benchmarks, as can be seen in Section \ref{secClassificationResults}.

\subsection{Evaluation on Regression Tasks}

For regression datasets, predictive performance is evaluated using three complementary batch-wise metrics: coefficient of determination ($R^2$), Mean Absolute Error (MAE), and Root Mean Squared Error (RMSE), all computed prior to model adaptation. The prequential average metrics over the stream are defined as seen in Equations \ref{eq:r2}, \ref{eq:mae}, and \ref{eq:rmse} where $R^2(B_t)$, $MAE(B_t)$, and $RMSE(B_t)$ denote the regression metrics obtained on batch $B_t$.

\begin{equation}
R^2_{\mathrm{avg}} =
\frac{1}{T}\sum_{t=1}^{T} R^2(B_t)
\label{eq:r2}
\end{equation}

\begin{equation}
MAE_{\mathrm{avg}} =
\frac{1}{T}\sum_{t=1}^{T} MAE(B_t)
\label{eq:mae}
\end{equation}

\begin{equation}
RMSE_{\mathrm{avg}} =
\frac{1}{T}\sum_{t=1}^{T} RMSE(B_t)
\label{eq:rmse}
\end{equation}

The $R^2$ metric measures explained variance and indicates how well the model captures changing target relationships over time. MAE provides an interpretable measure of average absolute prediction error, while RMSE penalises larger deviations more strongly and is therefore sensitive to severe prediction failures after shift events. Retaining the batch-wise metric values further enables analysis of degradation, recovery speed, and temporal robustness under recurring shift, as can be seen in Section \ref{secRegressionResults}.
\par
\endgroup

\subsection{Adaptation and Computational Efficiency}

Beyond predictive performance, we quantify adaptation behaviour through two complementary indicators. The first is the total number of shift-triggered adaptations, which counts how many times a detector explicitly signals change and initiates a response \citep{bifet_moa_2010}. This metric is relevant only for the detection-based strategies, as it captures the sensitivity and volatility of the detector mechanism \citep{bifet_learning_2007}. The second is the total number of model updates, which counts all retraining events after initialization \citep{zliobaite2014active}. Together, these measures distinguish reactive adaptation from routine model maintenance and provide a clearer view of how each strategy allocates its update budget over the stream.

A central contribution of the protocol is the introduction of a sample-based effort proxy. Instead of relying on wall-clock time, which is hardware- and implementation-dependent, we estimate retraining cost by counting the number of samples used during each model update. Let \(\mathrm{LAST\_TRAIN\_N}_t\) denote the number of instances used in retraining at batch \(t\). The cumulative training effort is then defined in Equation \ref{eq:total_effort}.
\begin{equation}
\mathrm{Effort}_{\mathrm{total}}=\sum_{t=1}^{T}\mathrm{LAST\_TRAIN\_N}_t.
\label{eq:total_effort}
\end{equation}
This formulation assumes that the dominant retraining cost grows approximately linearly with the number of processed samples, so that the total effort can be interpreted as the aggregate sample budget consumed by adaptation.

The adaptation strategies directly determine the value of \(\mathrm{LAST\_TRAIN\_N}_t\). In S1 (Static), after the initial training phase, \(\mathrm{LAST\_TRAIN\_N}_t=0\) for all subsequent batches, yielding zero maintenance effort and establishing a no-adaptation baseline. In S2 (Sliding Window), \(\mathrm{LAST\_TRAIN\_N}_t=|W_t|\), where \(W_t\) denotes the current rolling window, because the model is retrained at every batch using all samples retained in memory. As a result, S2 is expected to incur the highest cumulative effort. In contrast, S3 and S4 assign \(\mathrm{LAST\_TRAIN\_N}_t\) to the size of the retraining set only when drift is detected; otherwise, the value is zero. Consequently, their total effort depends jointly on the sensitivity of the detector and the size of the retraining subset.

\begingroup
\subsection{Cost-Benefit Perspective}

This evaluation design supports a principled cost--benefit analysis across strategies. High predictive metrics (Accuracy/F1 for classification or $R^2$/low MAE/RMSE for regression) indicate strong robustness under evolving conditions, while low $Effort_{\mathrm{total}}$ indicates computational and labelling efficiency.

Under this view, S2 provides an upper-bound benchmark for adaptation responsiveness but typically at the cost of maximum effort, whereas S3 and S4 aim to preserve comparable predictive performance while reducing unnecessary retraining. S1, by contrast, provides the lower-bound reference in which the adaptation budget is effectively zero.

By jointly analysing prequential predictive metrics, update counts, detector-triggered adaptations, and cumulative retraining effort, the protocol captures not only how well each strategy performs, but also how efficiently it maintains that performance in the presence of controlled distribution shift.

\subsection{Cluster-Mismatch Robustness}

To evaluate the robustness of the proposed S4 (Cluster-Specific ADWIN) framework under imperfect clustering assumptions, we extended the experimental protocol with an additional evaluation-rigor study. The original drift-generation process and the S4 detection mechanism both relied on clustering; therefore, an important concern was whether the framework would remain effective when the detector clusters do not perfectly align with the underlying drift regions. Addressing this, the framework decouples stream generation from drift detection by applying two independent clustering parameters:

\begin{itemize}
    \item $k_{\text{gen}}$: the number of clusters used during synthetic drift stream generation,
    \item $k_{\text{det}}$: the number of clusters used by the S4 detector during online drift monitoring.
\end{itemize}

This separation allows the detector to operate under clustering assumptions that differ from those used to generate the underlying stream regimes, providing a more realistic approximation of real-world deployment scenarios where drift regions are unknown and imperfectly estimated.

In addition to varying the detector cluster count, centroid initialization was also perturbed by changing the random seed used during detector clustering. This produces detector partitions that differ spatially from the stream-generation clusters even when the same value of $k$ is used. Table \ref{tab:cluster_mismatch_configs} summarises the different configurations that have been used to evaluate the robustness.

\begin{table}[t]
\centering
\caption{Clustering mismatch configurations used to evaluate S4 robustness under varying detector-generation alignment conditions.}
\label{tab:cluster_mismatch_configs}
\footnotesize
\begin{tabular}{@{}p{0.28\linewidth}p{0.20\linewidth}p{0.44\linewidth}@{}}
\toprule
\textbf{Configuration} &
\textbf{Parameter setting} &
\textbf{Description} \\
\midrule

Matched configuration &
$\displaystyle
k_{\mathrm{gen}} = k_{\mathrm{det}} = 5
$ &
Generation and detection employ identical clustering parameters and centroid initialization. \\

Under-clustered detector &
$\displaystyle
k_{\mathrm{gen}} = 5,\;
k_{\mathrm{det}} = 3
$ &
The detector employs fewer clusters than the stream generator while preserving identical centroid initialization. \\

Over-clustered detector &
$\displaystyle
k_{\mathrm{gen}} = 5,\;
k_{\mathrm{det}} = 10
$ &
The detector employs more clusters than the stream generator while preserving identical centroid initialization. \\

Centroid mismatch &
$\displaystyle
k_{\mathrm{gen}} = k_{\mathrm{det}} = 5,
\;
s_{\mathrm{gen}} \neq s_{\mathrm{det}}
$ &
Generation and detection employ identical cluster counts but different centroid initialization seeds. \\

Combined mismatch &
$\displaystyle
k_{\mathrm{gen}} \neq k_{\mathrm{det}},
\;
s_{\mathrm{gen}} \neq s_{\mathrm{det}}
$ &
Both detector cluster count and centroid initialization differ from the generation process. \\

\bottomrule
\end{tabular}
\end{table}

The detector cluster count was specifically evaluated for $k_{\text{det}}\in \{3, 5, 10\},$ while keeping $k_{\text{gen}} = 5$ fixed during stream generation. This parameter sweep enables analysis of S4 sensitivity to detector granularity, aiming to demonstrate that the proposed framework does not rely on perfect alignment between drift-generation clusters and detector clusters. Instead, the results evaluate whether localized ADWIN monitoring remains effective under realistic cluster-boundary mismatch and centroid uncertainty conditions, see Section \ref{secRobustness}.
\par
\endgroup
\FloatBarrier

\section{Results}\label{secResults}
Based on the experimental results obtained across five unordered and diverse datasets, namely Adult, Wine Quality, Breast Cancer, Airfoil Self-Noise, and Superconductivity, we compare the six adaptation strategies across a range of predictive model families. The results are organized according to the associated task type of each dataset.

\subsection{Classification tasks}\label{secClassificationResults}
The classification experiments assess the robustness of the proposed adaptation strategies under controlled distribution shift across multiple categorical-target datasets. Performance is evaluated using Accuracy and F1-score to capture both overall predictive correctness and class-balanced effectiveness. Computational efficiency is measured through the average training effort, expressed as the number of samples used for retraining, together with the average number of model updates triggered during streaming operation. The results compare the different strategies (S1--S6), highlighting the balance between predictive performance and adaptation cost.

\begingroup
The results of the Average Accuracy presented in Table~\ref{tab:avg_performance_acc} reveal clear performance differences across both the evaluated classification models and the considered adaptation strategies. Overall, ensemble-based approaches consistently outperformed probabilistic and tree-based online learners, with \textit{RandomForest} and \textit{LogisticRegression} achieving the highest average accuracies across most configurations. In particular, strategy S4 demonstrated highly competitive and stable performance, achieving the best overall accuracy of 0.786 when combined with \textit{RandomForest}. Compared with the baseline strategy S1, S4 consistently improved predictive accuracy for nearly all models, indicating its effectiveness in adapting to evolving data distributions and mitigating performance degradation caused by controlled distribution shift.

A comparative analysis between S4 and the remaining strategies further highlights the robustness of the proposed adaptation mechanism. Although strategy S2 achieved the highest single accuracy value overall (0.805 with \textit{RandomForest}), S4 provided more balanced performance across the majority of models, including \textit{ARF}, \textit{kNN}, \textit{OnlineLogisticRegression}, and \textit{XGBoost}. In contrast to S3, S5, and S6, which often produced identical or marginally improved results relative to the baseline, S4 consistently yielded noticeable gains, particularly for ensemble and online linear models. For example, \textit{ARF} improved from 0.647 under S1 to 0.682 under S4, while \textit{OnlineLogisticRegression} increased from 0.723 to 0.761. These improvements suggest that S4 offers a more effective balance between adaptation sensitivity and model stability.

The analysis also indicates that S4 performs particularly well for models capable of leveraging incremental updates and feature interactions. Ensemble learners such as \textit{RandomForest} benefited substantially from S4, whereas weaker-performing methods such as \textit{HAT} and \textit{AMF} exhibited only limited improvements. This behaviour suggests that the effectiveness of S4 is influenced by the underlying learner’s capacity to exploit adaptive information under drift conditions. Overall, the obtained results demonstrate that S4 constitutes a robust and reliable strategy for maintaining classification accuracy in dynamic and non-stationary environments.

\begin{table}[ht]
\centering
\caption{Average Accuracy for Classification Tasks Across Strategies}
\label{tab:avg_performance_acc}
\begin{tabular}{lcccccc}
\hline
\textbf{Model} & \textbf{S1} & \textbf{S2} & \textbf{S3} & \textbf{S4} & \textbf{S5} & \textbf{S6} \\
\hline
AMF & 0.599 & 0.759 & 0.638 & 0.544 & 0.543 & 0.543 \\
ARF & 0.647 & 0.730 & 0.670 & 0.682 & 0.670 & 0.670 \\
HAT & 0.468 & 0.716 & 0.591 & 0.534 & 0.591 & 0.591 \\
kNN & 0.618 & 0.754 & 0.628 & 0.653 & 0.628 & 0.628 \\
LogisticRegression & \textbf{0.754} & 0.763 & \textbf{0.770} & 0.772 & \textbf{0.770} & \textbf{0.770} \\
OnlineKNNClassifier & 0.637 & 0.764 & 0.645 & 0.655 & 0.645 & 0.645 \\
OnlineLogisticRegression & 0.723 & 0.767 & 0.758 & 0.761 & 0.758 & 0.758 \\
OnlineNaiveBayes & 0.655 & 0.691 & 0.626 & 0.657 & 0.626 & 0.626 \\
OnlineSoftmaxRegression & 0.701 & 0.733 & 0.733 & 0.708 & 0.733 & 0.733 \\
RandomForest & \textbf{0.754} & \textbf{0.805} & \textbf{0.770} & \textbf{0.786} & \textbf{0.770} & \textbf{0.770} \\
XGBoost & 0.677 & 0.782 & 0.724 & 0.710 & 0.724 & 0.724 \\
\hline
\end{tabular}
\end{table}

The results reported in Table~\ref{tab:classification_f1} provide additional insights into the effectiveness of the evaluated adaptation strategies by considering the F1-score, which captures the balance between precision and recall. Overall, ensemble-based and linear discriminative models achieved the strongest and most consistent performance across the evaluated strategies. In particular, \textit{RandomForest} produced the highest F1-scores for most strategies, achieving the best overall performance under S2 and S4 with values of 0.781 and 0.767, respectively. Similarly, \textit{LogisticRegression} demonstrated highly competitive behaviour, outperforming several online learning approaches and maintaining stable performance across all adaptation strategies.

A comparative analysis of S4 against the remaining strategies reveals that S4 provides robust and balanced performance improvements across multiple models. While S2 achieved the highest overall F1-score, S4 consistently outperformed the baseline strategy S1 and remained competitive with S3, S5, and S6. For example, \textit{ARF} improved from 0.639 under S1 to 0.675 under S4, while \textit{OnlineLogisticRegression} increased from 0.671 to 0.722. Likewise, \textit{RandomForest} improved substantially from 0.729 under S1 to 0.767 with S4, highlighting the ability of S4 to preserve both predictive precision and recall under changing data distributions. These results indicate that S4 offers a favorable trade-off between adaptation responsiveness and model stability.

The obtained F1-score patterns also demonstrate that S4 is particularly effective for models capable of exploiting adaptive updates and feature interactions. Ensemble learners and online linear methods benefited most from the strategy, whereas weaker-performing approaches such as \textit{HAT} exhibited comparatively limited gains. Although S2 produced slightly higher peak values for several models, S4 delivered more stable improvements across the broader set of classifiers, suggesting stronger generalization behaviour under drift conditions. Overall, the results confirm that S4 constitutes a reliable adaptation strategy for maintaining balanced classification performance in dynamic and non-stationary environments.

\begin{table}[ht]
\centering
\caption{Average F1-Score for Classification Tasks Across Strategies}
\label{tab:classification_f1}
\begin{tabular}{lcccccc}
\hline
\textbf{Model} & \textbf{S1} & \textbf{S2} & \textbf{S3} & \textbf{S4} & \textbf{S5} & \textbf{S6} \\
\hline
AMF & 0.593 & 0.750 & 0.641 & 0.541 & 0.553 & 0.553 \\
ARF & 0.639 & 0.718 & 0.654 & 0.675 & 0.654 & 0.654 \\
HAT & 0.372 & 0.675 & 0.517 & 0.459 & 0.517 & 0.517 \\
kNN & 0.618 & 0.741 & 0.608 & 0.651 & 0.608 & 0.608 \\
LogisticRegression & 0.738 & 0.743 & 0.752 & 0.756 & 0.752 & 0.752 \\
OnlineKNNClassifier & 0.639 & 0.756 & 0.644 & 0.655 & 0.644 & 0.644 \\
OnlineLogisticRegression & 0.671 & 0.735 & 0.718 & 0.722 & 0.718 & 0.718 \\
OnlineNaiveBayes & 0.625 & 0.638 & 0.602 & 0.626 & 0.602 & 0.602 \\
OnlineSoftmaxRegression & 0.663 & 0.716 & 0.713 & 0.688 & 0.713 & 0.713 \\
RandomForest & 0.729 & \textbf{0.781} & \textbf{0.755} & \textbf{0.767} & \textbf{0.755} & \textbf{0.755} \\
XGBoost & 0.673 & 0.765 & 0.723 & 0.709 & 0.723 & 0.723 \\
\hline
\end{tabular}
\end{table}
 
The results presented in Table~\ref{tab:classification_effort} provide a comparative analysis of the training effort required by the evaluated adaptation strategies across classification models. The reported values correspond to the average number of processed training samples used for adaptation and model updating. Overall, substantial differences can be observed between the evaluated strategies, highlighting the trade-off between predictive performance and computational adaptation cost in dynamic learning environments.

\begin{table}[ht]
\centering
\caption{Average Training Effort ($N_{train}$) for Classification Tasks Across Strategies}
\label{tab:classification_effort}
\begin{tabular}{lcccccc}
\hline
\textbf{Model} & \textbf{S1} & \textbf{S2} & \textbf{S3} & \textbf{S4} & \textbf{S5} & \textbf{S6} \\
\hline
AMF & \textbf{0.00} & 12493.33 & 1235.33 & \textbf{1223.88} & \textbf{353.00} & \textbf{353.00} \\
ARF & \textbf{0.00} & 12493.33 & 1216.33 & 5619.75 & 1216.33 & 1216.33 \\
HAT & \textbf{0.00} & 12493.33 & 1254.00 & 4807.08 & 1254.00 & 1254.00 \\
kNN & \textbf{0.00} & 12493.33 & 1253.67 & 4270.42 & 1253.67 & 1253.67 \\
LogisticRegression & \textbf{0.00} & 12493.33 & 1000.00 & 3583.33 & 1000.00 & 1000.00 \\
OnlineKNNClassifier & \textbf{0.00} & 12493.33 & 1235.00 & 4520.42 & 1235.00 & 1235.00 \\
OnlineLogisticRegression & \textbf{0.00} & 12493.33 & \textbf{666.67} & \textbf{3549.67} & \textbf{666.67} & \textbf{666.67} \\
OnlineNaiveBayes & \textbf{0.00} & 12493.33 & 1019.00 & 3887.42 & 1019.00 & 1019.00 \\
OnlineSoftmaxRegression & \textbf{0.00} & 12493.33 & 1315.67 & 4103.75 & 1315.67 & 1315.67 \\
RandomForest & \textbf{0.00} & 12493.33 & 1649.00 & 3932.67 & 1649.00 & 1649.00 \\
XGBoost & \textbf{0.00} & 12493.33 & 1451.33 & 3799.67 & 1451.33 & 1451.33 \\
\hline
\end{tabular}
\end{table}

A direct comparison of the strategies reveals that S2 incurred the highest training effort for all classifiers, with a constant average effort of 12493.33 samples. This behaviour indicates that S2 performs continuous or near-continuous adaptation, resulting in significantly higher computational requirements compared to the remaining approaches. In contrast, S1 required no additional adaptation effort, as reflected by the zero values observed across all classifiers, confirming its role as a static baseline without retraining or drift adaptation mechanisms.

The analysis further demonstrates that S4 achieved a substantially lower training effort than S2 while still maintaining strong predictive performance, as previously observed in the accuracy and F1-score evaluations. Although S4 required more adaptation effort than S3, S5, and S6, its computational cost remained considerably lower than that of S2 across all classifiers. For example, \textit{RandomForest} required an average effort of 3932.67 samples under S4 compared to 12493.33 under S2, corresponding to a reduction of more than 68\%. Similarly, \textit{OnlineLogisticRegression} reduced the effort from 12493.33 to 3549.67 samples while preserving competitive classification performance. These results suggest that S4 provides a more balanced compromise between adaptation effectiveness and computational efficiency.

Among the evaluated classifiers, \textit{OnlineLogisticRegression} consistently required the lowest adaptation effort for strategies S3, S5, and S6, with an average effort of 666.67 samples. Conversely, ensemble-based methods such as \textit{RandomForest} and boosting-based approaches generally required higher adaptation effort, reflecting the increased computational complexity associated with maintaining ensemble structures during drift adaptation. Nevertheless, despite the additional effort, these models also achieved superior predictive performance, indicating a favorable accuracy-effort trade-off.

Overall, the obtained results confirm that S4 constitutes an efficient adaptive strategy capable of significantly reducing computational overhead relative to highly reactive approaches such as S2, while still preserving strong classification performance under evolving data distributions.

\begin{table}[ht]
\centering
\caption{Average Number of Model Updates for Classification Tasks Across Strategies}
\label{tab:classification_updates}
\begin{tabular}{lcccccc}
\hline
\textbf{Model} & \textbf{S1} & \textbf{S2} & \textbf{S3} & \textbf{S4} & \textbf{S5} & \textbf{S6} \\
\hline
AMF & \textbf{0.00} & 11.33 & 1.67 & \textbf{1.00} & \textbf{1.00} & \textbf{1.00} \\
ARF & \textbf{0.00} & 11.33 & 1.33 & 3.00 & 1.33 & 1.33 \\
HAT & \textbf{0.00} & 11.33 & 2.00 & 2.75 & 2.00 & 2.00 \\
kNN & \textbf{0.00} & 11.33 & 2.00 & 2.21 & 2.00 & 2.00 \\
LogisticRegression & \textbf{0.00} & 11.33 & 1.00 & 1.79 & 1.00 & 1.00 \\
OnlineKNNClassifier & \textbf{0.00} & 11.33 & 1.67 & 2.33 & 1.67 & 1.67 \\
OnlineLogisticRegression & \textbf{0.00} & 11.33 & 0.67 & \textbf{1.83} & \textbf{0.67} & \textbf{0.67} \\
OnlineNaiveBayes & \textbf{0.00} & 11.33 & 1.33 & 1.96 & 1.33 & 1.33 \\
OnlineSoftmaxRegression & \textbf{0.00} & 11.33 & 1.67 & 2.13 & 1.67 & 1.67 \\
RandomForest & \textbf{0.00} & 11.33 & 2.00 & 2.08 & 2.00 & 2.00 \\
XGBoost & \textbf{0.00} & 11.33 & 2.00 & 1.96 & 2.00 & 2.00 \\
\hline
\end{tabular}
\end{table}

The results presented in Table~\ref{tab:classification_updates} provide a comparative evaluation of the average number of model updates performed by each adaptation strategy across the considered classification models. The reported values reflect the frequency of adaptation events triggered during learning and therefore provide an important indication of the responsiveness and computational behaviour of each strategy under dynamic data conditions.

A clear distinction can be observed between the evaluated strategies. As expected, S1 performed no updates across all classifiers, confirming its role as a static baseline without adaptive retraining capabilities. In contrast, S2 consistently produced the highest number of updates for every model, with an average of 11.33 updates. This result indicates that S2 relies on highly frequent adaptation mechanisms, leading to increased computational activity and training effort. Although this aggressive adaptation behaviour contributed to strong predictive performance in several cases, it also imposed substantially higher operational costs compared to the remaining strategies.

The comparative analysis further demonstrates that S4 achieved a considerably lower number of updates than S2 while maintaining competitive predictive performance in terms of both accuracy and F1-score. Across most classifiers, S4 required approximately two updates on average, representing a substantial reduction in adaptation frequency relative to S2. For example, \textit{RandomForest} reduced the average number of updates from 11.33 under S2 to only 2.08 under S4, while still achieving one of the highest predictive performances among all evaluated models. Similarly, \textit{LogisticRegression} required only 1.79 updates on average under S4 compared to 11.33 under S2. These findings indicate that S4 can effectively limit unnecessary adaptation events while still responding adequately to distributional changes.

Strategies S3, S5, and S6 generally exhibited the lowest update frequencies among the adaptive approaches, often requiring between one and two updates per model. In particular, \textit{OnlineLogisticRegression} achieved the minimum adaptation frequency under S3, S5, and S6, requiring only 0.67 updates on average. While these strategies provide strong computational efficiency, the predictive results obtained previously suggest that their lower update frequency may limit responsiveness to certain drift scenarios compared to S4 and S2.

Overall, the obtained results confirm that S4 offers a balanced trade-off between adaptation responsiveness and computational efficiency. Compared with the highly reactive behaviour of S2, S4 substantially reduced the number of model updates while preserving strong classification performance, demonstrating its suitability for dynamic and resource-constrained learning environments.

Figure \ref{fig:all_performance_grid} shows a conglomerate of bar charts that provides the performance and adaptation analysis of all classification datasets per strategy, where the total training effort graph shown in \ref{fig:All_total_training_effort}, measured as the cumulative number of samples processed during model fitting, provides a direct estimate of the computational cost associated with each strategy. The results reveal substantial differences in computational demand between the considered approaches. In particular, S2 incurred by far the highest training effort, exceeding 3.3 million processed samples, which confirms its highly reactive adaptation behaviour and frequent retraining operations. Although S2 achieved strong predictive performance in the accuracy and F1-score evaluations, the obtained results indicate that these gains come at a significant computational cost.

In contrast, S4 substantially reduced the overall training effort to approximately 1.0 million samples while still maintaining competitive predictive performance across most classifiers. This represents a reduction of nearly 70\% compared to S2, demonstrating that S4 achieves a more favorable balance between adaptation effectiveness and computational efficiency. Strategies S3, S5, and S6 exhibited the lowest non-zero effort values, each requiring approximately 0.3 million processed samples, indicating considerably lower adaptation activity and improved computational efficiency.

As expected, S1 required virtually no training effort because it operates as a static baseline without adaptive retraining. Overall, the results demonstrate that while highly reactive strategies such as S2 maximize adaptation frequency, they also introduce substantial computational overhead. Conversely, S4 provides a more efficient compromise by significantly reducing training effort while preserving strong predictive performance under evolving data distributions.

\begin{figure}[htbp]
    \centering
    \subfloat[Total Training Effort per strategy]{\includegraphics[width=0.45\textwidth]{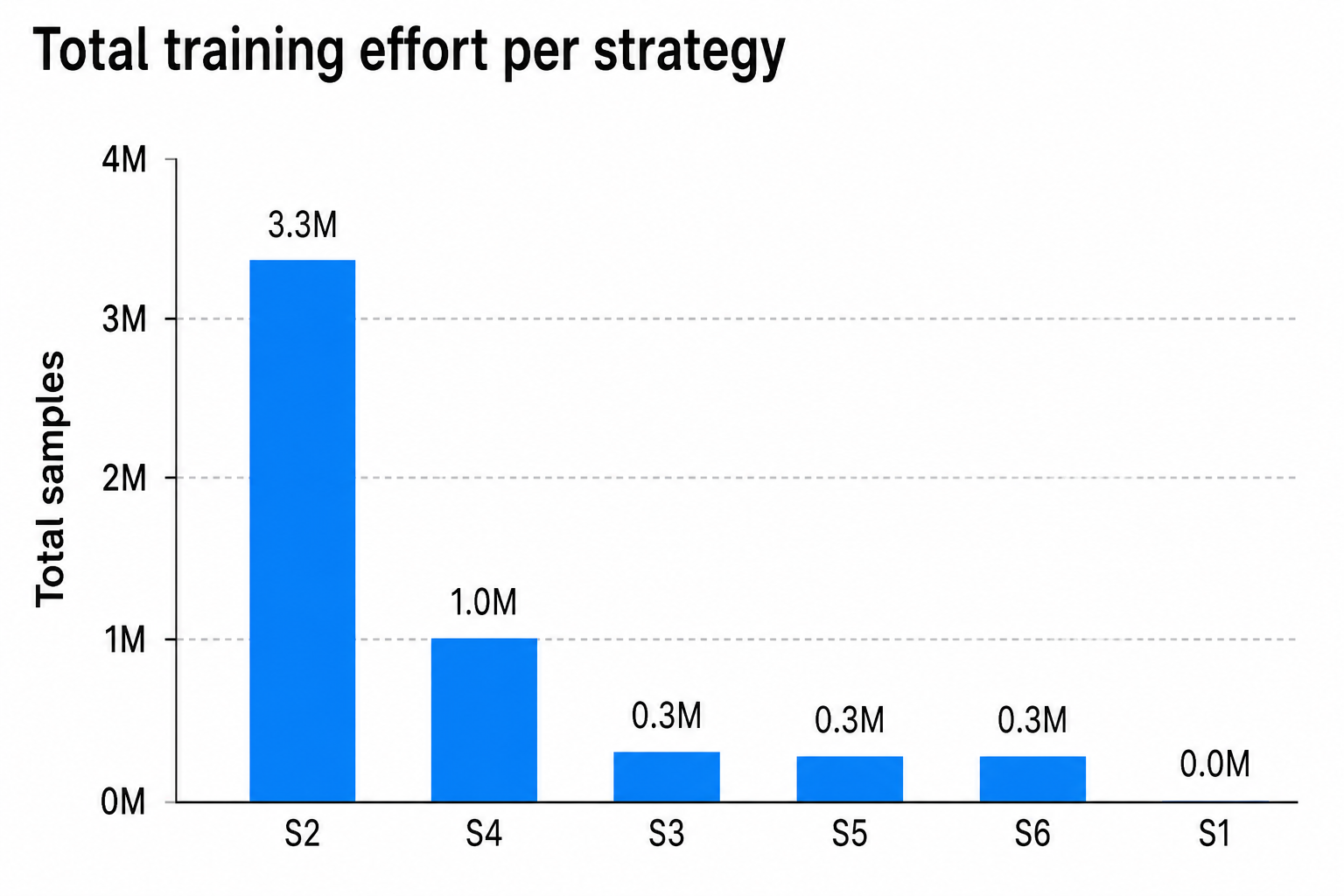}
    \label{fig:All_total_training_effort}}
    \hfill
    \subfloat[Total Model updates]{\includegraphics[width=0.45\textwidth]{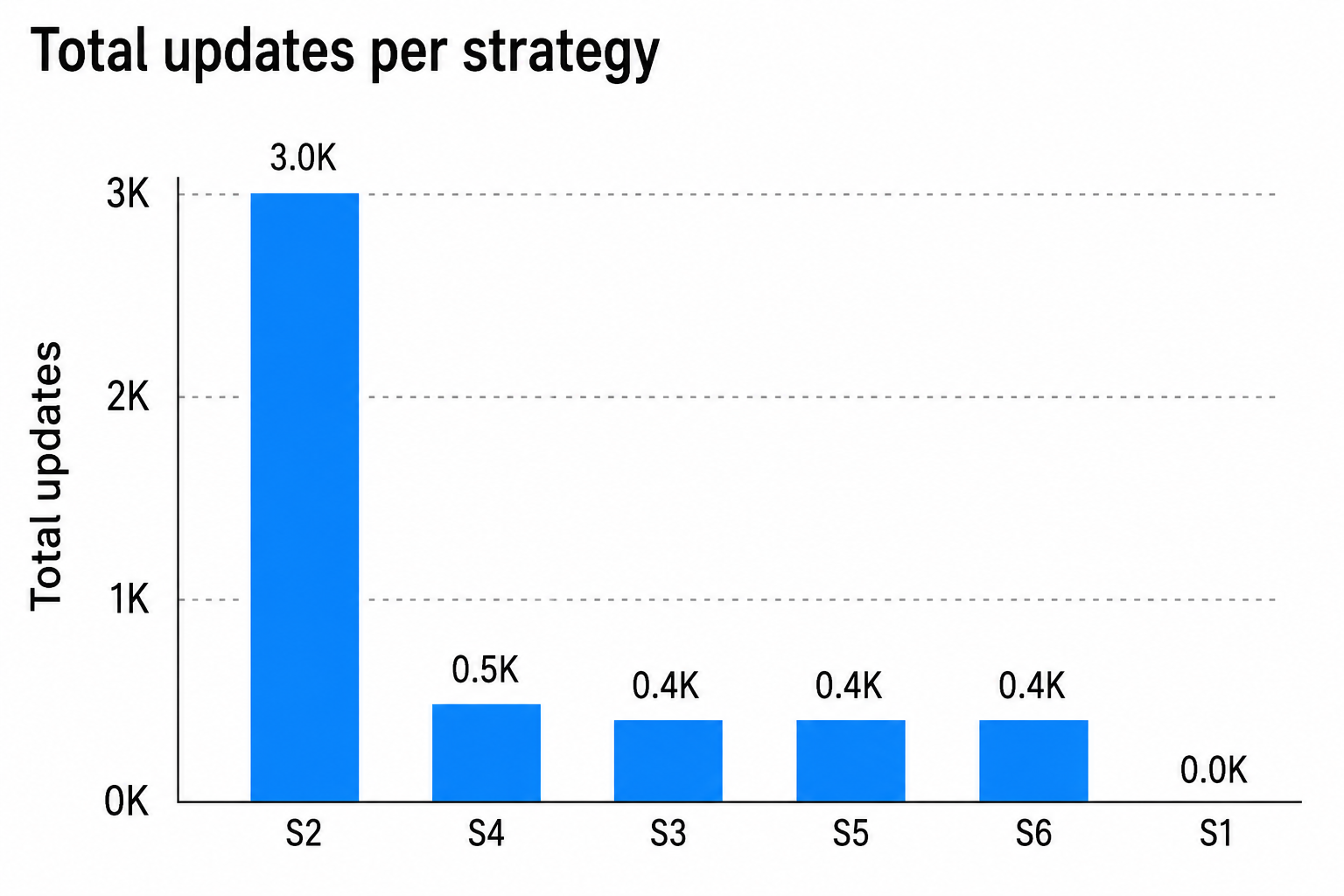}
    \label{fig:all_model_updates}}    
    \hfill
    \subfloat[Total Drift Triggers per strategy]{\includegraphics[width=0.45\textwidth]{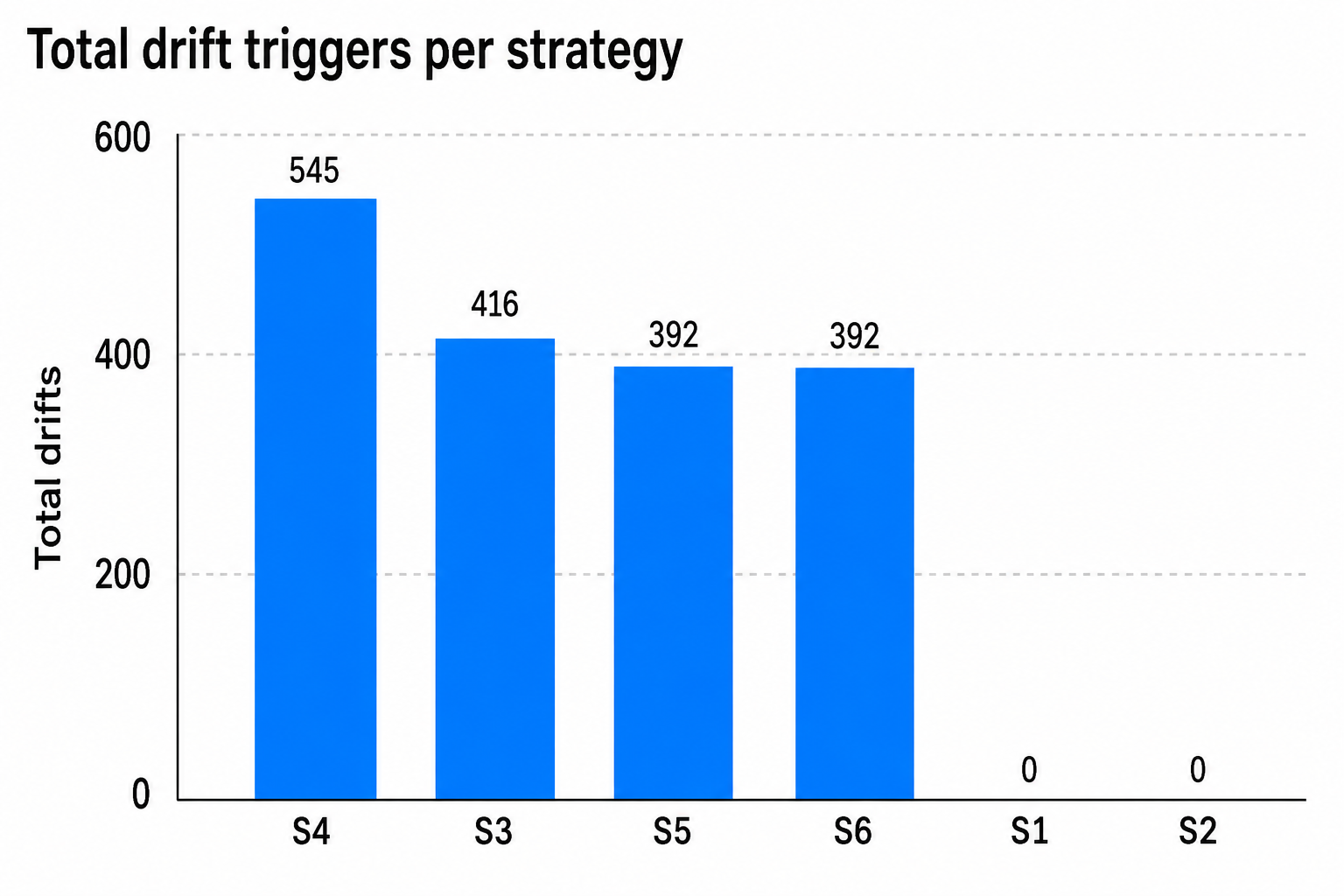}\label{ fig:acc_over_time_LR_Adult}
     \label{fig:all_drift_triggers}}
    \caption{Performance and Adaptation Effort Analysis for All Classification Datasets across Strategies S1-S6.}
    \label{fig:all_performance_grid}
\end{figure}

The total number of model updates graph shown in \ref{fig:all_drift_triggers} further clarifies the operational differences between the strategies. The results reveal substantial differences in adaptation frequency between the considered approaches, highlighting the trade-off between responsiveness to controlled distribution shift and computational overhead.

Among all strategies, S2 produced by far the highest number of updates, exceeding 3,000 total adaptation events. This behaviour confirms that S2 adopts a highly reactive adaptation mechanism that continuously retrains or updates models in response to changing data distributions. While this aggressive update behaviour contributed to strong predictive performance in several experiments, it also introduced significantly higher computational complexity and operational cost.

In contrast, S4 substantially reduced the total number of updates to approximately 500K while still maintaining competitive accuracy and F1-score performance. Compared with S2, this corresponds to a reduction of more than 80\% in update frequency, demonstrating that S4 can effectively limit unnecessary adaptation operations while preserving robust predictive capabilities. Strategies S3, S5, and S6 exhibited even lower update counts, each remaining near 400K total updates, indicating a more conservative adaptation behaviour with reduced computational demand.

As expected, S1 generated virtually no updates because it represents the static baseline without adaptive retraining. Overall, the results demonstrate that S4 provides a favorable compromise between adaptation responsiveness and computational efficiency, significantly reducing update frequency relative to highly reactive strategies such as S2 while still maintaining strong classification performance under evolving data distributions.

The last graph containing the total number of drift triggers seen in \ref{fig:all_drift_triggers} illustrates the sensitivity of the detection mechanisms embedded in the adaptive methods. The results reveal substantial differences in drift sensitivity and adaptation behaviour among the considered approaches. In particular, S4 produced the highest number of drift triggers, detecting a total of 545 drift events, followed by S3 with 416 triggers, while S5 and S6 each detected 392 events. In contrast, S1 and S2 produced no drift triggers because these strategies do not rely on explicit drift detection mechanisms.

The high number of detected drift events under S4 indicates that the strategy is highly responsive to distributional changes in the data stream. This behaviour aligns with the strong predictive performance previously observed for S4 in terms of both average accuracy and F1-score. By actively identifying changes in the underlying data distribution, S4 can initiate adaptive retraining processes more effectively, thereby improving model robustness under non-stationary environments. Importantly, despite detecting more drift events than the remaining adaptive strategies, S4 still maintained substantially lower training effort and update frequency compared with the highly reactive S2 approach.

A comparative analysis further shows that S3, S5, and S6 adopt more conservative drift detection behaviour, generating fewer triggers overall. While this reduces computational overhead and adaptation frequency, the lower number of detected drift events may limit responsiveness to abrupt or recurring distributional changes. Consequently, these strategies generally achieved slightly lower predictive performance compared with S4 in previous evaluations.

Overall, the results demonstrate that S4 achieves an effective balance between drift sensitivity, predictive performance, and computational efficiency. The strategy actively detects evolving data distributions while avoiding the excessive retraining behaviour observed in S2, making it a robust and practical solution for adaptive learning in dynamic environments.

\subsubsection{Adult}

The results displayed in table \ref{tab:adult_results} presents a detailed comparison of the evaluated adaptation strategies on the Adult dataset across multiple classifiers, considering predictive performance, adaptation frequency, and computational effort. Overall, the results demonstrate that adaptive strategies consistently improved classification performance compared with the static baseline S1, although the magnitude of improvement and associated computational cost varied considerably across classifiers and strategies.

\begin{longtable}{@{}llccrr@{}}
\caption{Detailed performance comparison on the Adult dataset.}
\label{tab:adult_results}\\
\toprule
\textbf{Classifier} & \textbf{Strat.} & \textbf{Avg. F1} & \textbf{Avg. Acc.} & \textbf{Updates} & \textbf{Effort (Samples)} \\
\midrule
\endfirsthead

\caption[]{Detailed performance comparison on the Adult dataset (continued).}\\
\toprule
\textbf{Classifier} & \textbf{Strat.} & \textbf{Avg. F1} & \textbf{Avg. Acc.} & \textbf{Updates} & \textbf{Effort (Samples)} \\
\midrule
\endhead

\midrule
\multicolumn{6}{r}{\textit{Continued on next page}}\\
\endfoot

\bottomrule
\endlastfoot

\textbf{AMF} & S1 & 0.754 & 0.752 & 0 & 0 \\
            & S2 & 0.823 & 0.833 & 14 & 20500 \\
            & S3 & 0.818 & 0.827 & 3 & 3000 \\
\midrule

\textbf{ARF} & S1 & 0.705 & 0.753 & 0 & 0 \\
            & S2 & 0.780 & 0.811 & 14 & 20500 \\
            & S3 & 0.756 & 0.796 & 3 & 3000 \\
            & S4 & 0.830 & 0.839 & 4 & 5100 \\
            & S5 & 0.756 & 0.796 & 3 & 3000 \\
            & S6 & 0.756 & 0.796 & 3 & 3000 \\
\midrule

\textbf{HAT} & S1 & 0.668 & 0.700 & 0 & 0 \\
            & S2 & 0.741 & 0.772 & 14 & 20500 \\
            & S3 & 0.692 & 0.724 & 3 & 3000 \\
            & S4 & 0.721 & 0.757 & 6 & 13500 \\
            & S5 & 0.692 & 0.724 & 3 & 3000 \\
            & S6 & 0.692 & 0.724 & 3 & 3000 \\
\midrule

\textbf{kNN} & S1 & 0.737 & 0.743 & 0 & 0 \\
            & S2 & 0.812 & 0.822 & 14 & 20500 \\
            & S3 & 0.801 & 0.810 & 3 & 3000 \\
            & S4 & 0.813 & 0.820 & 5 & 11700 \\
            & S5 & 0.801 & 0.810 & 3 & 3000 \\
            & S6 & 0.801 & 0.810 & 3 & 3000 \\
\midrule

\textbf{LogisticRegression} & S1 & 0.823 & 0.820 & 0 & 0 \\
            & S2 & 0.820 & 0.817 & 14 & 20500 \\
            & S3 & 0.823 & 0.821 & 1 & 1000 \\
            & S4 & 0.824 & 0.822 & 4 & 7700 \\
            & S5 & 0.823 & 0.821 & 1 & 1000 \\
            & S6 & 0.823 & 0.821 & 1 & 1000 \\
\midrule

\textbf{OnlineKNNClassifier} & S1 & 0.734 & 0.744 & 0 & 0 \\
            & S2 & 0.782 & 0.794 & 14 & 20500 \\
            & S3 & 0.778 & 0.790 & 2 & 2000 \\
            & S4 & 0.779 & 0.791 & 5 & 11700 \\
            & S5 & 0.778 & 0.790 & 2 & 2000 \\
            & S6 & 0.778 & 0.790 & 2 & 2000 \\
\midrule

\textbf{OnlineLogisticRegression} & S1 & 0.823 & 0.820 & 0 & 0 \\
            & S2 & 0.820 & 0.816 & 14 & 20500 \\
            & S3 & 0.823 & 0.821 & 1 & 1000 \\
            & S4 & 0.824 & 0.822 & 4 & 7700 \\
            & S5 & 0.823 & 0.821 & 1 & 1000 \\
            & S6 & 0.823 & 0.821 & 1 & 1000 \\
\midrule

\textbf{OnlineNaiveBayes} & S1 & 0.770 & 0.776 & 0 & 0 \\
        & S2 & 0.782 & 0.788 & 14 & 20500 \\
        & S3 & 0.784 & 0.792 & 2 & 2000 \\
        & S4 & 0.785 & 0.793 & 5 & 9700 \\
        & S5 & 0.784 & 0.792 & 2 & 2000 \\
        & S6 & 0.784 & 0.792 & 2 & 2000 \\
\midrule

\textbf{OnlineSoftmaxRegression} & S1 & 0.823 & 0.820 & 0 & 0 \\
        & S2 & 0.819 & 0.815 & 14 & 20500 \\
        & S3 & 0.823 & 0.821 & 1 & 1000 \\
        & S4 & 0.824 & 0.822 & 4 & 7700 \\
        & S5 & 0.823 & 0.821 & 1 & 1000 \\
        & S6 & 0.823 & 0.821 & 1 & 1000 \\
\midrule

\textbf{RandomForest} & S1 & 0.797 & 0.794 & 0 & 0 \\
        & S2 & 0.844 & 0.841 & 14 & 20500 \\
        & S3 & 0.843 & 0.839 & 3 & 3000 \\
        & S4 & 0.847 & 0.844 & 4 & 8700 \\
        & S5 & 0.843 & 0.839 & 3 & 3000 \\
        & S6 & 0.843 & 0.839 & 3 & 3000 \\
\midrule

\textbf{XGBoost} & S1 & 0.804 & 0.800 & 0 & 0 \\
        & S2 & 0.842 & 0.838 & 14 & 20500 \\
        & S3 & 0.838 & 0.834 & 3 & 3000 \\
        & S4 & 0.845 & 0.841 & 4 & 7500 \\
        & S5 & 0.838 & 0.834 & 3 & 3000 \\
        & S6 & 0.838 & 0.834 & 3 & 3000 \\

\end{longtable}
\FloatBarrier

The static baseline strategy S1 consistently required no updates or additional training effort, as expected, but generally produced the lowest predictive performance for most classifiers. This behaviour is particularly evident for adaptive tree-based approaches such as \textit{HAT}, \textit{ARF}, and \textit{AMF}, where both F1-score and accuracy improved substantially once adaptive retraining mechanisms were introduced. Nevertheless, linear models such as \textit{LogisticRegression}, \textit{OnlineLogisticRegression}, and \textit{OnlineSoftmaxRegression} exhibited relatively strong baseline performance under S1, suggesting higher inherent robustness to moderate distributional changes in the Adult dataset.

Strategy S2 achieved strong predictive performance across nearly all classifiers, often producing substantial improvements relative to S1. For instance, \textit{RandomForest} improved from an average accuracy of 0.794 under S1 to 0.841 under S2, while \textit{AMF} increased from 0.752 to 0.833. However, these gains were accompanied by the highest adaptation cost, requiring 14 updates and a training effort of 20,500 samples for every classifier. This indicates that S2 adopts an aggressive adaptation mechanism with continuous retraining behaviour, which, although effective in maximizing predictive performance, introduces considerable computational overhead.

Strategies S3, S5, and S6 demonstrated highly similar behaviour across all classifiers, consistently achieving competitive predictive performance with substantially lower computational requirements than S2. In most cases, these strategies required only one to three updates and training efforts between 1,000 and 3,000 samples. For example, \textit{LogisticRegression}, \textit{OnlineLogisticRegression}, and \textit{OnlineSoftmaxRegression} maintained accuracies above 0.821 while requiring only a single update and 1,000 training samples. These findings indicate that moderate adaptation mechanisms can preserve strong predictive performance while significantly reducing computational cost.

Among all evaluated strategies, S4 consistently provided the best balance between predictive performance and computational efficiency. In several classifiers, S4 achieved the highest overall performance, including \textit{RandomForest} (F1-score: 0.847, accuracy: 0.844), \textit{XGBoost} (F1-score: 0.845, accuracy: 0.841), and \textit{ARF} (F1-score: 0.830, accuracy: 0.839). Importantly, these gains were obtained with considerably lower adaptation effort than S2. For instance, \textit{RandomForest} under S4 required only four updates and 8,700 training samples compared with 14 updates and 20,500 samples under S2. Similar trends can be observed for \textit{XGBoost} and \textit{ARF}, demonstrating that S4 effectively improves predictive performance while controlling computational overhead.

The results also reveal that ensemble-based classifiers such as \textit{RandomForest}, \textit{XGBoost}, and \textit{ARF} benefited most from adaptive strategies, particularly S4 and S2, achieving the highest overall predictive metrics in the study. Conversely, simpler online probabilistic approaches such as \textit{OnlineNaiveBayes} exhibited more modest gains across strategies, suggesting limited sensitivity to the considered adaptation mechanisms. Linear online learners maintained highly stable behaviour across all strategies, indicating strong robustness under the relatively gradual drift characteristics of the Adult dataset.

Figure \ref{fig:adults_grid} displays the timelines that illustratinge the evolution of batch-wise accuracy for the Adult dataset across a selection of four classifiers showing the different adaptation strategies, where the vertical dashed lines indicate the imposed drift boundaries, corresponding to controlled cluster-induced distributional shifts. The ARF graph seen in Figure \ref{fig:acc_over_time_ARF_Adult} shows that before the drift boundaries, most adaptive strategies achieved high predictive performance, with S2, S4, and S6 reaching accuracies close to 0.98. After the major drift event around batch 9, all strategies experienced performance degradation, highlighting the impact of abrupt distributional changes on the data stream.

Among the evaluated approaches, S2 demonstrated the most stable recovery after drift, maintaining consistently higher accuracy during the later batches. S4 achieved strong performance before the drift event but exhibited a sharp temporary decline immediately afterward, indicating higher sensitivity to abrupt changes. In contrast, the static baseline S1 maintained lower but relatively stable performance throughout the stream, confirming the importance of adaptive retraining mechanisms under controlled distribution shift conditions.

The HAT graph seen in Figure \ref{fig:acc_over_time_HAT_Adult} displays that before the major drift event, most adaptive strategies achieved very high predictive performance, with S2, S3, S4, and S5 reaching accuracies close to 0.98, demonstrating strong adaptation capabilities during stable stream conditions.

Following the drift boundary around batch 9, all strategies experienced a substantial decrease in accuracy, highlighting the strong impact of abrupt distributional changes on the HAT classifier. Among the evaluated approaches, S2 demonstrated the most effective recovery behaviour, maintaining accuracies above 0.70 during the later stages of the stream. In contrast, S1 and S4 suffered a severe and prolonged degradation after the drift event, with accuracy values dropping below 0.40 for several batches. S6 showed moderate recovery toward the end of the stream but remained less stable than S2.

The kNN graph seen in Figure \ref{fig:acc_over_time_KNN_Adult} shows that before the major drift event, adaptive approaches such as S2, S4, and S6 achieved very high predictive performance, reaching accuracies close to 0.98, while the static baseline S1 maintained noticeably lower and less stable accuracy throughout the stream.

After the drift boundary around batch 9, all strategies experienced a significant reduction in performance, reflecting the impact of abrupt distributional changes on the kNN classifier. Among the evaluated approaches, S2 and S4 demonstrated the strongest recovery behaviour, gradually restoring accuracy to values above 0.70 during the later stages of the stream. In contrast, the static baseline S1 remained consistently below the adaptive strategies after the drift event, confirming the limitations of non-adaptive learning under evolving data conditions. S6 exhibited moderate recovery toward the end of the stream but showed greater instability compared with S2 and S4.

The graph for the Logistic Regression seen in Figure \ref{fig:acc_over_time_LG_Adult} shows that during the stable phases of the stream, most adaptive strategies achieved very high predictive performance, maintaining accuracies close to 0.98 before the major drift event. In contrast, the static baseline S1 exhibited lower stability after drift, while S2 showed a temporary performance degradation during the early batches before rapidly recovering.

Following the drift boundary around batch 9, all strategies experienced a noticeable decline in accuracy, reflecting the impact of abrupt distributional changes on the classifier. Among the evaluated approaches, S4 and S6 demonstrated the strongest recovery behaviour during the later stages of the stream, progressively restoring accuracy to values approaching 0.80 by the end of the evaluation period. S2 also maintained relatively stable post-drift performance, although with slightly lower final accuracy compared to S4. Conversely, the static baseline S1 experienced the largest sustained degradation after drift, confirming the limitations of non-adaptive learning in dynamic environments.

\begin{figure}[htbp]
    \centering
    \subfloat[ARF: Accuracy Over Time]{\includegraphics[width=0.45\textwidth]{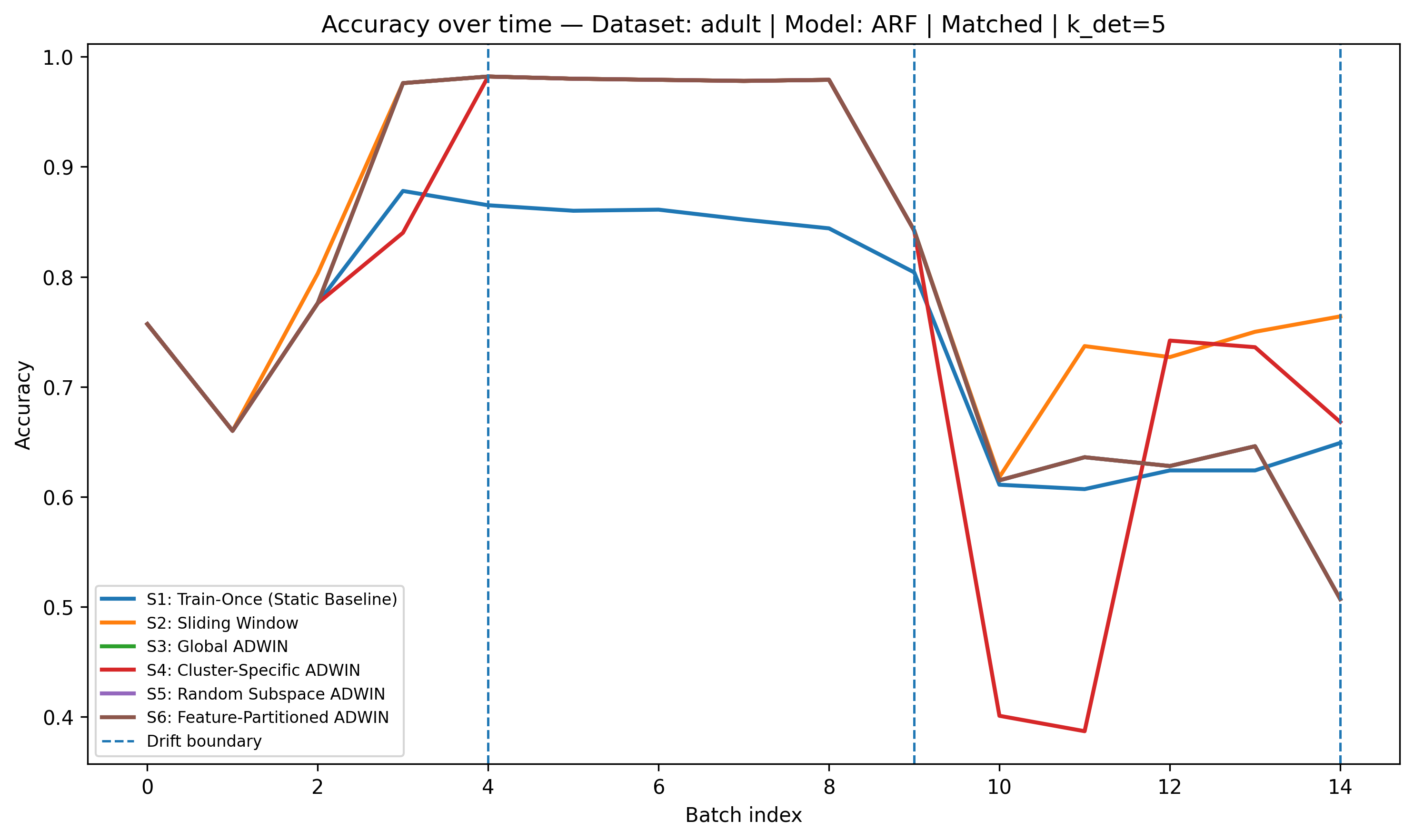}
    \label{fig:acc_over_time_ARF_Adult}}
    \hfill
    \subfloat[HAT: Accuracy Over Time]{\includegraphics[width=0.45\textwidth]{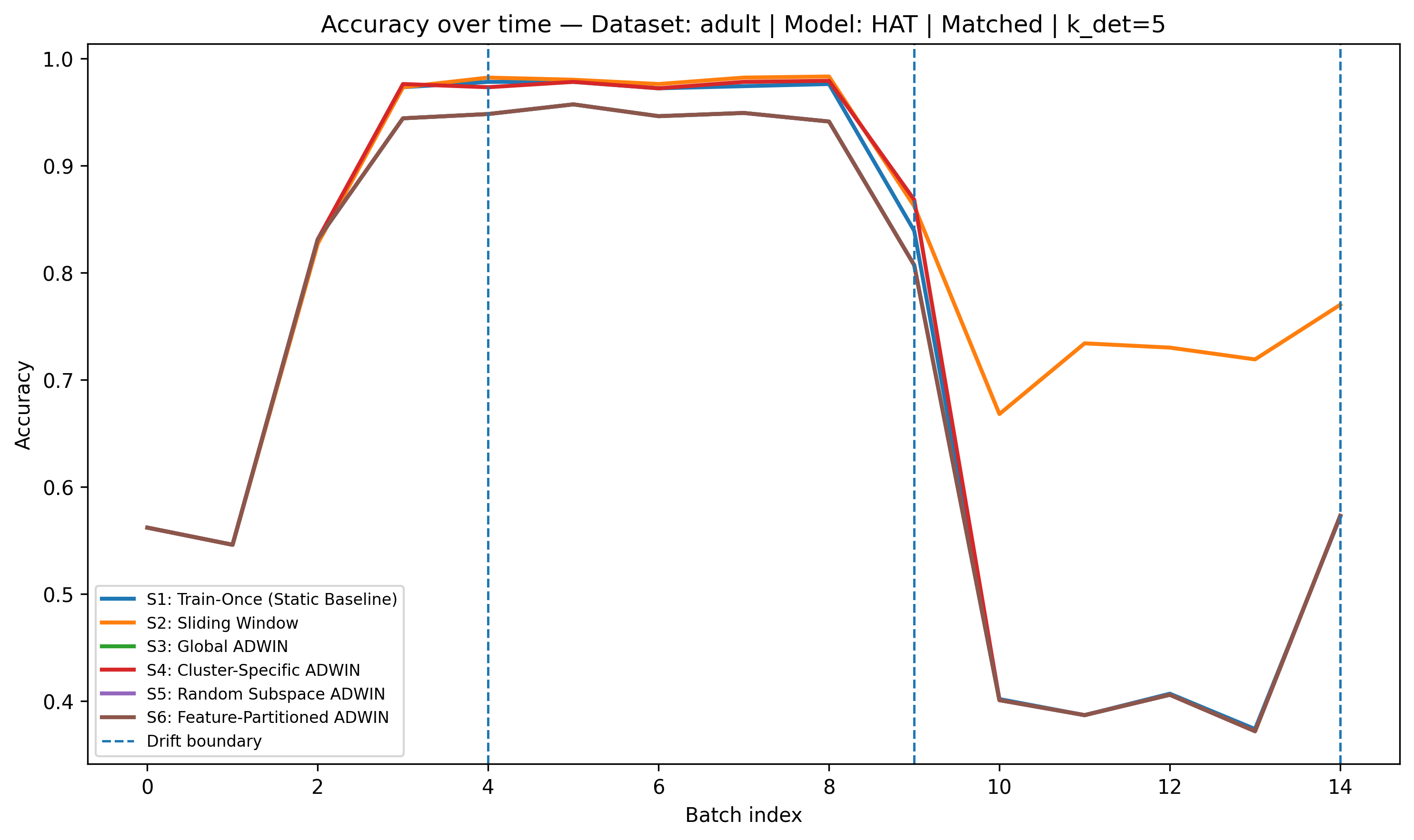}
    \label{fig:acc_over_time_HAT_Adult}}
    \hfill
    \subfloat[kNN: Accuracy Over Time]{\includegraphics[width=0.45\textwidth]{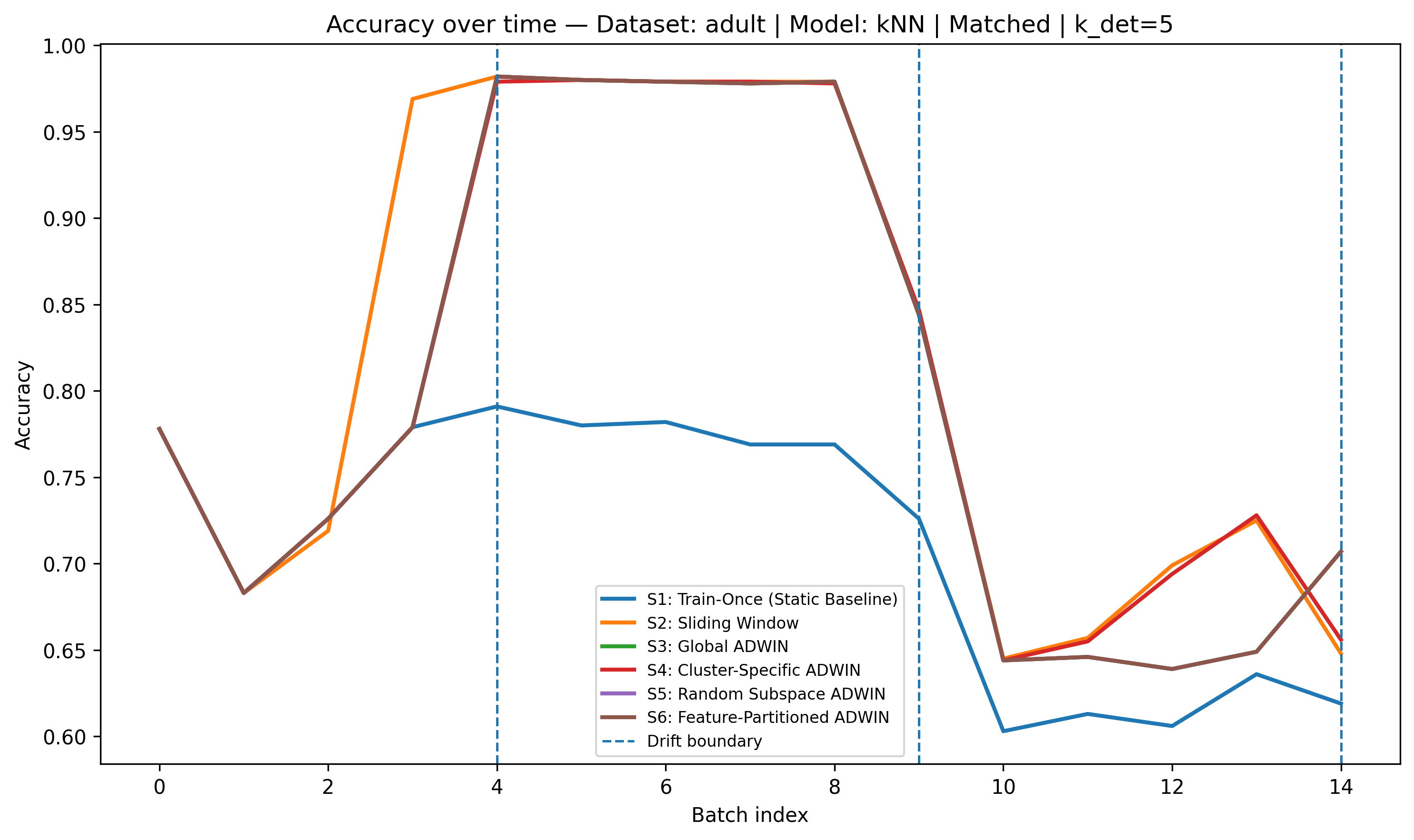}
    \label{fig:acc_over_time_KNN_Adult}}
    \hfill
    \subfloat[Logistic Regression: Accuracy Over Time]{\includegraphics[width=0.45\textwidth]{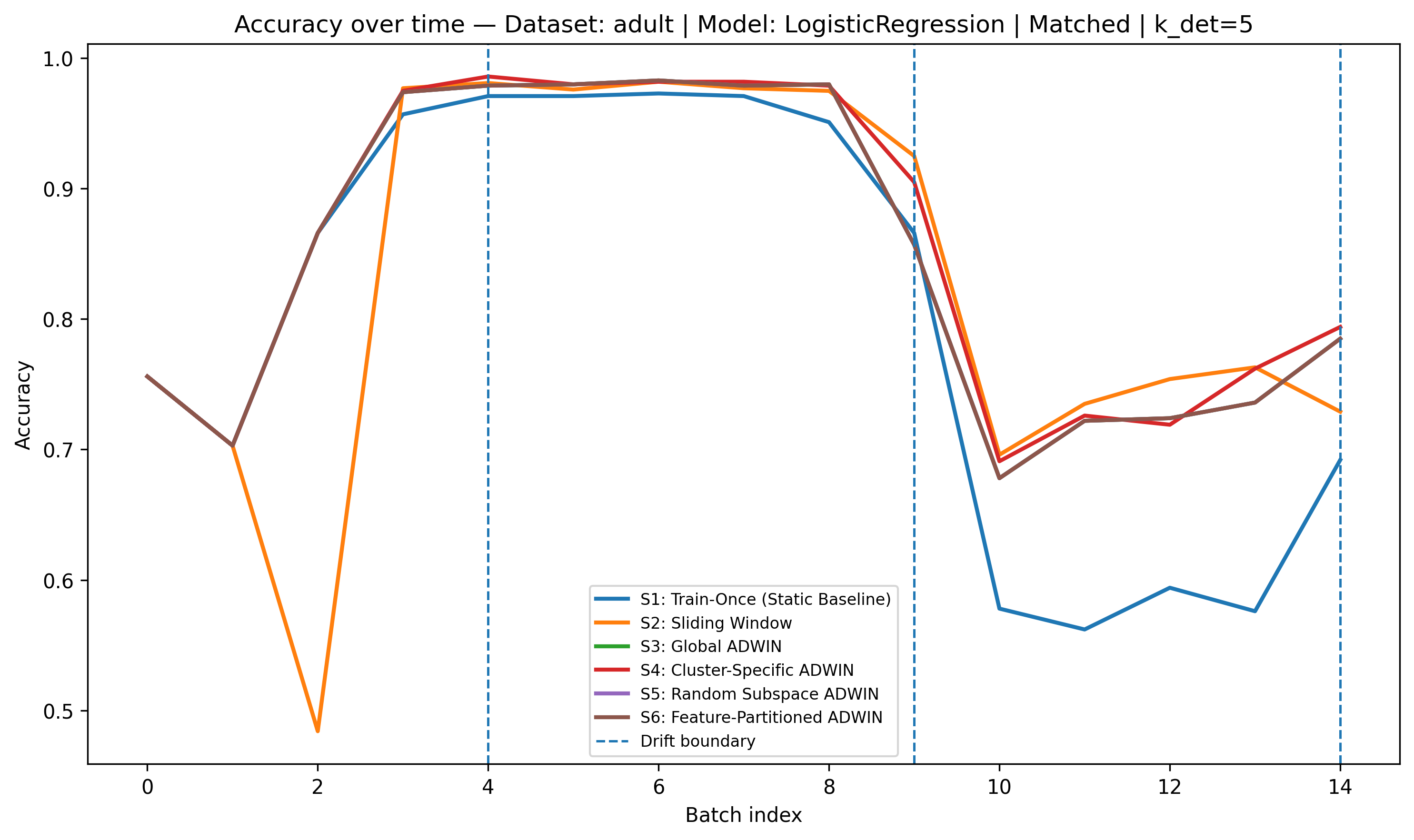}
    \label{fig:acc_over_time_LG_Adult}}
    \caption{Performance and Adaptation Effort Analysis for the Adult Dataset across Strategies S1-S6.}
    \label{fig:adults_grid}
\end{figure}

The timeline analyses on the Adult dataset demonstrate that adaptive retraining strategies significantly improve classifier robustness under controlled distribution shift compared with the static baseline S1, which consistently suffered substantial performance degradation after drift events. Across all evaluated classifiers, abrupt distributional changes caused noticeable drops in accuracy, highlighting the challenges of maintaining predictive stability in dynamic streaming environments.

Among the evaluated approaches, S2 generally achieved the most stable post-drift recovery, maintaining strong predictive performance across multiple classifiers, although at the cost of high update frequency and training effort. In contrast, S3, S5, and S6 provided lower computational overhead but exhibited more moderate recovery behaviour. Overall, S4 offered the best balance between adaptation responsiveness and computational efficiency, consistently achieving strong predictive stability and competitive recovery performance without the excessive retraining requirements observed in S2.
\FloatBarrier

\subsubsection{Wine Quality}

The results on the Wine Quality dataset displayed in table \ref{tab:wine_quality_results} reveal substantial variability in classifier behaviour and adaptation effectiveness on the Wine dataset. Overall, adaptive retraining strategies generally improved predictive performance relative to the static baseline S1, although the magnitude of these improvements varied considerably across classifiers. Compared with the Adult dataset, the Wine dataset exhibited lower overall predictive performance and greater sensitivity to adaptation strategy selection, suggesting a more challenging and unstable drift scenario.

\begin{longtable}{@{}llccrr@{}}
\caption{Detailed performance comparison on the Wine dataset.}
\label{tab:wine_quality_results}\\
\toprule
\textbf{Classifier} & \textbf{Strat.} & \textbf{Avg. F1} & \textbf{Avg. Acc.} & \textbf{Updates} & \textbf{Effort (Samples)} \\
\midrule
\endfirsthead

\caption[]{Detailed performance comparison on the Wine dataset (continued).}\\
\toprule
\textbf{Classifier} & \textbf{Strat.} & \textbf{Avg. F1} & \textbf{Avg. Acc.} & \textbf{Updates} & \textbf{Effort (Samples)} \\
\midrule
\endhead

\midrule
\multicolumn{6}{r}{\textit{Continued on next page}}\\
\endfoot

\bottomrule
\endlastfoot

\textbf{AMF} & S1 & 0.449 & 0.494 & 0 & 0 \\
             & S2 & 0.588 & 0.607 & 10 & 13947 \\
             & S3 & 0.504 & 0.520 & 1 & 649 \\
             & S4 & 0.557 & 0.579 & 2 & 2596 \\
             & S5 & 0.504 & 0.520 & 1 & 649 \\
             & S6 & 0.504 & 0.520 & 1 & 649 \\
\midrule
\textbf{ARF} & S1 & 0.384 & 0.452 & 0 & 0 \\
             & S2 & 0.454 & 0.505 & 10 & 13947 \\
             & S3 & 0.429 & 0.478 & 1 & 649 \\
             & S4 & 0.457 & 0.498 & 1 & 1298 \\
             & S5 & 0.429 & 0.478 & 1 & 649 \\
             & S6 & 0.429 & 0.478 & 1 & 649 \\
\midrule
\textbf{HAT} & S1 & 0.195 & 0.334 & 0 & 0 \\
             & S2 & 0.379 & 0.472 & 10 & 13947 \\
             & S3 & 0.338 & 0.446 & 1 & 649 \\
             & S4 & 0.331 & 0.438 & 2 & 2596 \\
             & S5 & 0.338 & 0.446 & 1 & 649 \\
             & S6 & 0.338 & 0.446 & 1 & 649 \\
\midrule
\textbf{kNN} & S1 & 0.429 & 0.493 & 0 & 0 \\
             & S2 & 0.526 & 0.564 & 10 & 13947 \\
             & S3 & 0.447 & 0.485 & 1 & 649 \\
             & S4 & 0.429 & 0.493 & 0 & 0 \\
             & S5 & 0.447 & 0.485 & 1 & 649 \\
             & S6 & 0.447 & 0.485 & 1 & 649 \\
\midrule
\textbf{LogisticRegression} & S1 & 0.464 & 0.506 & 0 & 0 \\
                            & S2 & 0.461 & 0.517 & 10 & 13947 \\
                            & S3 & 0.464 & 0.506 & 0 & 0 \\
                            & S4 & 0.464 & 0.506 & 0 & 0 \\
                            & S5 & 0.464 & 0.506 & 0 & 0 \\
                            & S6 & 0.464 & 0.506 & 0 & 0 \\
\midrule
\textbf{OnlineKNNClassifier} & S1 & 0.469 & 0.527 & 0 & 0 \\
                             & S2 & 0.536 & 0.558 & 10 & 13947 \\
                             & S3 & 0.540 & 0.566 & 1 & 649 \\
                             & S4 & 0.469 & 0.527 & 0 & 0 \\
                             & S5 & 0.540 & 0.566 & 1 & 649 \\
                             & S6 & 0.540 & 0.566 & 1 & 649 \\
\midrule
\textbf{OnlineLogisticRegression} & S1 & 0.435 & 0.531 & 0 & 0 \\
                                  & S2 & 0.470 & 0.552 & 10 & 13947 \\
                                  & S3 & 0.435 & 0.531 & 0 & 0 \\
                                  & S4 & 0.435 & 0.531 & 0 & 0 \\
                                  & S5 & 0.435 & 0.531 & 0 & 0 \\
                                  & S6 & 0.435 & 0.531 & 0 & 0 \\
\midrule
\textbf{OnlineNaiveBayes} & S1 & 0.455 & 0.481 & 0 & 0 \\
                          & S2 & 0.417 & 0.483 & 10 & 13947 \\
                          & S3 & 0.455 & 0.481 & 0 & 0 \\
                          & S4 & 0.455 & 0.481 & 0 & 0 \\
                          & S5 & 0.455 & 0.481 & 0 & 0 \\
                          & S6 & 0.455 & 0.481 & 0 & 0 \\
\midrule
\textbf{OnlineSoftmaxRegression} & S1 & 0.435 & 0.486 & 0 & 0 \\
                                 & S2 & 0.430 & 0.468 & 10 & 13947 \\
                                 & S3 & 0.445 & 0.483 & 3 & 1947 \\
                                 & S4 & 0.435 & 0.486 & 0 & 0 \\
                                 & S5 & 0.445 & 0.483 & 3 & 1947 \\
                                 & S6 & 0.445 & 0.483 & 3 & 1947 \\
\midrule
\textbf{RandomForest} & S1 & 0.479 & 0.550 & 0 & 0 \\
                      & S2 & 0.538 & 0.598 & 10 & 13947 \\
                      & S3 & 0.481 & 0.524 & 3 & 1947 \\
                      & S4 & 0.532 & 0.576 & 1 & 1298 \\
                      & S5 & 0.481 & 0.524 & 3 & 1947 \\
                      & S6 & 0.481 & 0.524 & 3 & 1947 \\
\midrule
\textbf{XGBoost} & S1 & 0.483 & 0.537 & 0 & 0 \\
                 & S2 & 0.542 & 0.598 & 10 & 13947 \\
                 & S3 & 0.544 & 0.583 & 2 & 1298 \\
                 & S4 & 0.483 & 0.537 & 0 & 0 \\
                 & S5 & 0.544 & 0.583 & 2 & 1298 \\
                 & S6 & 0.544 & 0.583 & 2 & 1298 \\

\end{longtable}

For the \textit{AMF} classifier, adaptive strategies significantly improved both F1-score and accuracy compared with the static baseline. In particular, S2 achieved the highest predictive performance, increasing accuracy from 0.494 to 0.607, although this improvement required the highest adaptation effort and update frequency. S4 also provided notable gains while requiring substantially fewer updates and training samples than S2, indicating a more efficient balance between adaptation and computational cost.

The \textit{ARF} classifier exhibited similar behaviour, with adaptive strategies consistently outperforming S1. Among the evaluated approaches, S4 achieved the best trade-off, slightly outperforming S3, S5, and S6 while requiring only a single update and relatively low training effort. Although S2 improved predictive performance further, its substantially higher computational overhead may limit its practical applicability in resource-constrained environments.

For \textit{HAT}, adaptive retraining produced substantial improvements over the baseline, particularly under S2, which achieved the highest F1-score and accuracy. However, S4 did not outperform the remaining adaptive approaches, suggesting that the cluster-based adaptation mechanism may be less effective for highly unstable tree-based learners on this dataset. Nevertheless, all adaptive strategies significantly improved performance relative to the static baseline.

The \textit{kNN} classifier benefited moderately from adaptation strategies. S2 achieved the highest predictive performance, while S3, S5, and S6 provided similar results with substantially lower computational requirements. Interestingly, S4 produced no measurable improvement relative to the baseline, indicating that the drift adaptation mechanism did not effectively enhance kNN performance under the considered drift conditions.

For \textit{LogisticRegression}, the evaluated adaptation strategies produced only marginal differences in performance. The classifier demonstrated strong stability across all strategies, with nearly identical F1-score and accuracy values under S1, S3, S4, S5, and S6. This behaviour suggests that LogisticRegression is inherently robust to the drift characteristics present in the Wine dataset and therefore benefits less from aggressive adaptation mechanisms.

A similar pattern was observed for \textit{OnlineLogisticRegression}, where adaptive retraining produced only limited improvements. S2 slightly increased predictive performance, but the remaining strategies exhibited nearly identical results to the baseline. These findings indicate that online linear models maintain relatively stable behaviour even without frequent adaptation under the evaluated conditions.

For \textit{OnlineNaiveBayes}, adaptive retraining did not provide consistent benefits. In fact, S2 slightly degraded predictive performance relative to the baseline, while S3–S6 produced nearly identical results to S1. This suggests that the probabilistic assumptions underlying the classifier may limit the effectiveness of the evaluated adaptation strategies on the Wine dataset.

The \textit{OnlineSoftmaxRegression} classifier also demonstrated limited sensitivity to adaptation strategies. While S3, S5, and S6 produced marginal improvements over the baseline, the gains remained relatively small. S2 slightly reduced performance despite requiring the highest computational effort, indicating that highly aggressive retraining may not be beneficial for this classifier under the considered drift conditions.

Among the ensemble-based approaches, \textit{RandomForest} achieved strong improvements under adaptive retraining. S2 produced the highest predictive performance overall, while S4 achieved comparable gains with substantially lower adaptation effort and update frequency. This result confirms the effectiveness of cluster-based adaptation for ensemble learners in dynamic environments.

Finally, \textit{XGBoost} demonstrated strong responsiveness to adaptive strategies. S2 and S3 achieved the highest predictive performance, while S4 produced no improvement relative to the baseline. Interestingly, S3, S5, and S6 provided competitive performance with very low adaptation cost, suggesting that moderate adaptation frequency may be sufficient for boosting-based models on this dataset.

Figure \ref{fig:wine_grid} displays a selection of timelines for the Wine Quality dataset, where these results shows a much lower overall accuracy and much higher sensivity to drift, with many models struggling to be above 0.50, and where every single model experiences a drastic drop in accuracy immediately after the first batch, suggesting a major initial change in data distribution.

The AMF graph seen in Figure \ref{fig:acc_over_time_AMF_Wine} reveal substantial fluctuations in predictive performance throughout the stream, highlighting the highly dynamic and drift-prone nature of the dataset. Frequent drift boundaries introduced repeated distributional changes, causing noticeable variations in accuracy for all evaluated strategies.

Among the adaptive approaches, S2 and S4 demonstrated the strongest overall adaptation capabilities. S2 maintained relatively stable performance after the middle batches and progressively improved during the later stages of the stream, ultimately reaching accuracies above 0.70. Similarly, S4 showed effective recovery behaviour following temporary degradations and achieved the highest final accuracy among all strategies, indicating strong responsiveness to recurring shift events. In contrast, the static baseline S1 exhibited greater instability and lower overall predictive performance after repeated drift occurrences, confirming the limitations of non-adaptive learning in continuously evolving environments.

Strategies S3, S5, and S6 displayed more conservative adaptation behaviour throughout the stream. Although these approaches generally achieved lower peak performance than S2 and S4, they maintained comparatively smoother accuracy transitions during several drift intervals. S3 and S5 exhibited highly similar behaviour, suggesting that moderate adaptation frequency can provide stable performance under recurring shift conditions without excessive retraining. S6 initially achieved very strong performance during the early batches but later experienced larger fluctuations and reduced stability after multiple drift events, indicating weaker long-term robustness compared with S2 and S4.

Figure \ref{fig:acc_over_time_OkNN_Wine}illustrates the temporal evolution of classification accuracy for the OnlineKNNClassifier on the Wine dataset under the evaluated adaptation strategies. The results reveal substantial variability in predictive performance throughout the stream, indicating the presence of frequent and challenging controlled distribution shift events. Accuracy fluctuations are particularly noticeable after successive drift boundaries, demonstrating the sensitivity of the classifier to evolving data distributions.

Among the evaluated strategies, S2 exhibited the most stable overall behaviour during the later stages of the stream, gradually improving performance and reaching accuracies above 0.70 by the final batches. S4 showed stronger responsiveness to certain drift events, achieving temporary performance peaks around the middle of the stream, although it also experienced larger fluctuations after subsequent drifts. In contrast, the static baseline S1 displayed weaker adaptation capability and lower overall stability under repeated distributional changes.

Strategies S3, S5, and S6 demonstrated more conservative adaptation behaviour across the stream. S3 and S5 maintained relatively stable but moderate performance levels, suggesting that limited adaptation frequency can provide robustness without excessive retraining. S6 initially achieved perfect predictive performance in the first batch but later suffered substantial fluctuations before recovering strongly toward the end of the stream, ultimately achieving one of the highest final accuracy values among all strategies. This behaviour indicates that feature-partitioned adaptation may provide stronger long-term recovery capabilities despite temporary instability during intermediate drift periods.

The OnlineLogisticRegression graph seen in Figure \ref{fig:acc_over_time_OLR_Wine} stablish that the performance fluctuations are more moderate, indicating that OnlineLogisticRegression exhibits relatively stable behaviour under recurring controlled distribution shift conditions. Nevertheless, repeated drift boundaries still introduced noticeable variations in predictive accuracy throughout the stream.

Among the evaluated strategies, S2 consistently achieved the strongest overall predictive performance during most phases of the stream. Although temporary degradations occurred after certain drift events, S2 repeatedly recovered and maintained comparatively higher accuracy values than the remaining approaches. S6 also demonstrated competitive adaptation behaviour, particularly during the early and final batches, where it achieved performance levels similar to S2. These results suggest that both sliding-window adaptation and feature-partitioned adaptation can effectively maintain OnlineLogisticRegression performance under evolving data distributions.

Strategies S3, S4, and S5 exhibited more conservative adaptation behaviour with smaller fluctuations throughout the stream. While these approaches generally produced lower peak accuracies than S2, they maintained relatively stable transitions between drift intervals, indicating improved robustness against abrupt performance degradation. In contrast, the static baseline S1 remained consistently below the adaptive approaches, further confirming the importance of adaptive retraining mechanisms in dynamic streaming environments.

Finally, the Random Forest graph seen in Figure \ref{fig:acc_over_time_RF_Wine} reveal considerable performance variability across the stream, confirming the presence of frequent controlled distribution shift events and the challenging nature of the dataset. Repeated drift boundaries caused noticeable fluctuations in predictive accuracy for all strategies, although the adaptive approaches generally maintained stronger recovery capabilities than the static baseline.

Among the evaluated strategies, S2 demonstrated the most stable overall adaptation behaviour throughout the stream. After temporary degradations caused by drift events, S2 consistently recovered and progressively improved performance during the later batches, ultimately achieving the highest final accuracy among all strategies. S4 also exhibited strong responsiveness to evolving distributions, reaching competitive accuracy levels during several drift intervals and maintaining relatively stable performance after recovery phases.

Strategies S3 and S5 displayed more conservative adaptation behaviour, producing smoother transitions but generally lower peak performance compared with S2 and S4. These approaches appeared less reactive to abrupt drift events, which reduced instability but also limited recovery speed following substantial distributional changes. In contrast, S6 showed highly unstable behaviour throughout the stream. Although it achieved perfect predictive performance during the initial batch, its accuracy later fluctuated considerably and experienced severe degradation around the middle stages of the stream before partially recovering toward the end. This indicates that feature-partitioned adaptation may be highly sensitive to recurring shift conditions in ensemble-based classifiers.

\begin{figure}[htbp]
    \centering
    \subfloat[AMF: Accuracy Over Time]{\includegraphics[width=0.45\textwidth]{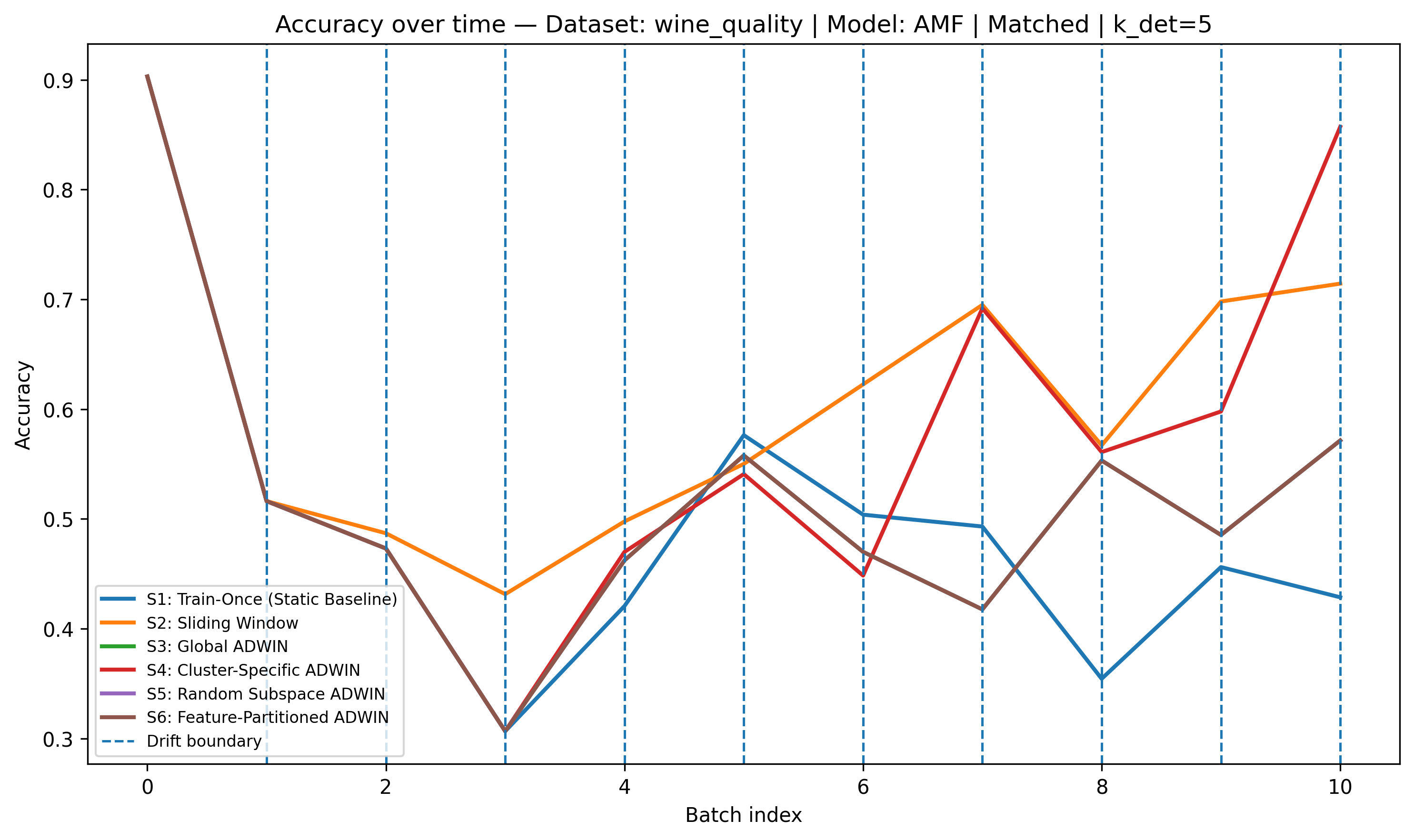}
    \label{fig:acc_over_time_AMF_Wine}}
    \hfill
    \subfloat[Online kNN: Accuracy Over Time]{\includegraphics[width=0.45\textwidth]{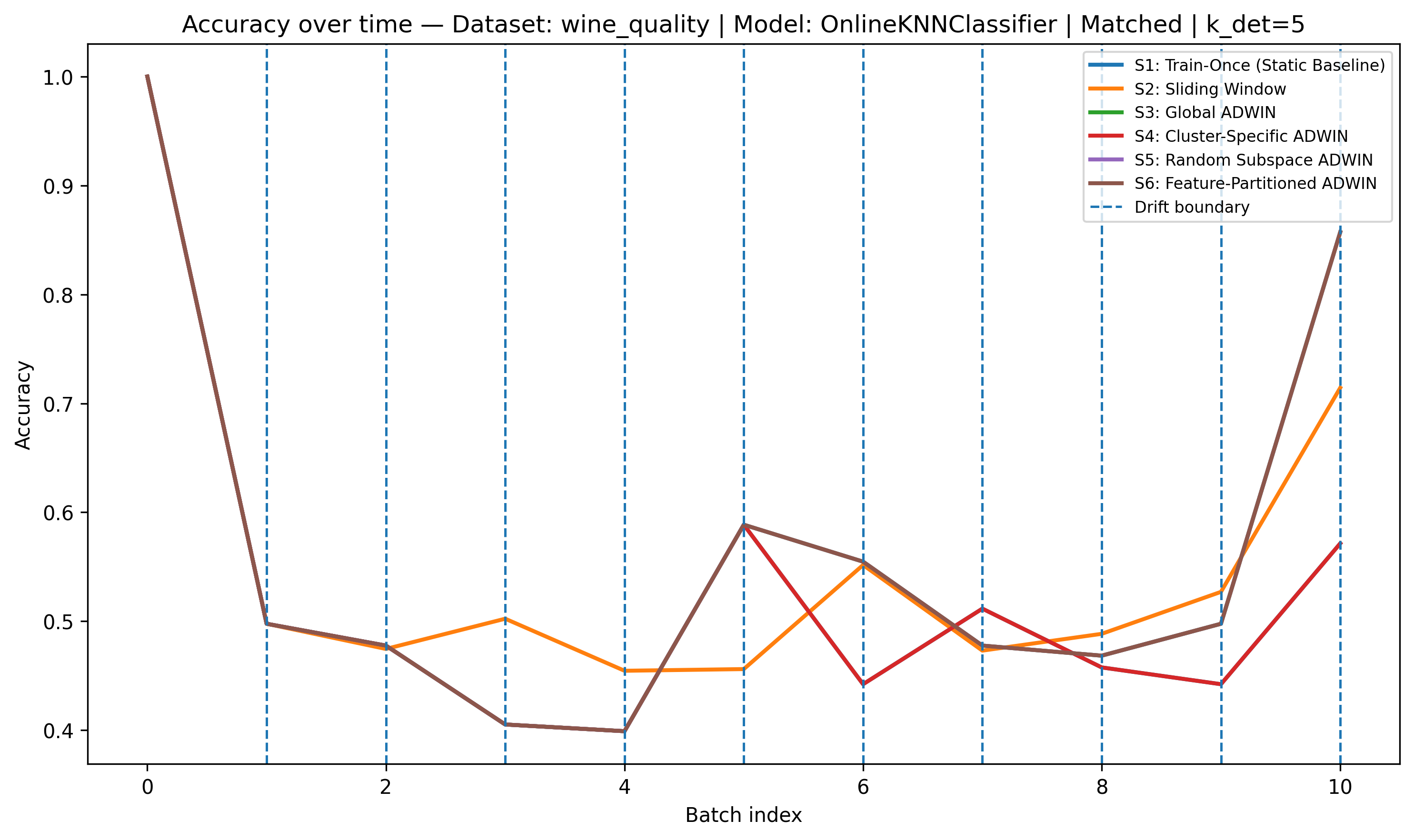}
    \label{fig:acc_over_time_OkNN_Wine}}
    \hfill
    \subfloat[Online LR: Accuracy Over Time]{\includegraphics[width=0.45\textwidth]{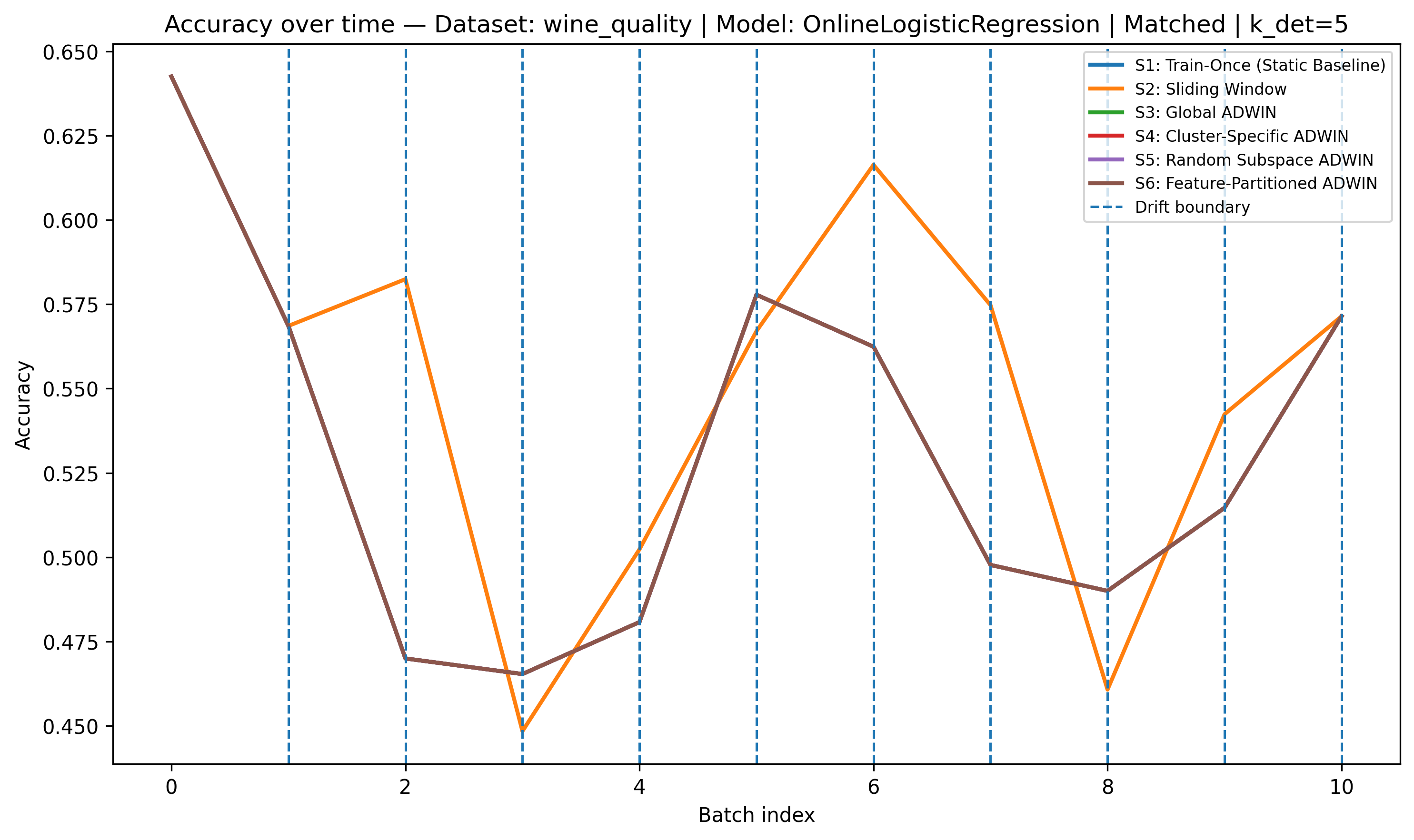}
    \label{fig:acc_over_time_OLR_Wine}}
    \hfill
    \subfloat[Random Forest: Accuracy Over Time]{\includegraphics[width=0.45\textwidth]{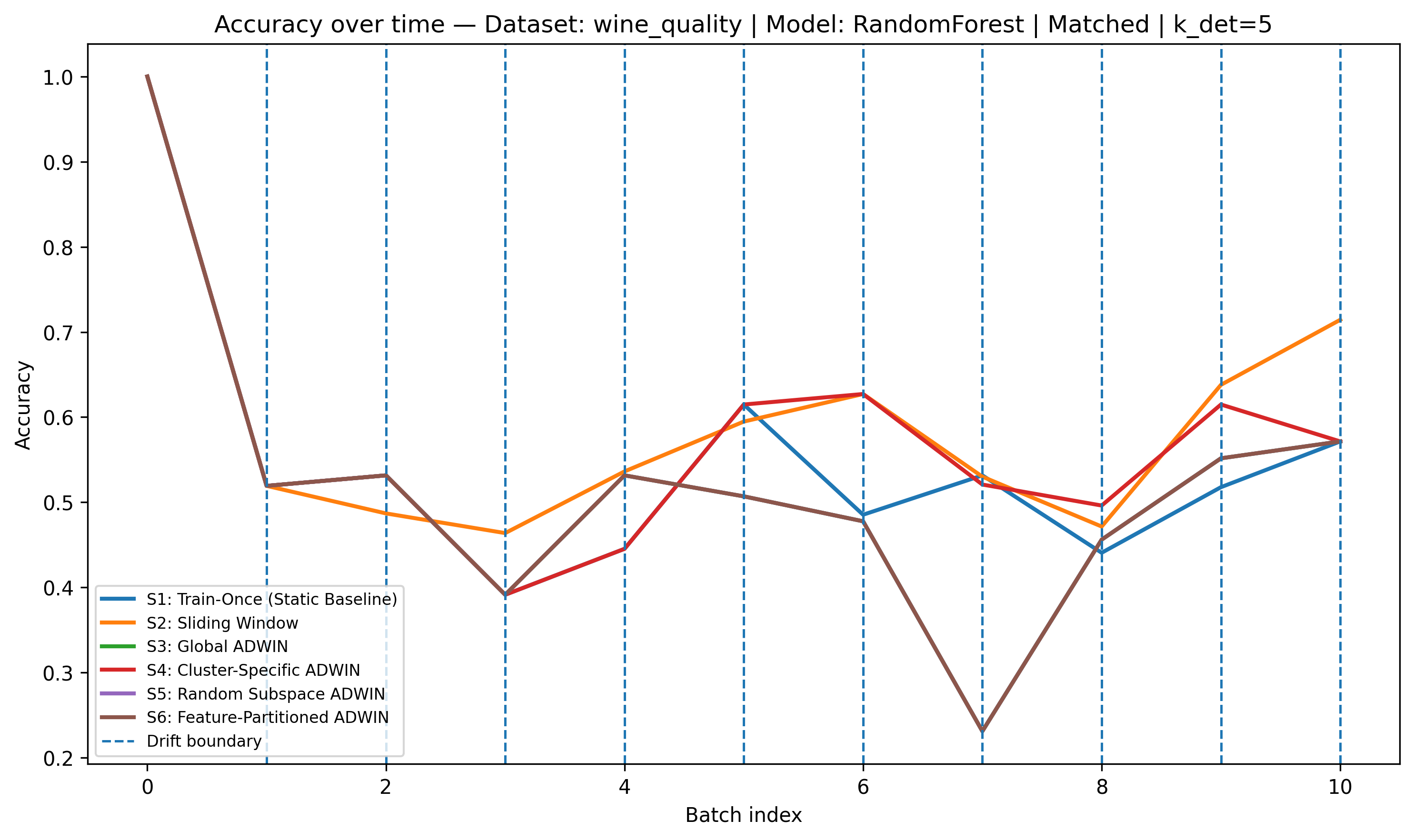}
    \label{fig:acc_over_time_RF_Wine}}
    \caption{Performance and Adaptation Effort Analysis for the Wine Quality Dataset across Strategies S1-S6.}
    \label{fig:wine_grid}
\end{figure}
\FloatBarrier

Overall, the timeline analyses on the Wine dataset demonstrate that recurring controlled distribution shift significantly affects classifier stability and predictive performance, producing substantial fluctuations in accuracy throughout the stream. Across all evaluated classifiers, the static baseline S1 consistently exhibited weaker robustness and limited recovery capability after drift events, confirming the importance of adaptive retraining in dynamic environments. Among the evaluated approaches, S2 generally achieved the most stable recovery and strongest overall predictive performance, although at the cost of higher retraining effort and update frequency. In contrast, S3 and S5 provided more conservative and computationally efficient adaptation behaviour with smoother transitions but slower recovery after abrupt drifts. Overall, S4 offered one of the best trade-offs between predictive stability and adaptation responsiveness, consistently demonstrating effective recovery capabilities while avoiding the excessive adaptation overhead associated with highly reactive strategies such as S2.

\subsubsection{Breast Cancer}

The Breast Cancer dataset represents the smallest stream evaluated in this study and exhibits generally high baseline performance for most classifiers, see table \ref{tab:breast_cancer_results}. Overall, the dataset appears considerably more stable and less affected by severe controlled distribution shift, as reflected by the consistently high F1-scores and accuracies obtained by most classifiers, even under the static baseline S1. Nevertheless, adaptive retraining strategies still provided measurable improvements for several classifiers, particularly under the more reactive adaptation mechanism S2.

\begin{longtable}{@{}llccrr@{}}
\caption{Detailed performance comparison on the Breast Cancer dataset.}
\label{tab:breast_cancer_results}\\
\toprule
\textbf{Classifier} & \textbf{Strat.} & \textbf{Avg. F1} & \textbf{Avg. Acc.} & \textbf{Updates} & \textbf{Effort (Samples)} \\
\midrule
\endfirsthead

\caption[]{Detailed performance comparison on the Breast Cancer dataset (continued).}\\
\toprule
\textbf{Classifier} & \textbf{Strat.} & \textbf{Avg. F1} & \textbf{Avg. Acc.} & \textbf{Updates} & \textbf{Effort (Samples)} \\
\midrule
\endhead

\midrule
\multicolumn{6}{r}{\textit{Continued on next page}}\\
\endfoot

\bottomrule
\endlastfoot

\textbf{AMF} & S1 & 0.577 & 0.552 & 0 & 0 \\
             & S2 & 0.839 & 0.837 & 10 & 3033 \\
             & S3 & 0.602 & 0.566 & 1 & 57 \\
             & S4 & 0.577 & 0.552 & 0 & 0 \\
             & S5 & 0.602 & 0.566 & 1 & 57 \\
             & S6 & 0.602 & 0.566 & 1 & 57 \\
\midrule

\textbf{ARF} & S1 & 0.778 & 0.738 & 0 & 0 \\
             & S2 & 0.881 & 0.851 & 10 & 3033 \\
             & S3 & 0.778 & 0.738 & 0 & 0 \\
             & S4 & 0.778 & 0.738 & 0 & 0 \\
             & S5 & 0.778 & 0.738 & 0 & 0 \\
             & S6 & 0.778 & 0.738 & 0 & 0 \\
\midrule

\textbf{HAT} & S1 & 0.274 & 0.353 & 0 & 0 \\
             & S2 & 0.840 & 0.856 & 10 & 3033 \\
             & S3 & 0.572 & 0.623 & 2 & 113 \\
             & S4 & 0.274 & 0.353 & 0 & 0 \\
             & S5 & 0.572 & 0.623 & 2 & 113 \\
             & S6 & 0.572 & 0.623 & 2 & 113 \\
\midrule

\textbf{kNN} & S1 & 0.876 & 0.854 & 0 & 0 \\
             & S2 & 0.915 & 0.904 & 10 & 3033 \\
             & S3 & 0.876 & 0.854 & 0 & 0 \\
             & S4 & 0.876 & 0.854 & 0 & 0 \\
             & S5 & 0.876 & 0.854 & 0 & 0 \\
             & S6 & 0.876 & 0.854 & 0 & 0 \\
\midrule

\textbf{LogisticRegression} & S1 & 0.952 & 0.956 & 0 & 0 \\
                            & S2 & 0.939 & 0.944 & 10 & 3033 \\
                            & S3 & 0.952 & 0.956 & 0 & 0 \\
                            & S4 & 0.952 & 0.956 & 0 & 0 \\
                            & S5 & 0.952 & 0.956 & 0 & 0 \\
                            & S6 & 0.952 & 0.956 & 0 & 0 \\
\midrule

\textbf{OnlineKNNClassifier} & S1 & 0.703 & 0.661 & 0 & 0 \\
                             & S2 & 0.915 & 0.907 & 10 & 3033 \\
                             & S3 & 0.573 & 0.547 & 1 & 56 \\
                             & S4 & 0.703 & 0.661 & 0 & 0 \\
                             & S5 & 0.573 & 0.547 & 1 & 56 \\
                             & S6 & 0.573 & 0.547 & 1 & 56 \\
\midrule

\textbf{OnlineLogisticRegression} & S1 & 0.884 & 0.899 & 0 & 0 \\
                                  & S2 & 0.923 & 0.933 & 10 & 3033 \\
                                  & S3 & 0.884 & 0.899 & 0 & 0 \\
                                  & S4 & 0.884 & 0.899 & 0 & 0 \\
                                  & S5 & 0.884 & 0.899 & 0 & 0 \\
                                  & S6 & 0.884 & 0.899 & 0 & 0 \\
\midrule

\textbf{OnlineNaiveBayes} & S1 & 0.913 & 0.917 & 0 & 0 \\
                          & S2 & 0.929 & 0.933 & 10 & 3033 \\
                          & S3 & 0.913 & 0.917 & 0 & 0 \\
                          & S4 & 0.913 & 0.917 & 0 & 0 \\
                          & S5 & 0.913 & 0.917 & 0 & 0 \\
                          & S6 & 0.913 & 0.917 & 0 & 0 \\
\midrule

\textbf{OnlineSoftmaxRegression} & S1 & 0.935 & 0.938 & 0 & 0 \\
                                 & S2 & 0.944 & 0.947 & 10 & 3033 \\
                                 & S3 & 0.935 & 0.938 & 0 & 0 \\
                                 & S4 & 0.935 & 0.938 & 0 & 0 \\
                                 & S5 & 0.935 & 0.938 & 0 & 0 \\
                                 & S6 & 0.935 & 0.938 & 0 & 0 \\
\midrule

\textbf{RandomForest} & S1 & 0.926 & 0.922 & 0 & 0 \\
                      & S2 & 0.959 & 0.956 & 10 & 3033 \\
                      & S3 & 0.926 & 0.922 & 0 & 0 \\
                      & S4 & 0.926 & 0.922 & 0 & 0 \\
                      & S5 & 0.926 & 0.922 & 0 & 0 \\
                      & S6 & 0.926 & 0.922 & 0 & 0 \\
\midrule

\textbf{XGBoost} & S1 & 0.947 & 0.944 & 0 & 0 \\
                 & S2 & 0.960 & 0.956 & 10 & 3033 \\
                 & S3 & 0.947 & 0.944 & 0 & 0 \\
                 & S4 & 0.947 & 0.944 & 0 & 0 \\
                 & S5 & 0.947 & 0.944 & 0 & 0 \\
                 & S6 & 0.947 & 0.944 & 0 & 0 \\

\end{longtable}

Among the evaluated approaches, S2 consistently achieved the highest predictive performance across nearly all classifiers. Significant improvements can be observed for adaptive tree-based and neighborhood-based models such as \textit{AMF}, \textit{ARF}, \textit{HAT}, \textit{kNN}, and \textit{OnlineKNNClassifier}. For example, \textit{HAT} improved dramatically from an average accuracy of 0.353 under S1 to 0.856 under S2, while \textit{AMF} increased from 0.552 to 0.837. Similarly, \textit{OnlineKNNClassifier} improved from 0.661 to 0.907, demonstrating the strong effectiveness of continuous retraining under evolving conditions. However, these gains were achieved at the cost of the highest adaptation effort, requiring 10 updates and 3,033 training samples for all classifiers.

Strategies S3, S5, and S6 generally exhibited highly similar behaviour across classifiers, producing only minor improvements relative to the static baseline. In many cases, these strategies achieved nearly identical performance to S1 while requiring either zero or very few updates. This trend suggests that the Breast Cancer dataset contains relatively mild distributional changes, limiting the benefits of moderate adaptation mechanisms. Nevertheless, for more drift-sensitive classifiers such as \textit{HAT} and \textit{AMF}, these strategies still provided noticeable improvements over the static baseline while maintaining extremely low computational overhead.

In contrast to the Adult and Wine datasets, S4 produced limited performance gains for most classifiers. For many models, including \textit{ARF}, \textit{kNN}, \textit{RandomForest}, and \textit{XGBoost}, S4 achieved results identical to the static baseline. This indicates that the cluster-based adaptation mechanism was triggered infrequently or provided limited additional benefit under the relatively stable conditions of the Breast Cancer dataset. Consequently, the advantages of more sophisticated drift-aware adaptation strategies become less pronounced when the underlying data distribution remains comparatively consistent over time.

Linear and probabilistic models such as \textit{LogisticRegression}, \textit{OnlineLogisticRegression}, \textit{OnlineNaiveBayes}, and \textit{OnlineSoftmaxRegression} demonstrated exceptionally strong baseline performance across all strategies. In several cases, S1 already achieved accuracies above 0.93, leaving limited room for improvement through adaptive retraining. These findings indicate that such models exhibit strong inherent robustness on the Breast Cancer dataset and are less sensitive to the evaluated drift conditions.

Ensemble-based classifiers such as \textit{RandomForest} and \textit{XGBoost} consistently achieved the highest overall predictive performance in the study. While S2 further improved their already strong baseline results, the remaining strategies produced nearly identical performance to S1, again suggesting relatively stable stream conditions with limited severe drift.

Figure \ref{fig:breast_grid} displays the timeline for the Breast Cancer dataset, showing the performance of four chosen classifiers for the different adaptation strategies, where it is common to see lines overlapping, as similar performance is obtained between adaptive strategies as compared with the Adult and Wine datasets, the observed performance fluctuations are considerably smaller. 

The ARF graph seen in Figure \ref{fig:acc_over_time_ARF_Breast} shows that among the evaluated approaches, S2 demonstrated the strongest and most consistent adaptation behaviour. After temporary degradations during the early batches, S2 rapidly recovered and maintained very high accuracy values close to 1.0 for most of the remaining stream, indicating excellent robustness to evolving data distributions. In contrast, S6 exhibited substantially greater variability, alternating between high performance and noticeable degradations after several drift events. Although S6 achieved perfect accuracy at certain points in the stream, its recovery behaviour remained less stable than S2 over time.

The remaining strategies, including S1, S3, S4, and S5, produced nearly identical behaviour and therefore overlap in the timeline. Their performance remained relatively stable across the stream but consistently below the highly adaptive S2 strategy. This suggests that the drift conditions present in the Breast Cancer dataset are sufficiently mild for simpler or less reactive adaptation mechanisms to maintain strong predictive performance without requiring extensive retraining.

The HAT graph seen in Figure \ref{fig:acc_over_time_HAT_Breast} reveal substantial differences in adaptation behaviour across the strategies, with repeated drift boundaries causing noticeable fluctuations in predictive performance throughout the stream. Compared with the Adult and Wine datasets, the Breast Cancer dataset still appears relatively stable overall; however, the HAT classifier demonstrates higher sensitivity to drift events than several of the previously analyzed models.

Among the evaluated approaches, S2 consistently achieved the strongest predictive performance and the most stable recovery behaviour. After an initial degradation during the early batches, S2 rapidly adapted to the evolving data distribution and maintained accuracies close to 1.0 for most of the remaining stream. This indicates that the sliding-window adaptation mechanism provides highly effective drift responsiveness for HAT under the considered streaming conditions.

In contrast, S4 exhibited highly unstable behaviour across the stream. Although it achieved temporary improvements at certain points, its performance frequently degraded to very low accuracy values after several drift events, suggesting limited robustness and inconsistent recovery capabilities. Similarly, S6 demonstrated strong fluctuations throughout the evaluation period. While S6 reached perfect accuracy during some batches and achieved strong late-stage recovery, it also suffered severe temporary degradations following multiple drift boundaries, indicating greater sensitivity to abrupt distributional changes.

The remaining strategies, including S1, S3, and S5, overlapped closely throughout most of the stream and therefore appear as a single trajectory in the timeline. These approaches maintained relatively stable but considerably lower predictive performance than S2, suggesting that moderate or limited adaptation mechanisms were insufficient to fully capture the evolving patterns affecting the HAT classifier.

For Online Naive Bayes graph seen in Figure \ref{fig:acc_over_time_ONB_Breast} shows that between the the evaluated approaches, S2 achieved the strongest and most stable overall performance across the stream. After recovering from an early degradation, S2 rapidly adapted to the evolving data distribution and maintained accuracy values close to 1.0 during most of the later batches. This behaviour indicates that the sliding-window adaptation mechanism provides highly effective responsiveness and recovery under the considered drift conditions.

S4 also demonstrated strong adaptation capabilities, particularly during the second half of the stream, where it maintained comparatively high and stable predictive performance after recovering from temporary degradations. Although S4 exhibited larger fluctuations than S2 following certain drift events, it consistently outperformed the more conservative adaptation approaches during the later stages of the evaluation period. In contrast, S6 displayed substantially greater instability throughout the stream. While S6 achieved perfect accuracy at specific points, it also suffered sharp performance drops after multiple drift boundaries, indicating weaker robustness and less consistent recovery behaviour under recurring distributional changes.

The remaining strategies, including S1, S3, and S5, overlapped closely throughout most of the stream and therefore appear as a single trajectory in the timeline. These approaches maintained relatively stable predictive performance but generally remained below the more adaptive S2 and S4 strategies. This suggests that moderate adaptation mechanisms can preserve reasonable stability on the Breast Cancer dataset, although stronger retraining strategies still provide measurable performance advantages.

The Online Softmax Regressor graph seen in Figure \ref{fig:acc_over_time_OSR_Breast} also experiences similar results, where S2 demonstrated the strongest and most consistent adaptation behaviour throughout the stream. After a temporary degradation around the middle batches, S2 rapidly recovered and maintained accuracy values close to 1.0 during most of the later stages of the evaluation period. This indicates that the sliding-window adaptation mechanism effectively preserves predictive performance under evolving data distributions while providing strong recovery capabilities after drift events.

S6 also exhibited strong predictive performance across the stream, maintaining relatively high accuracy values despite experiencing larger fluctuations after certain drift boundaries. Although S6 showed a more pronounced temporary degradation during the middle batches, it consistently recovered and achieved stable performance during the later stages of the stream. This suggests that feature-partitioned adaptation can provide effective long-term robustness for OnlineSoftmaxRegression under recurring shift conditions.

The remaining strategies, including S1, S3, S4, and S5, overlapped almost entirely throughout the timeline and therefore appear as a single trajectory in the graph. Their performance remained highly stable across all drift intervals, indicating that even limited or non-adaptive approaches are capable of maintaining strong predictive performance on the Breast Cancer dataset. This behaviour further confirms that the dataset presents relatively mild distributional changes compared with the Adult and Wine datasets.

\begin{figure}[htbp]
    \centering
    \subfloat[LogisticRegression: Accuracy Over Time]{\includegraphics[width=0.45\textwidth]{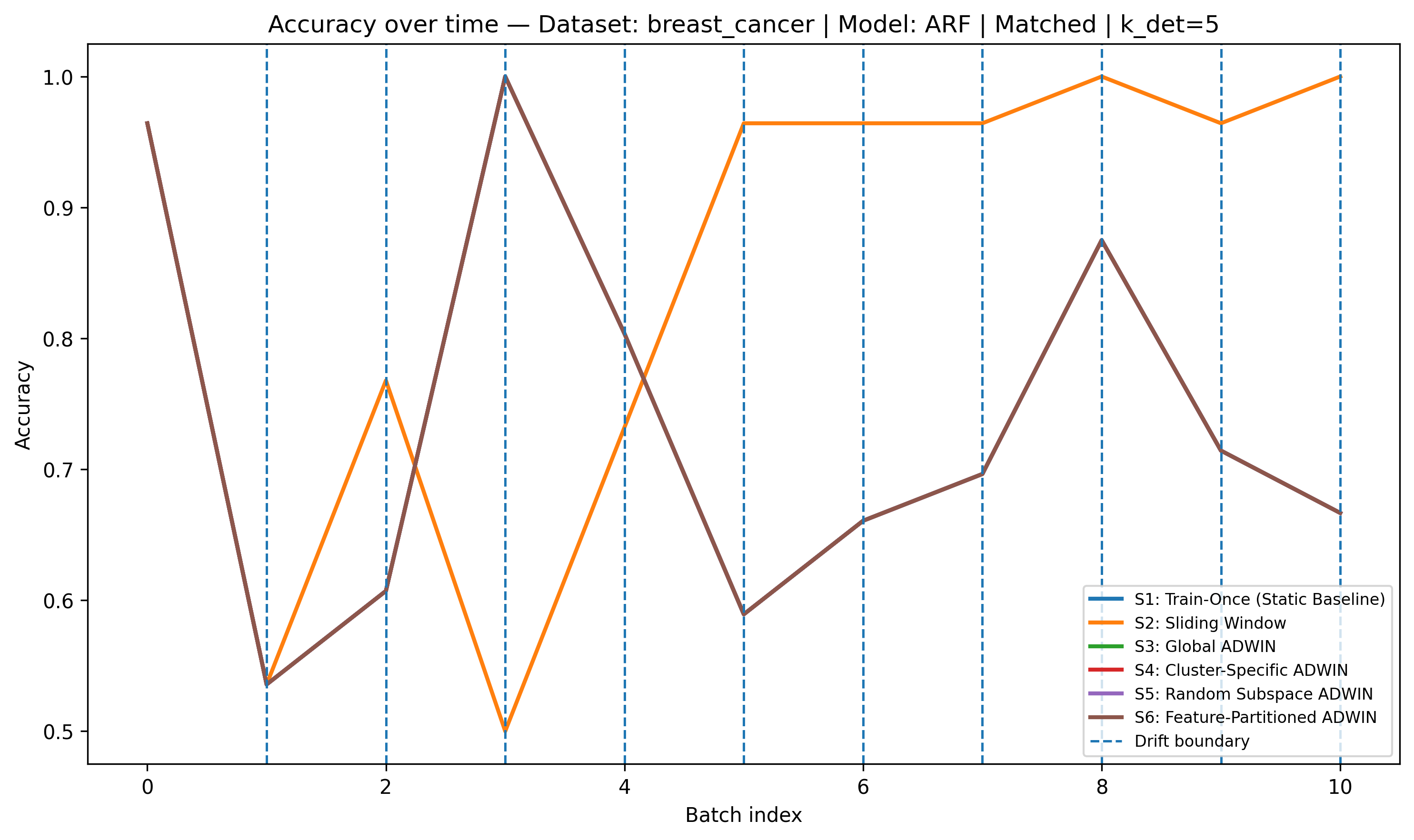}
    \label{fig:acc_over_time_ARF_Breast}}
    \hfill
    \subfloat[RandomForest: Accuracy Over Time]{\includegraphics[width=0.45\textwidth]{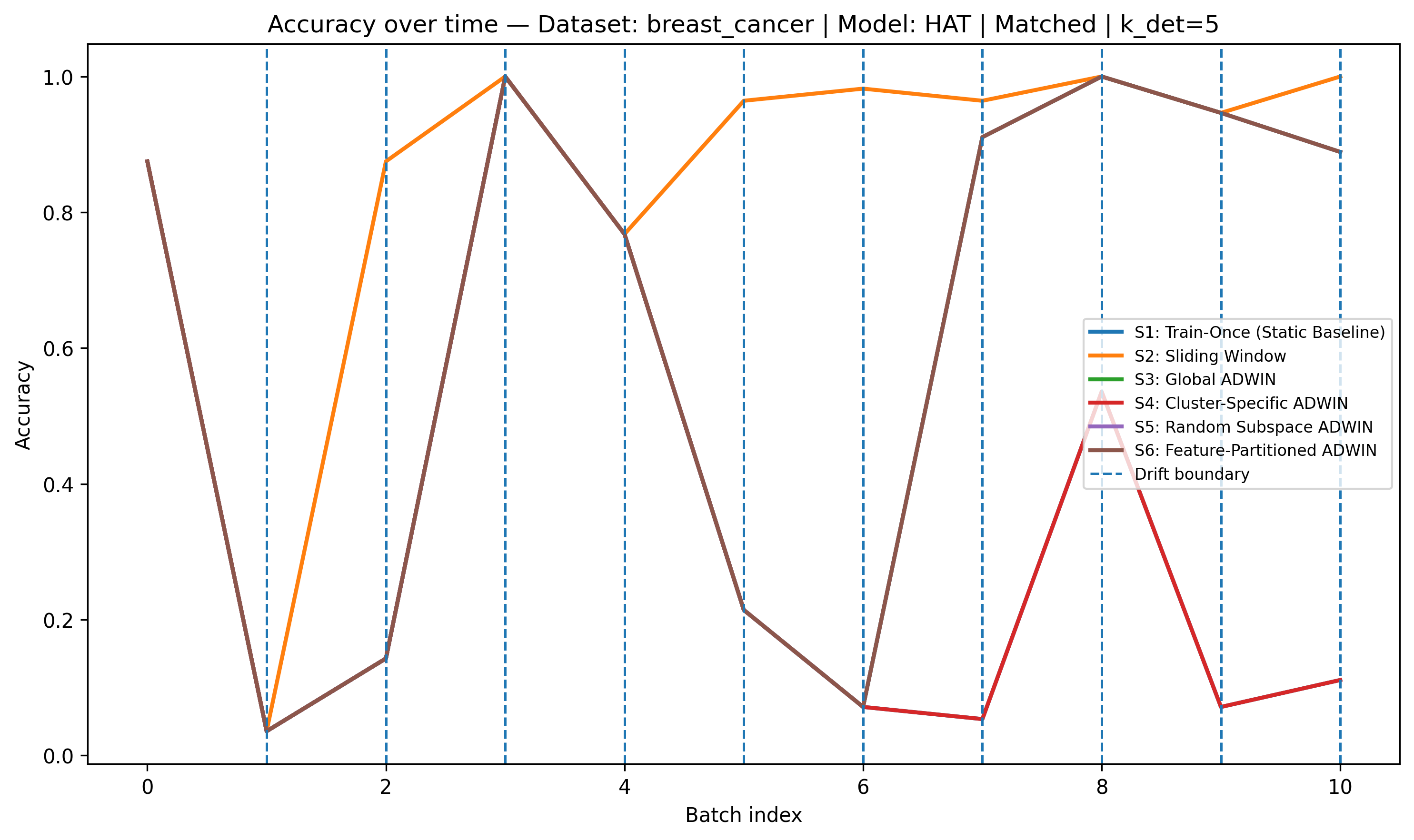}
    \label{fig:acc_over_time_HAT_Breast}}
    \hfill
    \subfloat[kNN: Accuracy Over Time]{\includegraphics[width=0.45\textwidth]{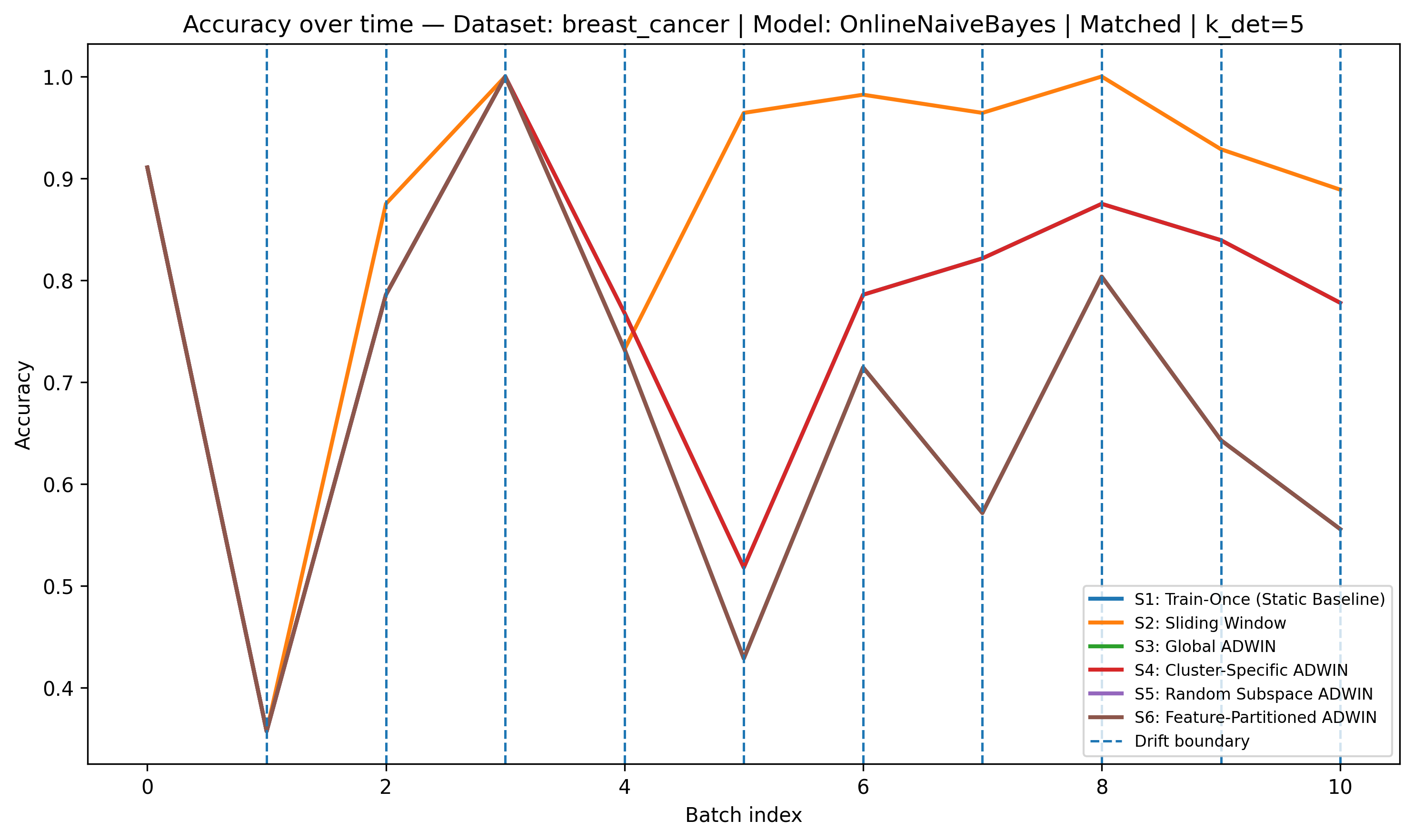}
    \label{fig:acc_over_time_ONB_Breast}}
    \hfill
    \subfloat[XGBoost: Accuracy Over Time]{\includegraphics[width=0.45\textwidth]{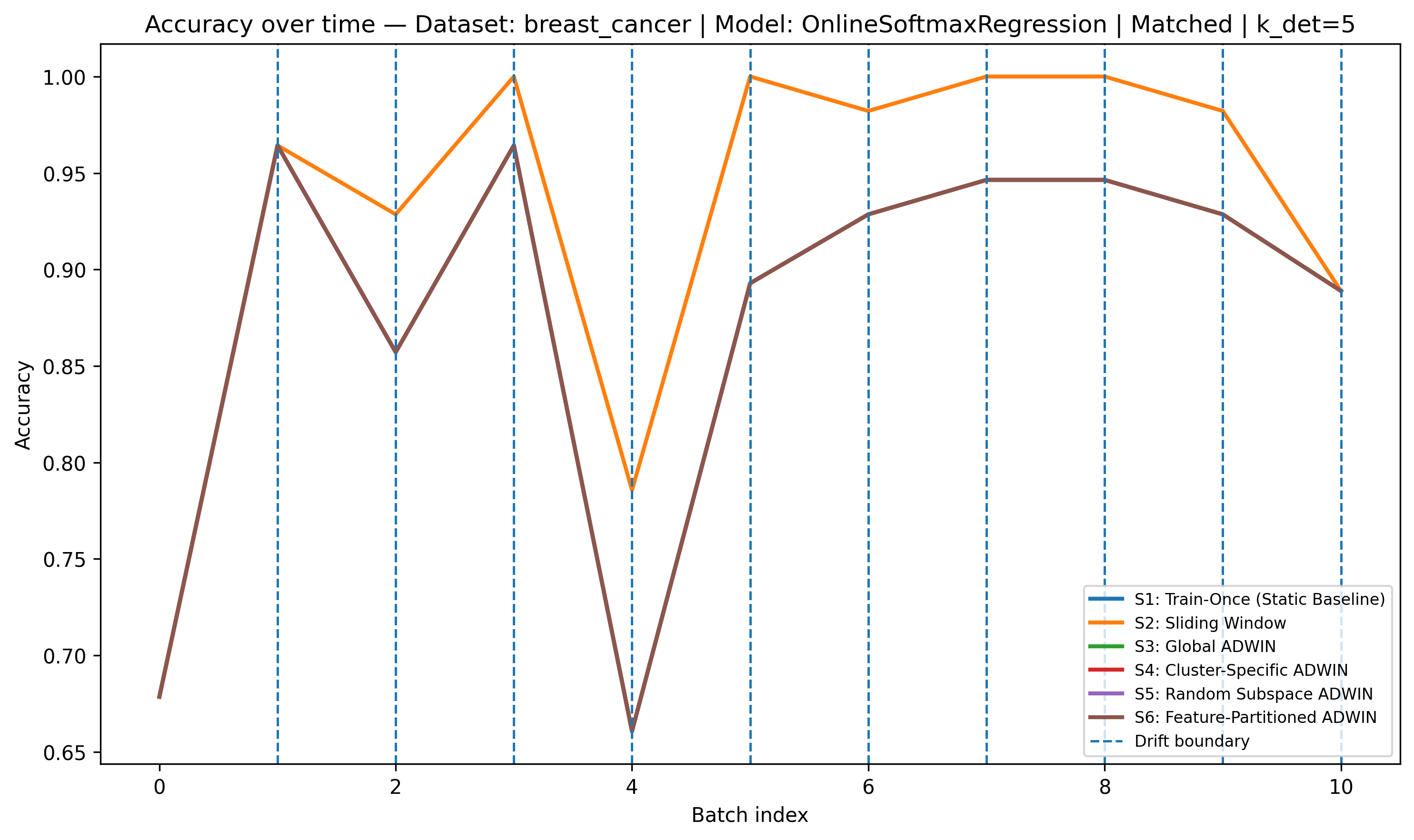}
    \label{fig:acc_over_time_OSR_Breast}}
    \caption{Performance and Adaptation Effort Analysis for the Breast Cancer Dataset across Strategies S1-S6.}
    \label{fig:breast_grid}
\end{figure}
\FloatBarrier

Overall, the timeline analyses for the Breast Cancer dataset indicate that the evaluated classifiers operated under relatively mild controlled distribution shift conditions compared with the Adult and Wine datasets. Most classifiers maintained consistently high predictive performance throughout the stream, and even the static baseline S1 achieved competitive results in several cases, suggesting greater inherent stability in the data distribution. Among the evaluated adaptation strategies, S2 consistently provided the strongest and most stable performance, frequently maintaining accuracies close to 1.0 and demonstrating effective recovery after drift events, although at the cost of higher retraining effort. S4 also showed strong adaptation capabilities for several classifiers, often providing effective recovery and competitive predictive stability while requiring less aggressive retraining than S2. In contrast, S3 and S5 exhibited more conservative but stable behaviour, frequently producing results similar to the baseline due to the limited severity of drift. The results confirm that adaptive retraining improves robustness under evolving conditions, while also showing that aggressive adaptation may provide diminishing returns in relatively stable streaming environments.

\FloatBarrier
\begingroup
\subsection{Regression tasks}\label{secRegressionResults}
The regression experiments evaluate the effectiveness of the proposed adaptation strategies under controlled distribution shift across two unordered and continuous-target datasets, where the performance is assessed using standard regression metrics, including $R^2$, Mean Absolute Error (MAE), and Root Mean Squared Error (RMSE), while computational cost is quantified through the number of model updates and total training effort (samples used for retraining).

Table \ref{tab:avg_r2_regression} presents the average $R^2$ performance obtained by the evaluated regression models across all adaptation strategies. Overall, the results indicate that adaptive retraining strategies substantially improve regression performance compared with the static baseline S1, particularly for ensemble-based and instance-based regressors. Among the evaluated approaches, S2 consistently achieved the strongest predictive performance across most models, confirming the effectiveness of aggressive sliding-window adaptation under evolving regression streams.

The highest overall performance was obtained by \textit{XGBRegressor} under S2, achieving an average $R^2$ value of 0.490, followed by \textit{RandomForestRegressor} with 0.418 and \textit{ARFRegressor} with 0.385. Similarly, \textit{kNNRegressor} achieved positive and stable predictive performance under the adaptive strategies, particularly under S2 and S3. These results demonstrate that ensemble-based and neighborhood-based regressors benefit significantly from continuous adaptation mechanisms capable of responding to distributional changes in the stream.

In contrast, the static baseline S1 consistently produced poor predictive performance for nearly all regression models, often resulting in strongly negative $R^2$ values. This behaviour indicates substantial prediction errors under non-adaptive learning conditions and highlights the inability of static regression models to maintain accurate predictions under evolving data distributions. The degradation was particularly severe for \textit{HoeffdingTreeRegressor}, which produced extremely large negative $R^2$ values across all strategies, suggesting unstable or highly unsuitable behaviour for the evaluated regression scenario.

Strategies S3, S5, and S6 generally exhibited nearly identical performance across most regressors, indicating similar adaptation behaviour and retraining patterns. These strategies often improved performance relative to the static baseline but remained consistently below the more aggressive S2 strategy. S4 demonstrated intermediate behaviour, occasionally outperforming S3, S5, and S6, particularly for \textit{LinearRegression} and \textit{RandomForestRegressor}, while still remaining below the strongest results obtained by S2.

\begin{table}[ht]
\centering
\caption{Average $R^2$ Performance for Regression Tasks Across Strategies}
\label{tab:avg_r2_regression}
\begin{tabular}{lcccccc}
\toprule
\textbf{Model} & \textbf{S1} & \textbf{S2} & \textbf{S3} & \textbf{S4} & \textbf{S5} & \textbf{S6} \\
\midrule
ARFRegressor & -0.331 & \textbf{0.385} & 0.176 & 0.143 & 0.176 & 0.176 \\

HoeffdingTreeRegressor & 
$-7.59 \times 10^{6}$ & 
$-1.43 \times 10^{21}$ & 
$-5.03 \times 10^{20}$ & 
$-1.27 \times 10^{21}$ & 
$-5.03 \times 10^{20}$ & 
$-5.03 \times 10^{20}$ \\

LinearRegression & 
-14.500 & 
-4.792 & 
-2.847 & 
\textbf{-2.642} & 
-2.847 & 
-2.847 \\

RandomForestRegressor & 
-2.859 & 
\textbf{0.418} & 
-0.009 & 
0.064 & 
-0.009 & 
-0.009 \\

XGBRegressor & 
-4.659 & 
\textbf{0.490} & 
-0.102 & 
-0.073 & 
-0.102 & 
-0.102 \\

kNNRegressor & 
-0.288 & 
\textbf{0.365} & 
0.263 & 
0.222 & 
0.263 & 
0.263 \\
\bottomrule
\end{tabular}
\end{table}

Table \ref{tab:avg_mae_regression} presents the average Mean Absolute Error (MAE) obtained by the evaluated regression models across all adaptation strategies. Since lower MAE values indicate better predictive performance, the results demonstrate that adaptive retraining strategies significantly improve regression accuracy compared with the static baseline S1 for most models. Overall, S2 consistently achieved the lowest MAE values across the majority of regressors, confirming the effectiveness of aggressive sliding-window adaptation under evolving regression streams.

Among the evaluated models, \textit{XGBRegressor} achieved the best overall performance under S2 with an MAE of 0.377, followed closely by \textit{RandomForestRegressor} with 0.402 and \textit{ARFRegressor} with 0.411. Similarly, \textit{kNNRegressor} also benefited substantially from adaptive retraining, reducing the MAE from 0.515 under the static baseline to 0.422 under S2. These findings indicate that ensemble-based and instance-based regressors are highly responsive to adaptive updating mechanisms when operating under evolving data distributions.

The static baseline S1 generally produced substantially higher error values across most models, particularly for \textit{LinearRegression}, \textit{RandomForestRegressor}, and \textit{XGBRegressor}. This behaviour confirms the limitations of non-adaptive regression models in continuously changing streaming environments. In contrast, adaptive strategies were able to substantially reduce prediction errors by continuously updating the models in response to evolving patterns in the data stream.

Strategies S3, S5, and S6 again produced nearly identical performance across all regressors, indicating highly similar retraining behaviour and adaptation frequency. Although these approaches consistently improved performance relative to S1, their MAE values remained above those achieved by the more aggressive S2 strategy. S4 generally provided intermediate performance between S2 and the more conservative approaches, occasionally achieving the best result for certain models, such as \textit{LinearRegression}, where it obtained the lowest overall MAE value of 0.981.

Unlike the remaining regressors, \textit{HoeffdingTreeRegressor} exhibited comparatively poor and unstable performance across all strategies, with S2 even producing slightly higher MAE values than the static baseline. This behaviour is consistent with the previously observed instability in the $R^2$ analysis and suggests that this model struggles to adapt effectively under the evaluated regression drift conditions.

\begin{table}[ht]
\centering
\caption{Average MAE Performance for Regression Tasks Across Strategies}
\label{tab:avg_mae_regression}
\begin{tabular}{lcccccc}
\toprule
\textbf{Model} & \textbf{S1} & \textbf{S2} & \textbf{S3} & \textbf{S4} & \textbf{S5} & \textbf{S6} \\
\midrule

ARFRegressor & 
0.516 & 
\textbf{0.411} & 
0.462 & 
0.476 & 
0.462 & 
0.462 \\

HoeffdingTreeRegressor & 
\textbf{0.590} & 
0.673 & 
0.644 & 
0.664 & 
0.644 & 
0.644 \\

LinearRegression & 
1.599 & 
1.220 & 
1.000 & 
\textbf{0.981} & 
1.000 & 
1.000 \\

RandomForestRegressor & 
1.005 & 
\textbf{0.402} & 
0.653 & 
0.599 & 
0.653 & 
0.653 \\

XGBRegressor & 
1.207 & 
\textbf{0.377} & 
0.711 & 
0.698 & 
0.711 & 
0.711 \\

kNNRegressor & 
0.515 & 
\textbf{0.422} & 
0.475 & 
0.488 & 
0.475 & 
0.475 \\

\bottomrule
\end{tabular}
\end{table}

Table \ref{tab:avg_rmse_regression} presents the average Root Mean Squared Error (RMSE) obtained by the evaluated regression models across all adaptation strategies. Since lower RMSE values indicate better predictive performance, the results demonstrate that adaptive retraining substantially improves regression accuracy compared with the static baseline S1 for most models. Overall, S2 consistently achieved the lowest RMSE values across the majority of regressors, confirming the effectiveness of aggressive adaptation under evolving regression streams.

Among the evaluated models, \textit{XGBRegressor} achieved the best overall performance under S2 with an RMSE of 8.173, followed closely by \textit{RandomForestRegressor} with 8.487 and \textit{kNNRegressor} with 9.747. Similarly, \textit{ARFRegressor} showed significant improvement under adaptive retraining, reducing the RMSE from 14.946 under the static baseline to 9.878 under S2. These findings indicate that ensemble-based and instance-based regressors benefit substantially from continuous retraining mechanisms capable of adapting to changing data distributions.

Strategies S3, S5, and S6 again produced nearly identical results across most regressors, indicating similar adaptation behaviour and retraining frequency. While these strategies consistently improved performance relative to the static baseline, their RMSE values generally remained higher than those achieved by S2. S4 provided intermediate performance and, in the case of \textit{LinearRegression}, achieved the best overall result with an RMSE of 13.893, suggesting that moderate adaptation can be beneficial for simpler linear models.

In contrast, \textit{HoeffdingTreeRegressor} exhibited highly unstable behaviour across all strategies, producing extremely large RMSE values, particularly under S2 and S4. This instability is consistent with the previously observed MAE and $R^2$ results and indicates that the model struggles to maintain predictive stability under the evaluated regression drift conditions.

\begin{table}[ht]
\centering
\caption{Average RMSE Performance for Regression Tasks Across Strategies}
\label{tab:avg_rmse_regression}
\begin{tabular}{lcccccc}
\toprule
\textbf{Model} & \textbf{S1} & \textbf{S2} & \textbf{S3} & \textbf{S4} & \textbf{S5} & \textbf{S6} \\
\midrule

ARFRegressor & 
14.946 & 
\textbf{9.878} & 
10.450 & 
11.626 & 
10.450 & 
10.450 \\

HoeffdingTreeRegressor & 
15345.422 & 
$1.39 \times 10^{11}$ & 
$6.97 \times 10^{10}$ & 
$1.29 \times 10^{11}$ & 
$6.97 \times 10^{10}$ & 
$6.97 \times 10^{10}$ \\

LinearRegression & 
26.207 & 
22.918 & 
14.493 & 
\textbf{13.893} & 
14.493 & 
14.493 \\

RandomForestRegressor & 
14.716 & 
\textbf{8.487} & 
9.753 & 
9.374 & 
9.753 & 
9.753 \\

XGBRegressor & 
16.031 & 
\textbf{8.173} & 
9.556 & 
9.236 & 
9.556 & 
9.556 \\

kNNRegressor & 
13.991 & 
\textbf{9.747} & 
9.939 & 
10.095 & 
9.939 & 
9.939 \\

\bottomrule
\end{tabular}
\end{table}

Table \ref{tab:avg_training_samples_regression} presents the average number of training samples processed by each regression model across the evaluated adaptation strategies. The results reveal substantial differences in retraining effort between the strategies, highlighting the trade-off between predictive performance and computational cost in evolving regression streams.

As expected, the static baseline S1 required no additional training samples for any regressor, since the models were trained only once and remained unchanged throughout the stream. In contrast, S2 consistently incurred the highest retraining effort across all models, processing approximately 14,301 samples on average. This confirms that the sliding-window strategy performs the most aggressive adaptation, continuously retraining the models in response to detected changes in the data distribution.

Strategies S3, S5, and S6 again exhibited nearly identical behaviour across all regressors, requiring substantially fewer training samples than S2. Depending on the model, these approaches processed between approximately 2,200 and 3,700 samples on average, indicating considerably lower adaptation overhead while still providing meaningful performance improvements over the static baseline. This consistent behaviour further suggests that these strategies trigger retraining under similar drift conditions and adaptation frequencies.

S4 demonstrated intermediate retraining effort between the conservative and aggressive adaptation strategies. Across most regressors, S4 required between approximately 9,000 and 10,800 training samples, reflecting a more selective adaptation mechanism that retrains less frequently than S2 but substantially more often than S3, S5, and S6. Although this higher retraining effort generally resulted in improved predictive performance compared with the more conservative strategies, the gains were often smaller than those achieved by the fully adaptive S2 strategy.

\begin{table}[ht]
\centering
\caption{Average Training Samples for Regression Tasks Across Strategies}
\label{tab:avg_training_samples_regression}
\begin{tabular}{lcccccc}
\toprule
\textbf{Model} & \textbf{S1} & \textbf{S2} & \textbf{S3} & \textbf{S4} & \textbf{S5} & \textbf{S6} \\
\midrule

ARFRegressor & 
\textbf{0.00} & 
14301.50 & 
3725.00 & 
9506.25 & 
3725.00 & 
3725.00 \\

HoeffdingTreeRegressor & 
\textbf{0.00} & 
14301.50 & 
2225.00 & 
10850.00 & 
2225.00 & 
2225.00 \\

kNNRegressor & 
\textbf{0.00} & 
14301.50 & 
3500.00 & 
9125.00 & 
3500.00 & 
3500.00 \\

LinearRegression & 
\textbf{0.00} & 
14301.50 & 
2725.00 & 
9537.50 & 
2725.00 & 
2725.00 \\

RandomForestRegressor & 
\textbf{0.00} & 
14301.50 & 
3150.00 & 
9693.75 & 
3150.00 & 
3150.00 \\

XGBRegressor & 
\textbf{0.00} & 
14301.50 & 
3650.00 & 
9050.00 & 
3650.00 & 
3650.00 \\

\bottomrule
\end{tabular}
\end{table}

Table \ref{tab:avg_updates_regression} presents the average number of model updates performed by each regression algorithm across the evaluated adaptation strategies. The results reveal clear differences in adaptation frequency, highlighting the trade-off between responsiveness to controlled distribution shift and computational overhead in evolving regression streams.

As expected, the static baseline S1 performed no updates for any regressor, since the models remained fixed after the initial training phase. In contrast, S2 consistently triggered the highest number of updates across all models, reaching 12 updates on average for every regressor. This behaviour confirms that the sliding-window strategy represents the most aggressive adaptation mechanism, continuously retraining the models in response to detected distributional changes.

Strategies S3, S5, and S6 again demonstrated nearly identical update behaviour across all regressors, requiring substantially fewer updates than S2. Depending on the model, these approaches performed between approximately 3.5 and 5 updates on average, indicating a more conservative adaptation mechanism that reacts less frequently to drift events while still maintaining predictive improvements over the static baseline. This consistent behaviour suggests that these strategies share similar drift-triggering and retraining dynamics.

S4 exhibited intermediate adaptation behaviour between the conservative and aggressive strategies. Across the evaluated regressors, S4 performed between approximately 4.5 and 7.1 updates on average, reflecting a more selective retraining mechanism that adapts more frequently than S3, S5, and S6 but less aggressively than S2. This moderate update frequency generally corresponded to improved predictive performance while avoiding the substantial retraining overhead associated with S2.

\begin{table}[ht]
\centering
\caption{Average Number of Updates for Regression Tasks Across Strategies}
\label{tab:avg_updates_regression}
\begin{tabular}{lcccccc}
\toprule
\textbf{Model} & \textbf{S1} & \textbf{S2} & \textbf{S3} & \textbf{S4} & \textbf{S5} & \textbf{S6} \\
\midrule

ARFRegressor & 
\textbf{0.00} & 
12.00 & 
5.00 & 
6.19 & 
5.00 & 
5.00 \\

HoeffdingTreeRegressor & 
\textbf{0.00} & 
12.00 & 
3.50 & 
7.13 & 
3.50 & 
3.50 \\

kNNRegressor & 
\textbf{0.00} & 
12.00 & 
3.50 & 
4.56 & 
3.50 & 
3.50 \\

LinearRegression & 
\textbf{0.00} & 
12.00 & 
4.00 & 
5.94 & 
4.00 & 
4.00 \\

RandomForestRegressor & 
\textbf{0.00} & 
12.00 & 
4.00 & 
5.75 & 
4.00 & 
4.00 \\

XGBRegressor & 
\textbf{0.00} & 
12.00 & 
4.50 & 
5.38 & 
4.50 & 
4.50 \\

\bottomrule
\end{tabular}
\end{table}

\FloatBarrier
\subsubsection{Airfoil Self-Noise}
The Airfoil Self-Noise dataset is used to evaluate the proposed adaptation strategies on a real-world regression task involving aerodynamic noise prediction under streaming conditions. This dataset provides a suitable benchmark for controlled distribution shift analysis because changes in operating conditions and evolving feature-target relationships can significantly affect predictive performance over time. Table \ref{tab:airfoil_results} summarises the detailed experimental results. Overall, the results demonstrate that adaptive retraining substantially improves predictive performance compared with the static baseline S1 for most regressors, particularly for ensemble-based models. Among the evaluated strategies, S2 consistently achieved the strongest predictive performance across nearly all regressors, although at the cost of the highest retraining effort and update frequency.

The best overall performance was obtained by \textit{XGBRegressor} under S2, achieving the highest average $R^2$ value of 0.706 while simultaneously producing the lowest MAE (2.214) and RMSE (3.100). Similarly, \textit{RandomForestRegressor} demonstrated strong performance under S2, obtaining an average $R^2$ of 0.678 with comparatively low prediction errors. \textit{ARFRegressor} and \textit{kNNRegressor} also benefited significantly from adaptive retraining, with S2 substantially improving predictive accuracy relative to the static baseline. These findings indicate that adaptive ensemble-based and instance-based regressors are highly effective under evolving regression streams.

Strategies S3, S5, and S6 again produced nearly identical results across most regressors, suggesting similar retraining behaviour and drift adaptation frequency. While these approaches generally improved predictive performance relative to S1, their results remained consistently below those obtained by S2. S4 typically provided intermediate performance between the conservative and aggressive adaptation strategies, often achieving moderate improvements while requiring substantially fewer updates and training samples than S2. This behaviour suggests that S4 offers a more balanced compromise between predictive performance and computational overhead.

The static baseline S1 consistently produced weaker predictive performance across most regressors, particularly for \textit{LinearRegression} and \textit{HoeffdingTreeRegressor}, which exhibited strongly negative $R^2$ values and large prediction errors. Although adaptive retraining improved the performance of \textit{HoeffdingTreeRegressor}, the model remained unstable across all strategies, indicating limited suitability for the evaluated regression drift conditions.

An interesting observation is the behaviour of \textit{kNNRegressor}, where strategies S3 through S6 produced identical results to the static baseline. This indicates that no effective retraining or adaptation was triggered for these strategies under the considered drift conditions, whereas S2 substantially improved performance through continuous retraining.

\begin{longtable}{@{}llccrrr@{}}
\caption{Detailed performance comparison on the Airfoil dataset.}
\label{tab:airfoil_results}\\

\toprule
\textbf{Regressor} & \textbf{Strat.} & \textbf{Avg. $R^2$} & \textbf{Avg. MAE} & \textbf{Avg. RMSE} & \textbf{Updates} & \textbf{Effort (Samples)} \\
\midrule
\endfirsthead

\caption[]{Detailed performance comparison on the Airfoil dataset (continued).}\\
\toprule
\textbf{Regressor} & \textbf{Strat.} & \textbf{Avg. $R^2$} & \textbf{Avg. MAE} & \textbf{Avg. RMSE} & \textbf{Updates} & \textbf{Effort (Samples)} \\
\midrule
\endhead

\midrule
\multicolumn{7}{r}{\textit{Continued on next page}}\\
\endfoot

\bottomrule
\endlastfoot

\textbf{ARFRegressor} 
& S1 & -0.052 & 5.169 & 6.269 & 0 & 0 \\
& S2 & 0.520 & 3.258 & 4.235 & 12 & 8103 \\
& S3 & 0.312 & 4.080 & 5.072 & 3 & 450 \\
& S4 & 0.400 & 3.731 & 4.568 & 4 & 1200 \\
& S5 & 0.312 & 4.080 & 5.072 & 3 & 450 \\
& S6 & 0.312 & 4.080 & 5.072 & 3 & 450 \\
\midrule

\textbf{HoeffdingTreeRegressor}
& S1 & -189.934 & 74.026 & 108.714 & 0 & 0 \\
& S2 & -69.721 & 28.318 & 53.605 & 12 & 8103 \\
& S3 & -95.327 & 44.208 & 65.434 & 3 & 450 \\
& S4 & -64.124 & 30.131 & 51.433 & 3 & 900 \\
& S5 & -95.327 & 44.208 & 65.434 & 3 & 450 \\
& S6 & -95.327 & 44.208 & 65.434 & 3 & 450 \\
\midrule

\textbf{LinearRegression}
& S1 & -5.905 & 12.246 & 16.069 & 0 & 0 \\
& S2 & -2.500 & 7.303 & 11.443 & 12 & 8103 \\
& S3 & -2.672 & 7.635 & 11.720 & 3 & 450 \\
& S4 & -2.724 & 7.666 & 11.803 & 2 & 600 \\
& S5 & -2.672 & 7.635 & 11.720 & 3 & 450 \\
& S6 & -2.672 & 7.635 & 11.720 & 3 & 450 \\
\midrule

\textbf{RandomForestRegressor}
& S1 & 0.111 & 4.305 & 5.766 & 0 & 0 \\
& S2 & 0.678 & 2.357 & 3.470 & 12 & 8103 \\
& S3 & 0.404 & 3.453 & 4.719 & 2 & 300 \\
& S4 & 0.514 & 3.089 & 4.262 & 2 & 600 \\
& S5 & 0.404 & 3.453 & 4.719 & 2 & 300 \\
& S6 & 0.404 & 3.453 & 4.719 & 2 & 300 \\
\midrule

\textbf{XGBRegressor}
& S1 & -0.063 & 4.621 & 6.301 & 0 & 0 \\
& S2 & 0.706 & 2.214 & 3.100 & 12 & 8103 \\
& S3 & 0.364 & 3.448 & 4.875 & 2 & 300 \\
& S4 & 0.423 & 3.201 & 4.642 & 1 & 300 \\
& S5 & 0.364 & 3.448 & 4.875 & 2 & 300 \\
& S6 & 0.364 & 3.448 & 4.875 & 2 & 300 \\
\midrule

\textbf{kNNRegressor}
& S1 & 0.305 & 3.887 & 5.099 & 0 & 0 \\
& S2 & 0.558 & 3.111 & 4.065 & 12 & 8103 \\
& S3 & 0.305 & 3.887 & 5.099 & 0 & 0 \\
& S4 & 0.305 & 3.887 & 5.099 & 0 & 0 \\
& S5 & 0.305 & 3.887 & 5.099 & 0 & 0 \\
& S6 & 0.305 & 3.887 & 5.099 & 0 & 0 \\

\end{longtable}

Figure \ref{fig:airfoil_grid} provides a complementary grid-based analysis of the Airfoil Self-Noise experiments by combining batch-wise $R^2$ trajectories with adaptation behaviour across all four strategies. This visual perspective highlights how predictive performance evolves after each drift boundary and enables direct comparison between predictive stability and retraining frequency.

For Adaptive Random Forest Regressor (Figure \ref{fig:r2_over_time_ARFR_airfoil}), where the static baseline S1 exhibited highly unstable behaviour, frequently producing negative $R^2$ values after multiple drift boundaries, indicating poor generalization under non-adaptive learning. In contrast, S2 consistently achieved the strongest and most stable predictive performance, rapidly recovering after drift events and maintaining high positive $R^2$ values throughout most of the stream. S4 also demonstrated effective adaptation capabilities, particularly during the later stages of the stream, where it achieved performance close to S2. S6 provided moderate but stable recovery behaviour, consistently outperforming the static baseline while remaining below the stronger adaptive strategies. Overall, the results confirm that aggressive and cluster-aware retraining strategies significantly enhance ARFRegressor robustness under recurring regression drift conditions.

The Hoeffding Tree Regressor timeline (Figure \ref{fig:r2_over_time_HTR_airfoil}) reveals highly unstable regression performance under evolving data conditions. The static baseline S1 consistently produced extremely negative $R^2$ values throughout the stream, indicating severe predictive degradation and poor robustness to controlled distribution shift. Although all adaptive strategies initially experienced substantial performance drops following early drift events, S2 and S4 demonstrated the strongest recovery behaviour, rapidly converging toward near-zero or slightly positive $R^2$ values during the later stages of the stream. S6 exhibited moderate improvement compared with the static baseline but remained less stable than S2 and S4, particularly during intermediate drift intervals. Overall, the results indicate that adaptive retraining substantially mitigates the severe degradation observed in the non-adaptive baseline; however, HoeffdingTreeRegressor remains considerably less stable and less effective than the other evaluated regression models under recurring shift conditions.

When looking at the kNN Regressor timeline (Figure \ref{fig:r2_over_time_kNNr_airfoil}), it can be difficult to appreciate that S1, S3, S4, S5, and S6 overlap entirely across the stream, indicating that these strategies produced equivalent predictive behaviour and did not provide measurable improvement over the static baseline. Their $R^2$ values fluctuate with the drift boundaries, but the adaptation mechanisms in S3-S6 do not appear to alter the model trajectory relative to S1. In contrast, S2 is the only strategy that clearly separates from the overlapping group, showing stronger recovery after drift events and maintaining higher positive $R^2$ values during most later batches. Overall, the results indicate that, for kNNRegressor on the Airfoil dataset, the sliding-window strategy S2 is the only adaptation mechanism that substantially improves performance, while S3, S4, S5, and S6 behave similarly to the non-adaptive baseline.

Then, the Random Forest Regressor timeline (Figure \ref{fig:r2_over_time_RFR_airfoil}) demonstrates that adaptive retraining substantially improves regression performance under recurring shift conditions. The static baseline S1 exhibited unstable behaviour throughout the stream, including several strong degradations and negative $R^2$ values after major drift boundaries, indicating limited robustness to evolving data distributions.

Among the evaluated strategies, S2 consistently achieved the strongest and most stable performance. After initial fluctuations, S2 rapidly recovered following drift events and maintained high positive $R^2$ values throughout most of the stream, ultimately reaching near-perfect performance in the final batches. S4 also showed strong adaptation capability, particularly during the later stages of the stream, where it significantly outperformed the static baseline and maintained stable positive predictive performance.

In contrast, S3, S5, and S6 largely overlap across the timeline, indicating that these strategies produced very similar predictive behaviour under the evaluated drift conditions. Although they generally improved upon the static baseline during several intervals, their performance remained less stable and consistently below that achieved by S2. Overall, the results suggest that aggressive adaptive retraining, particularly through the sliding-window strategy S2, provides the most effective robustness improvements for RandomForestRegressor on the Airfoil dataset.

\begin{figure}[htbp]
    \centering
    \subfloat[Adaptive Random Forest Regressor: $R^2$ Over Time]{\includegraphics[width=0.45\textwidth]{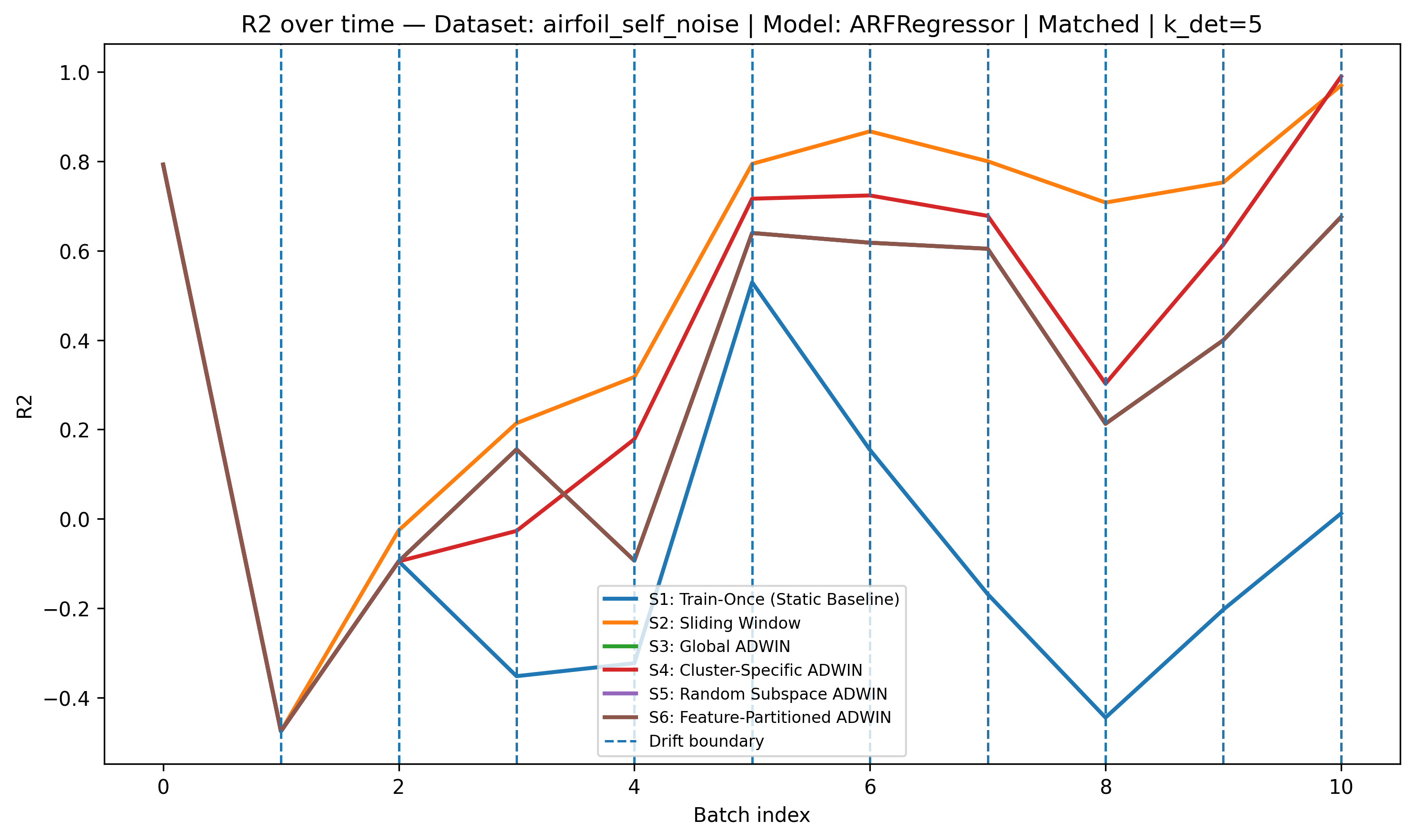}
    \label{fig:r2_over_time_ARFR_airfoil}}
    \hfill
    \subfloat[Hoeffding Tree Regressor: $R^2$ Over Time]{\includegraphics[width=0.45\textwidth]{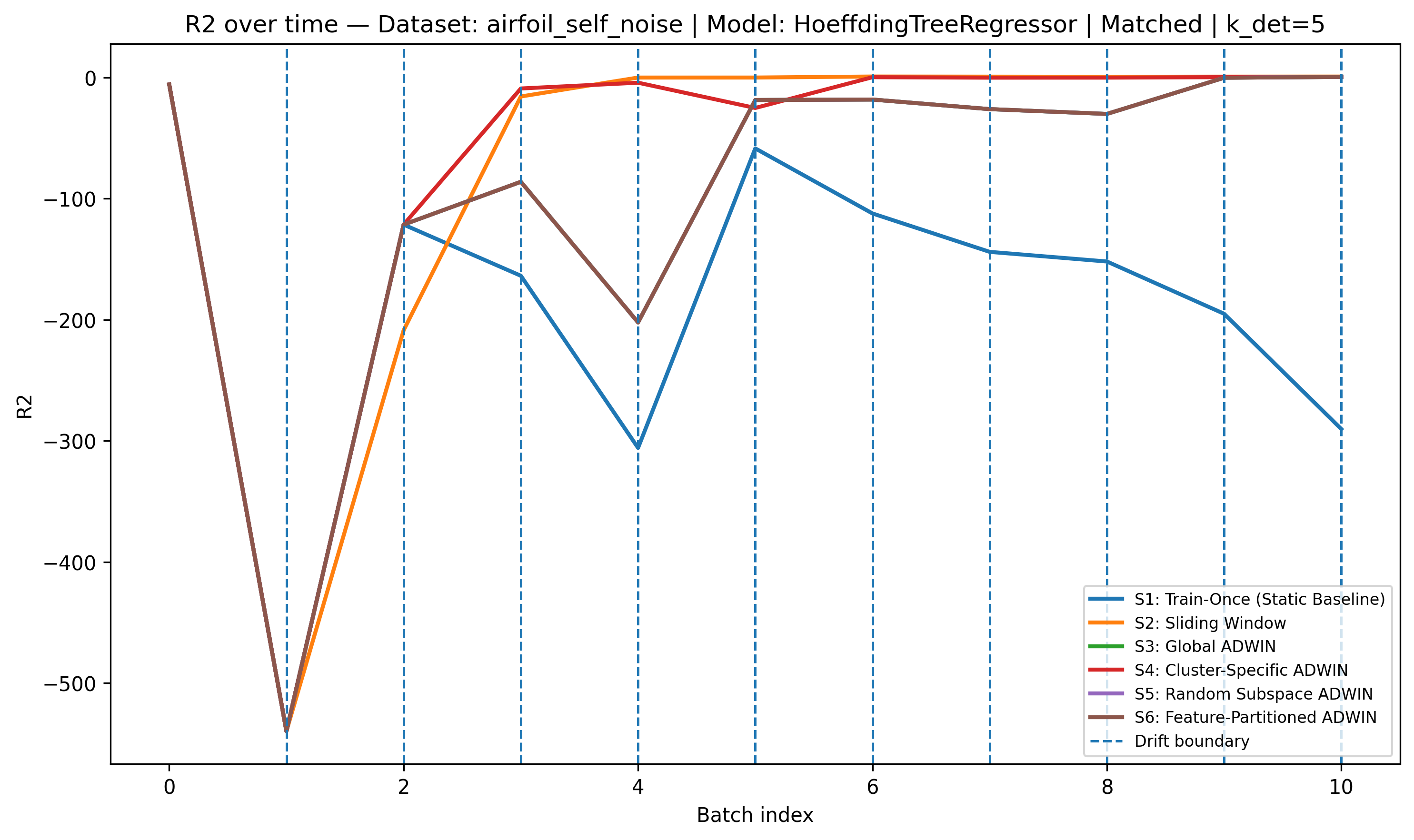}
    \label{fig:r2_over_time_HTR_airfoil}}
    \hfill
    \subfloat[kNN Regressor: $R^2$ Over Time]{\includegraphics[width=0.45\textwidth]{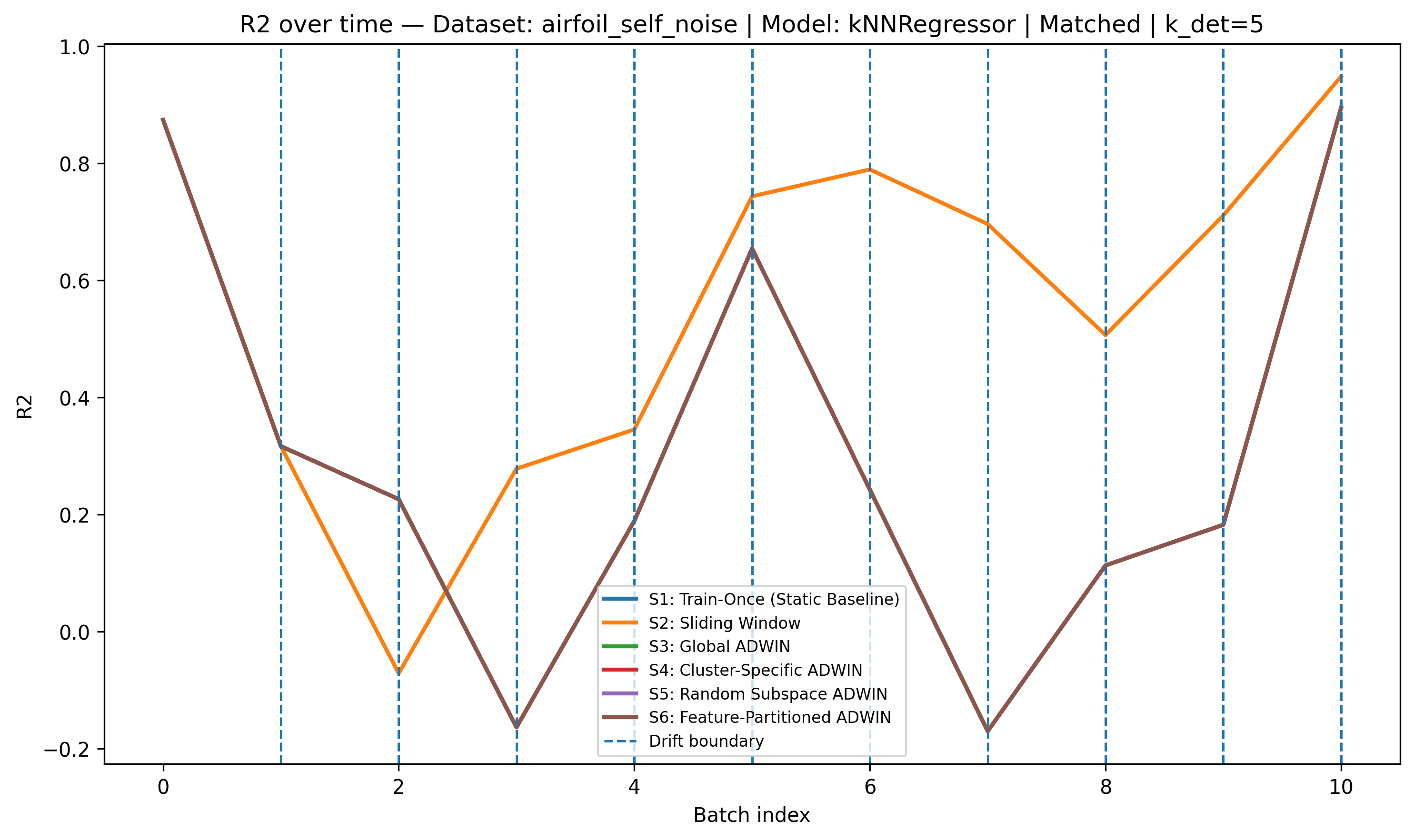}
    \label{fig:r2_over_time_kNNr_airfoil}}
    \hfill
    \subfloat[Random Forest Regressor: $R^2$ Over Time]{\includegraphics[width=0.45\textwidth]{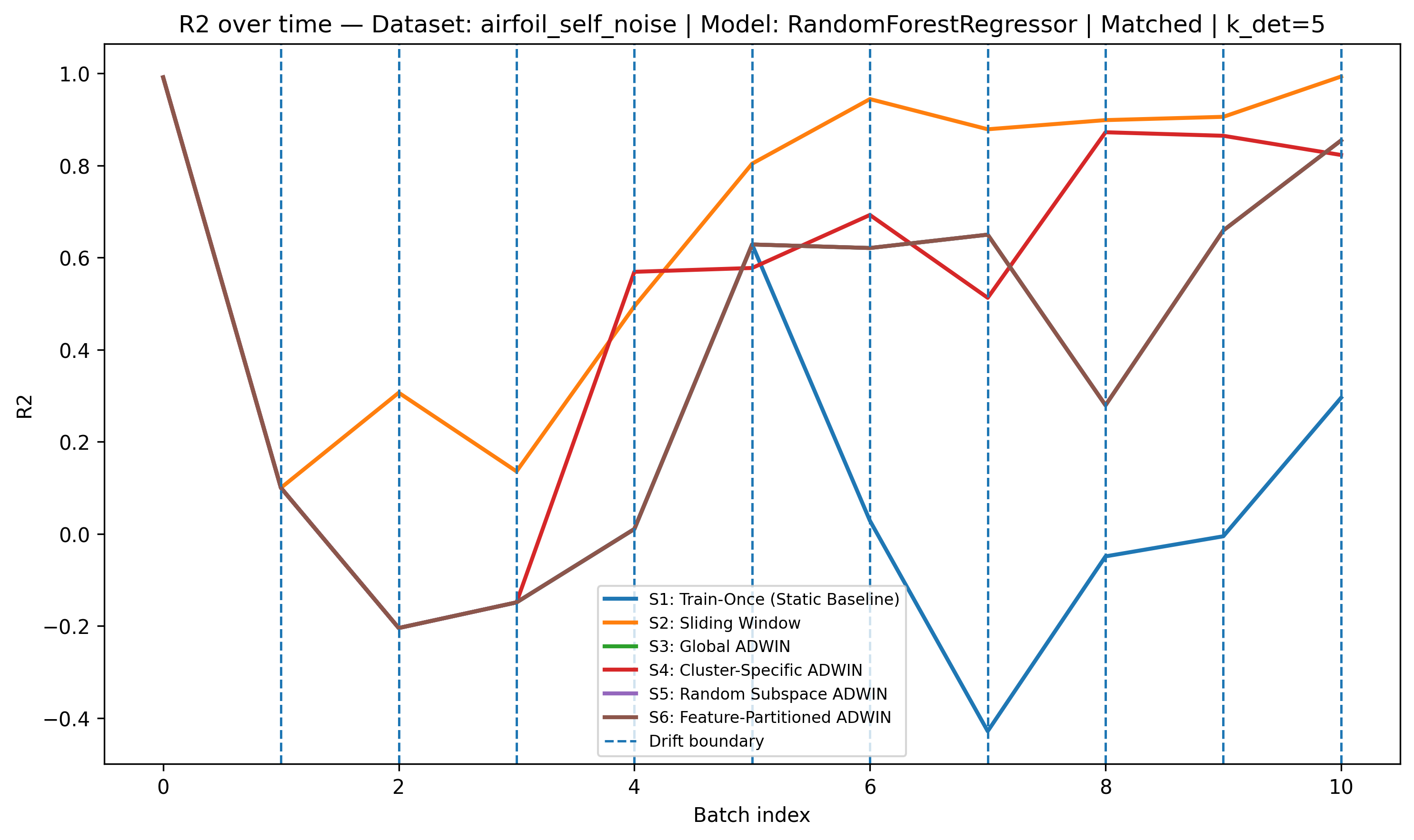}
    \label{fig:r2_over_time_RFR_airfoil}}
    \caption{Performance and Adaptation Analysis for the Airfoil Self-Noise Dataset across Strategies S1-S6.}
    \label{fig:airfoil_grid}
\end{figure}

Overall, the Airfoil Self-Noise results consistently demonstrate that adaptive retraining strategies significantly improve regression robustness under recurring controlled distribution shift conditions when compared with the non-adaptive baseline S1. Across most evaluated regressors, the static strategy frequently suffered from unstable behaviour, including sharp degradations and prolonged negative $R^2$ values after drift boundaries, highlighting the difficulty of maintaining predictive performance in evolving environments without adaptation. Among all strategies, S2 (Sliding Window) repeatedly emerged as the most effective and stable approach, consistently recovering rapidly after drift events and maintaining the highest overall predictive performance throughout the streams. S4 (Cluster-Specific ADWIN) also showed strong adaptation capabilities in several models, particularly for RandomForestRegressor and ARFRegressor, where it achieved substantial improvements over the baseline. In contrast, S3, S5, and S6 often exhibited highly overlapping trajectories, indicating similar adaptation behaviour and limited differentiation in their impact on predictive performance for several regressors. Overall, the results confirm that aggressive retraining approaches, especially sliding-window adaptation, provide the most reliable mechanism for handling recurring shift in non-stationary regression streams on the Airfoil dataset.

\FloatBarrier
\subsubsection{Superconductivty data}
The Superconductivty dataset is used to evaluate the proposed adaptation strategies on a complex real-world regression task involving the prediction of critical temperature from material properties under streaming conditions. This dataset presents a challenging benchmark for controlled distribution shift analysis due to its high dimensionality, nonlinear feature interactions, and the possibility of changing feature relevance over time. Table \ref{tab:superconductivity_results} summarises the detailed experimental results across the adaptation strategies.

These results show that adaptive retraining strategies substantially improve regression performance on the Superconductivity dataset when compared with the non-adaptive baseline S1. Across most regressors, the static baseline produced strongly negative $R^2$ values alongside high MAE and RMSE scores, indicating poor predictive capability under evolving data conditions. Among all evaluated strategies, S2 consistently delivered the strongest overall performance, achieving the highest positive $R^2$ values and the lowest prediction errors for ARFRegressor, RandomForestRegressor, and XGBRegressor. These improvements, however, came at the cost of the highest adaptation effort and update frequency.

Cluster-aware adaptation through S4 also demonstrated competitive performance, particularly for LinearRegression, where it achieved the best overall balance between predictive accuracy and error reduction. In contrast, S3, S5, and S6 produced nearly identical results across all regressors, suggesting that these strategies triggered highly similar adaptation behaviour under the evaluated drift conditions. Although they generally improved over the static baseline, their predictive gains remained below those achieved by S2 and, in some cases, S4.

The HoeffdingTreeRegressor remained highly unstable across all strategies, producing extremely negative $R^2$ values and exceptionally large error magnitudes, indicating that this model struggled to adapt effectively to the dataset regardless of the retraining mechanism. Overall, the results confirm that aggressive adaptive retraining strategies, especially sliding-window adaptation (S2), provide the most robust and consistent performance improvements for non-stationary regression streams in the Superconductivity dataset.

\begin{longtable}{@{}llccrrr@{}}
\caption{Detailed performance comparison on the Superconductivity dataset.}
\label{tab:superconductivity_results}\\

\toprule
\textbf{Regressor} & \textbf{Strat.} & \textbf{Avg. $R^2$} & \textbf{Avg. MAE} & \textbf{Avg. RMSE} & \textbf{Updates} & \textbf{Effort (Samples)} \\
\midrule
\endfirsthead

\caption[]{Detailed performance comparison on the Superconductivity dataset (continued).}\\
\toprule
\textbf{Regressor} & \textbf{Strat.} & \textbf{Avg. $R^2$} & \textbf{Avg. MAE} & \textbf{Avg. RMSE} & \textbf{Updates} & \textbf{Effort (Samples)} \\
\midrule
\endhead

\midrule
\multicolumn{7}{r}{\textit{Continued on next page}}\\
\endfoot

\bottomrule
\endlastfoot

\textbf{ARFRegressor}
& S1 & -0.611 & 18.110 & 23.623 & 0 & 0 \\
& S2 & 0.251 & 11.239 & 15.723 & 14 & 20500 \\
& S3 & 0.040 & 11.702 & 15.829 & 7 & 7000 \\
& S4 & 0.079 & 12.683 & 17.159 & 10 & 20000 \\
& S5 & 0.040 & 11.702 & 15.829 & 7 & 7000 \\
& S6 & 0.040 & 11.702 & 15.829 & 7 & 7000 \\
\midrule

\textbf{HoeffdingTreeRegressor}
& S1 & $-1.52 \times 10^{7}$ & 24018.212 & 30610.225 & 0 & 0 \\
& S2 & $-2.85 \times 10^{21}$ & $1.89 \times 10^{11}$ & $2.77 \times 10^{11}$ & 14 & 20500 \\
& S3 & $-1.01 \times 10^{21}$ & $1.04 \times 10^{11}$ & $1.39 \times 10^{11}$ & 4 & 4000 \\
& S4 & $-2.03 \times 10^{21}$ & $1.77 \times 10^{11}$ & $2.63 \times 10^{11}$ & 9 & 18000 \\
& S5 & $-1.01 \times 10^{21}$ & $1.04 \times 10^{11}$ & $1.39 \times 10^{11}$ & 4 & 4000 \\
& S6 & $-1.01 \times 10^{21}$ & $1.04 \times 10^{11}$ & $1.39 \times 10^{11}$ & 4 & 4000 \\
\midrule

\textbf{kNNRegressor}
& S1 & -0.880 & 17.143 & 22.997 & 0 & 0 \\
& S2 & 0.172 & 10.362 & 15.542 & 14 & 20500 \\
& S3 & 0.221 & 9.979 & 14.894 & 7 & 7000 \\
& S4 & 0.138 & 9.815 & 15.025 & 13 & 26000 \\
& S5 & 0.221 & 9.979 & 14.894 & 7 & 7000 \\
& S6 & 0.221 & 9.979 & 14.894 & 7 & 7000 \\
\midrule

\textbf{LinearRegression}
& S1 & -23.094 & 27.655 & 37.631 & 0 & 0 \\
& S2 & -7.084 & 25.126 & 37.386 & 14 & 20500 \\
& S3 & -3.021 & 14.943 & 20.090 & 5 & 5000 \\
& S4 & -2.559 & 13.849 & 18.734 & 12 & 24000 \\
& S5 & -3.021 & 14.943 & 20.090 & 5 & 5000 \\
& S6 & -3.021 & 14.943 & 20.090 & 5 & 5000 \\
\midrule

\textbf{RandomForestRegressor}
& S1 & -5.829 & 19.176 & 23.926 & 0 & 0 \\
& S2 & 0.159 & 9.390 & 13.947 & 14 & 20500 \\
& S3 & -0.422 & 10.705 & 15.062 & 6 & 6000 \\
& S4 & -0.284 & 9.669 & 14.346 & 13 & 26000 \\
& S5 & -0.422 & 10.705 & 15.062 & 6 & 6000 \\
& S6 & -0.422 & 10.705 & 15.062 & 6 & 6000 \\
\midrule

\textbf{XGBRegressor}
& S1 & -9.254 & 18.339 & 26.134 & 0 & 0 \\
& S2 & 0.273 & 9.089 & 13.496 & 14 & 20500 \\
& S3 & -0.569 & 9.735 & 14.672 & 7 & 7000 \\
& S4 & -0.445 & 9.062 & 13.815 & 10 & 20000 \\
& S5 & -0.569 & 9.735 & 14.672 & 7 & 7000 \\
& S6 & -0.569 & 9.735 & 14.672 & 7 & 7000 \\

\end{longtable}

Figure \ref{fig:super_grid} provides a complementary grid-based comparison of the batch-wise $R^2$ trajectories for the Superconductivity dataset across all evaluated regressors. The figure illustrates how predictive performance evolves over time under recurring shift conditions while contrasting the behaviour of the four adaptation strategies. By visualising model performance at each batch, the timelines provide detailed insight into predictive stability, sensitivity to drift boundaries, and recovery after distributional changes.

\begin{figure}[htbp]
    \centering
    \subfloat[Linear Regressor: $R^2$ Over Time]{\includegraphics[width=0.45\textwidth]{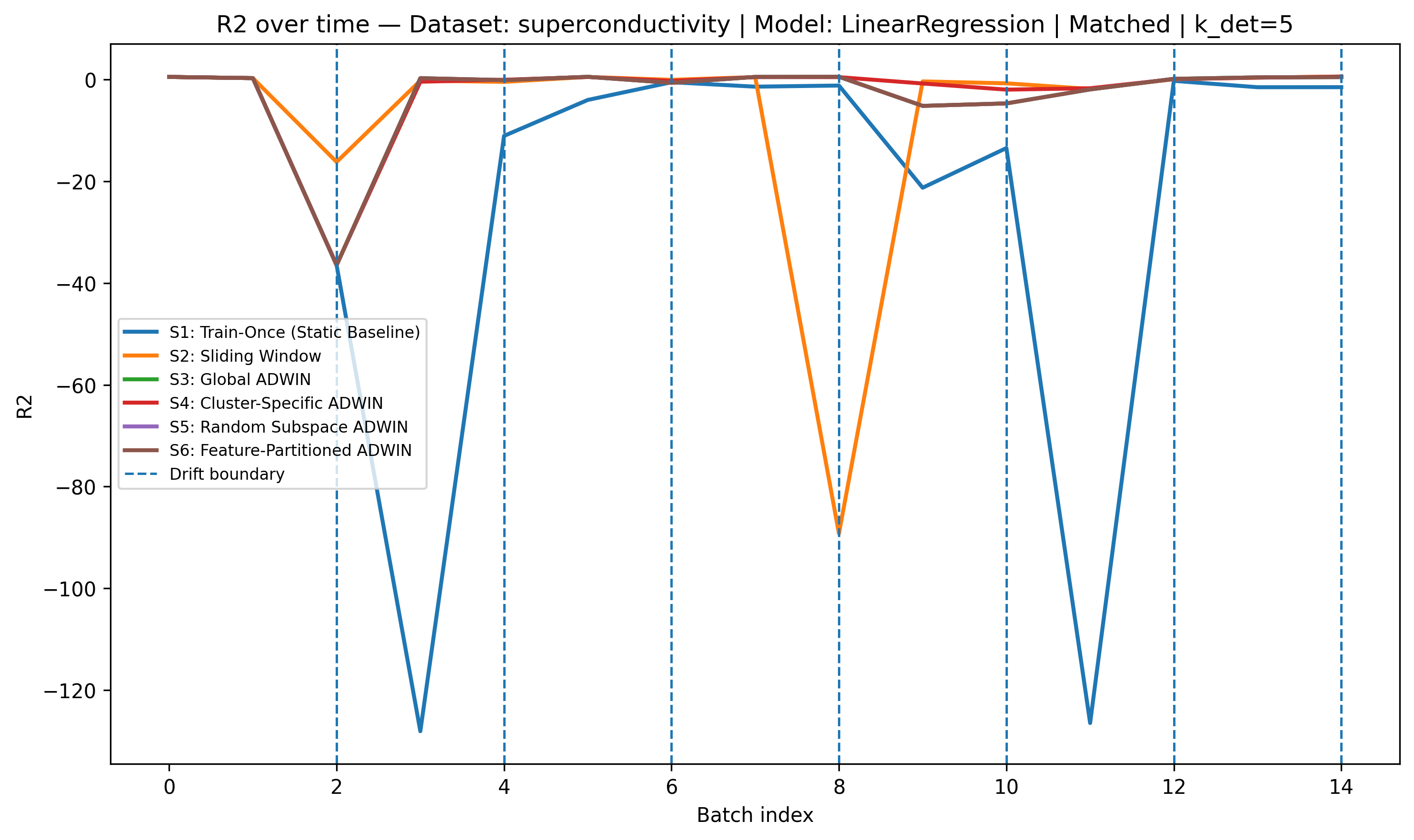}
    \label{fig:r2_time_LR_superconductivity}}
    \hfill
    \subfloat[Random Forest Regressor: $R^2$ Over Time]{\includegraphics[width=0.45\textwidth]{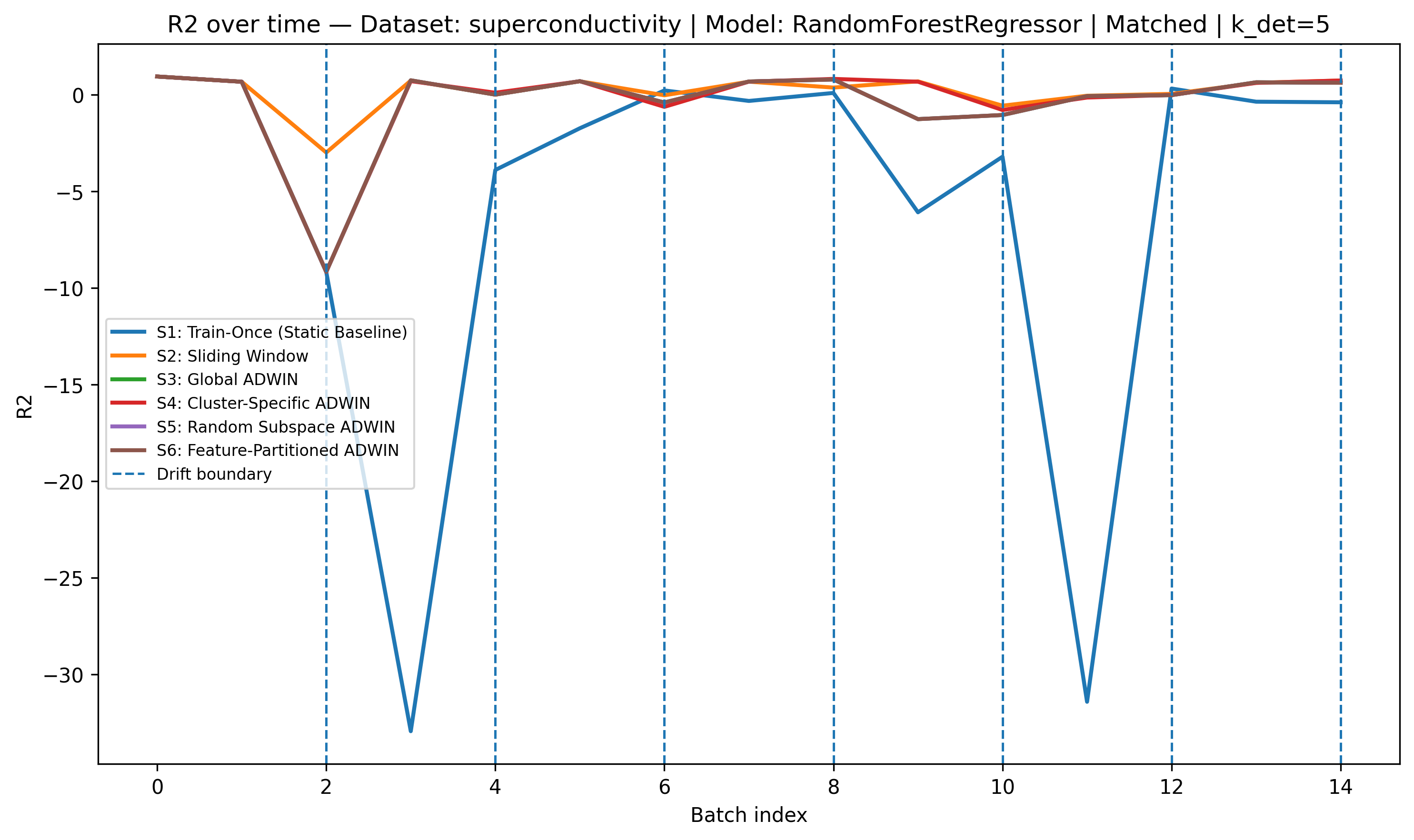}
    \label{fig:r2_time_RF_superconductivity}}
    \hfill
    \subfloat[XGBoost Regressor: $R^2$ Over Time]{\includegraphics[width=0.45\textwidth]{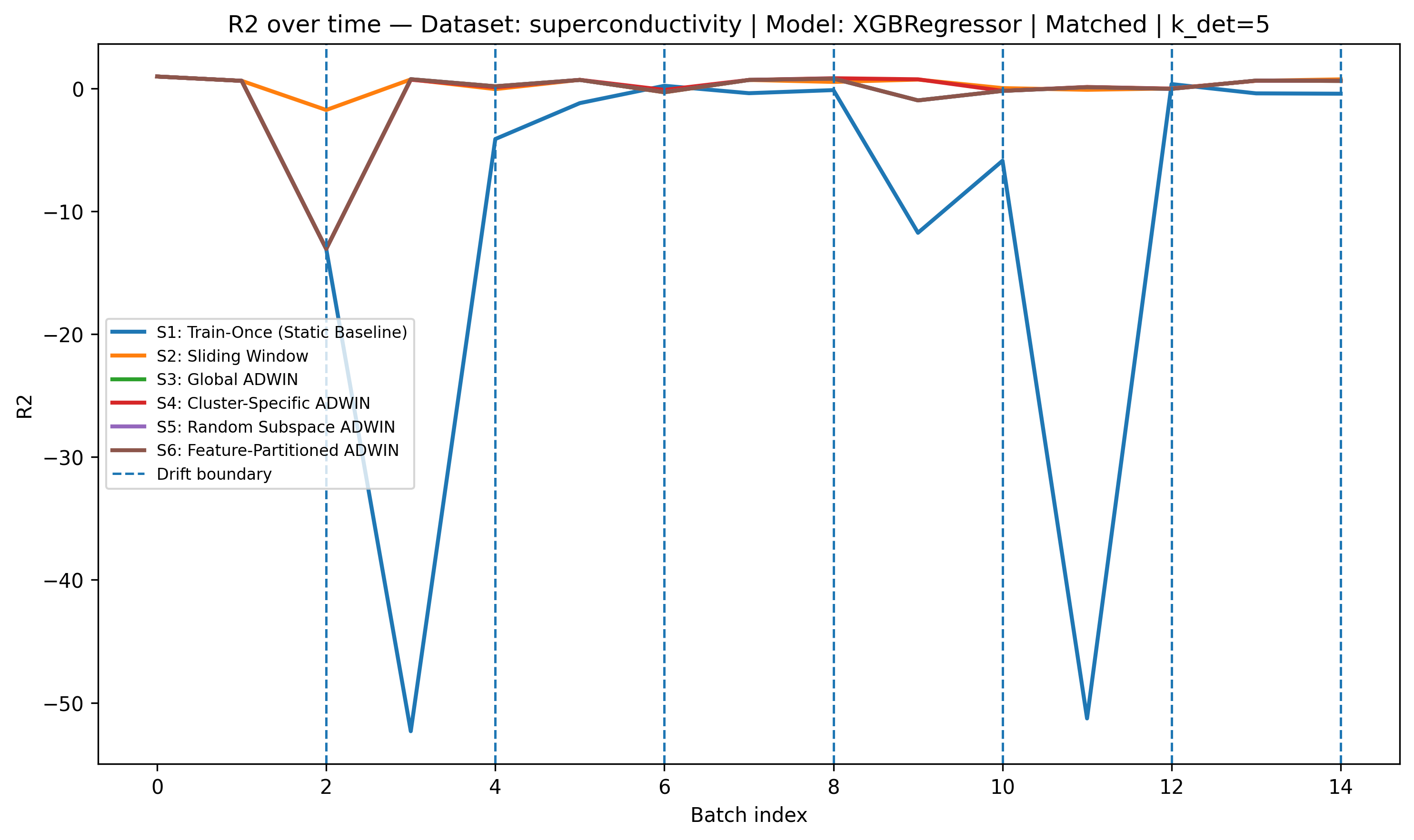}
    \label{fig:r2_time_XGB_superconductivity}}
    \hfill
    \subfloat[kNN Regression: $R^2$ Over Time]{\includegraphics[width=0.45\textwidth]{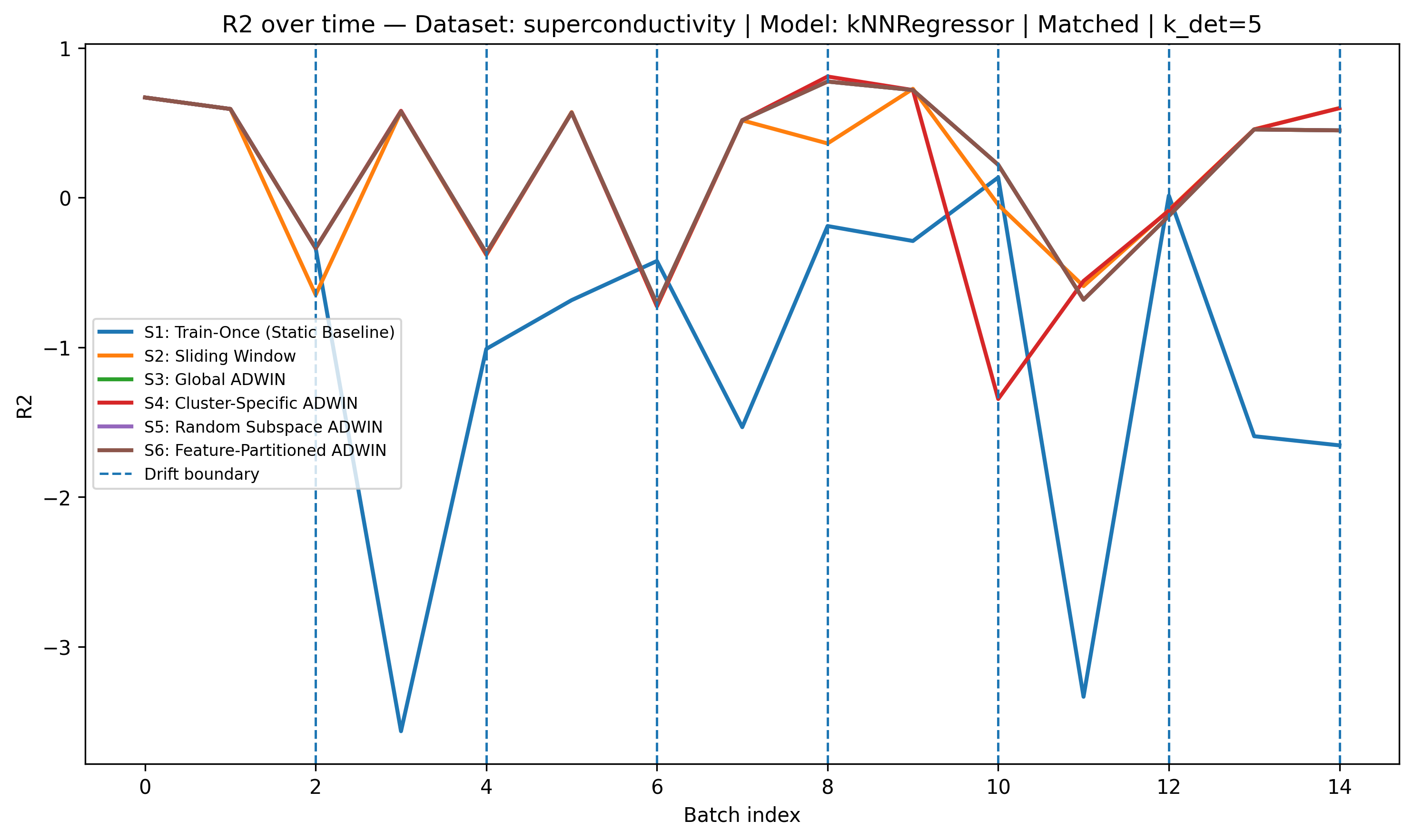}
    \label{fig:r2_time_KNN_superconductivity}}
    \caption{Performance and Adaptation Effort Analysis for the Superconductivity dataset across Strategies S1-S6.}
    \label{fig:super_grid}
\end{figure}
\par
\endgroup

The timeline for the Linear Regression (Figure \ref{fig:r2_time_LR_superconductivity}) shows that the non-adaptive baseline S1 exhibited severe instability throughout the stream, including multiple extreme negative $R^2$ degradations after major drift boundaries, demonstrating poor robustness to evolving data distributions.

In contrast, the adaptive strategies generally maintained far more stable predictive behaviour. S4 showed the strongest overall robustness, remaining consistently close to zero or slightly positive throughout most of the stream while avoiding the catastrophic degradations observed in S1. S3, S5, and S6 overlap almost entirely across the timeline, indicating nearly identical adaptation behaviour and predictive outcomes under the evaluated drift conditions. These strategies maintained relatively stable performance and consistently outperformed the static baseline.

Although S2 initially demonstrated competitive adaptation performance, it experienced a major instability around the middle of the stream, where a severe negative $R^2$ drop occurred before rapidly recovering in later batches.

The Random Forest timeline (Figure \ref{fig:r2_time_RF_superconductivity}) shows once again that non-adaptive strategy experienced severe instability throughout the stream, including multiple extreme negative $R^2$ drops after several drift boundaries, indicating poor resilience to evolving data distributions.

In contrast, the adaptive strategies maintained considerably more stable and controlled predictive behaviour. S2 consistently provided the strongest and most stable performance across the stream, remaining close to zero or slightly positive even after major drift events. S4 also demonstrated strong adaptation capability, closely tracking S2 throughout most batches while maintaining stable predictive performance.

Additionally, S3, S5, and S6 overlap almost entirely during the majority of the timeline, indicating highly similar adaptation behaviour and predictive outcomes under the evaluated drift conditions. Although these strategies showed small fluctuations at specific batches, they consistently avoided the severe degradations observed in the static baseline.

The timeline for the XGBoost Regressor (Figure \ref{fig:r2_time_XGB_superconductivity}) highlights a strong contrast between the non-adaptive baseline and the adaptive retraining strategies. The static baseline S1 exhibited severe instability, including multiple extreme negative $R^2$ drops after several drift boundaries, indicating substantial degradation under evolving data conditions. In contrast, all adaptive strategies maintained considerably more stable behaviour throughout the stream.

S2 consistently delivered the most stable overall performance, remaining close to zero or slightly positive across nearly all batches and rapidly recovering after drift events. S4 also demonstrated strong robustness, closely tracking S2 during most stages of the stream while maintaining stable predictive performance. Additionally, S3, S5, and S6 overlap almost entirely throughout the timeline, indicating nearly identical adaptation behaviour and predictive outcomes under the evaluated drift conditions. Although these strategies occasionally experienced small fluctuations, they consistently outperformed the static baseline by avoiding the severe degradations observed in S1.

For kNN (Figure \ref{fig:r2_time_KNN_superconductivity}), exhibited highly unstable behaviour, frequently producing large negative $R^2$ values after several drift boundaries, indicating poor adaptation to evolving data distributions. In contrast, S2, S4, and S6 maintained predominantly positive $R^2$ values across most of the stream and demonstrated stronger recovery following drift events. S4 achieved the highest peak performance during the middle stages of the stream, while S2 provided the most stable overall trajectory with fewer severe degradations. S3 and S5 overlap almost entirely with S6, indicating nearly identical adaptation behaviour and predictive performance.

Overall, the timeline analyses for the Superconductivity dataset across all evaluated regressors, demonstrates that the static strategy frequently suffered from severe negative $R^2$ degradations after drift boundaries, highlighting its inability to cope with evolving data distributions. In contrast, adaptive approaches maintained considerably more stable trajectories throughout the streams. Among the evaluated strategies, S2 and S4 repeatedly emerged as the strongest adaptation mechanisms, consistently providing the highest stability and most reliable recovery after drift events for XGBRegressor and RandomForestRegressor, while S4 showed particularly robust behaviour for LinearRegression. For kNNRegressor, adaptive retraining also produced substantial improvements in predictive stability and overall robustness compared with the static baseline.

\FloatBarrier

\begingroup
\subsection{Cluster-Mismatch Robustness and Sensitivity Results}\label{secRobustness}

Table~\ref{tab:combined_cluster_mismatch_results} presents the combined classification and regression robustness results for the proposed S4 (Cluster-Specific ADWIN) framework under varying detector mismatch conditions. The evaluation considers both cluster-count mismatch and centroid initialization mismatch between the synthetic stream generator and the online detector.

\begin{table*}[t]
\centering
\caption{Combined classification and regression robustness evaluation for S4 (Cluster-Specific ADWIN).}
\label{tab:combined_cluster_mismatch_results}

\renewcommand{\arraystretch}{1.2}

\resizebox{\textwidth}{!}{
\begin{tabular}{lccccccccc}
\hline
\textbf{Setting} &
\textbf{$k_g$} &
\textbf{$k_d$} &
\textbf{GSeed} &
\textbf{DSeed} &
\textbf{Avg F1} &
\textbf{Avg Acc.} &
\textbf{Avg RMSE} &
\textbf{Avg MAE} &
\textbf{Avg Adapt.} \\
\hline

Matched
& 5 & 5 & 42 & 42
& 0.6580 & 0.6803
& 21.96B
& 14.75B
& 3.34 \\

Centroid mismatch
& 5 & 5 & 42 & 7
& 0.6659 & 0.6834
& 20.95B
& 15.22B
& 3.20 \\

Under-cluster detector
& 5 & 3 & 42 & 42
& \textbf{0.6729}
& \textbf{0.6904}
& 22.15B
& 14.97B
& \textbf{2.55} \\

Over-cluster detector
& 5 & 10 & 42 & 42
& 0.6580
& 0.6798
& 23.21B
& 16.07B
& 3.52 \\

Centroids \& $k$ mismatch
& 5 & 3 & 42 & 7
& 0.6622
& 0.6797
& \textbf{19.73B}
& \textbf{13.00B}
& 2.70 \\

Different centroids + $k$ mismatch
& 5 & 10 & 42 & 7
& 0.6606
& 0.6830
& 21.65B
& 14.68B
& 3.23 \\

\hline
\end{tabular}
}
\end{table*}

Across both predictive tasks, the results demonstrate that S4 remains comparatively stable even when detector clusters do not perfectly align with the drift-generation structure. Performance degradation under mismatch conditions is generally limited, indicating that the localized ADWIN adaptation mechanism is robust to imperfect cluster boundaries and centroid uncertainty.

For classification, the strongest overall performance was obtained under the under-cluster detector configuration ($k_{\text{det}}=3$), which achieved the highest average F1 score ($0.6729$) and average accuracy ($0.6904$), while also producing the fewest adaptation events ($2.55$ on average). This suggests that moderate detector generalization can reduce unnecessary retraining while still preserving sensitivity to dominant drift regions.

The over-cluster detector configuration ($k_{\text{det}}=10$) produced slightly lower classification performance together with increased adaptation frequency. This behaviour is consistent with the expectation that excessive partitioning creates more localized detectors, which may become overly sensitive to smaller fluctuations within the feature space.

Regression behaviour follows a similar trend. The best RMSE and MAE values were obtained under the combined centroid and cluster-count mismatch configuration with $k_{\text{det}}=3$, indicating that S4 can maintain stable regression adaptation performance even under substantial detector mismatch. While the absolute regression error values vary across configurations, the differences remain relatively moderate considering the scale and heterogeneity of the evaluated regression datasets.

The centroid mismatch experiments further demonstrate that the framework does not depend on exact spatial alignment between generator and detector clusters. Even when detector centroids were initialized using different random seeds, predictive performance remained stable across both classification and regression metrics.

Overall, these results indicate that the proposed S4 framework generalizes beyond idealized clustering assumptions and remains effective under realistic conditions where drift regions and detector partitions are only approximately aligned.

\subsection{Adaptive Window Dynamics Across Clusters}\label{secWindowDynamics}
To further analyse the localized behaviour of the proposed S4 (Cluster-Specific ADWIN) framework, we visualized the adaptive ADWIN window dynamics for contrasting detector clusters within the same data stream. Specifically, each plot compares a relatively stable cluster against a high-drift cluster, showing how the adaptive window sizes evolve independently over time. This directly addresses the reviewer’s request to illustrate how S4 manages different feature-space regions with varying sensitivity levels.

The key intuition behind these plots is that ADWIN dynamically expands its window when the observed error distribution remains stable and rapidly contracts or resets when drift is detected. Consequently, stable regions accumulate progressively larger windows, whereas high-drift regions exhibit abrupt collapses corresponding to detected distribution shifts. Because S4 assigns an independent ADWIN detector to each cluster, these behaviours emerge locally rather than globally.

Figure~\ref{fig:s4_window_dynamics_grid} presents representative examples across both classification and regression tasks under several robustness configurations, including matched clustering, centroid mismatch, and combined centroid plus cluster-count mismatch settings.

\begin{figure*}[t]
\centering

\begin{minipage}[t]{0.32\textwidth}
    \centering
    \includegraphics[width=\linewidth]{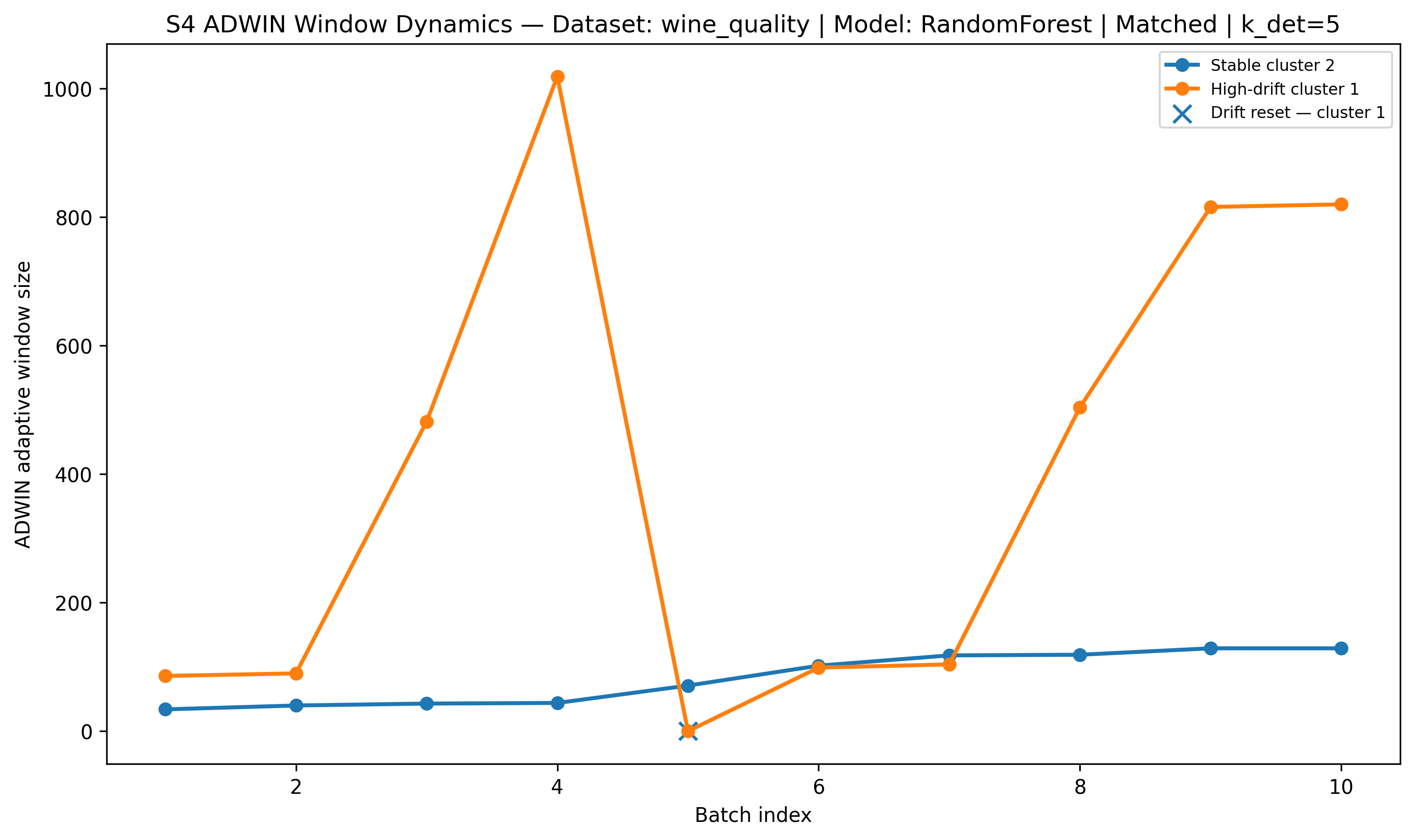}
    
    \vspace{0.2cm}
    \footnotesize
    \textbf{(a)} Wine Quality -- RandomForest\\
    Matched setting: $k_{gen}=5$, $k_{det}=5$
\end{minipage}
\hfill
\begin{minipage}[t]{0.32\textwidth}
    \centering
    \includegraphics[width=\linewidth]{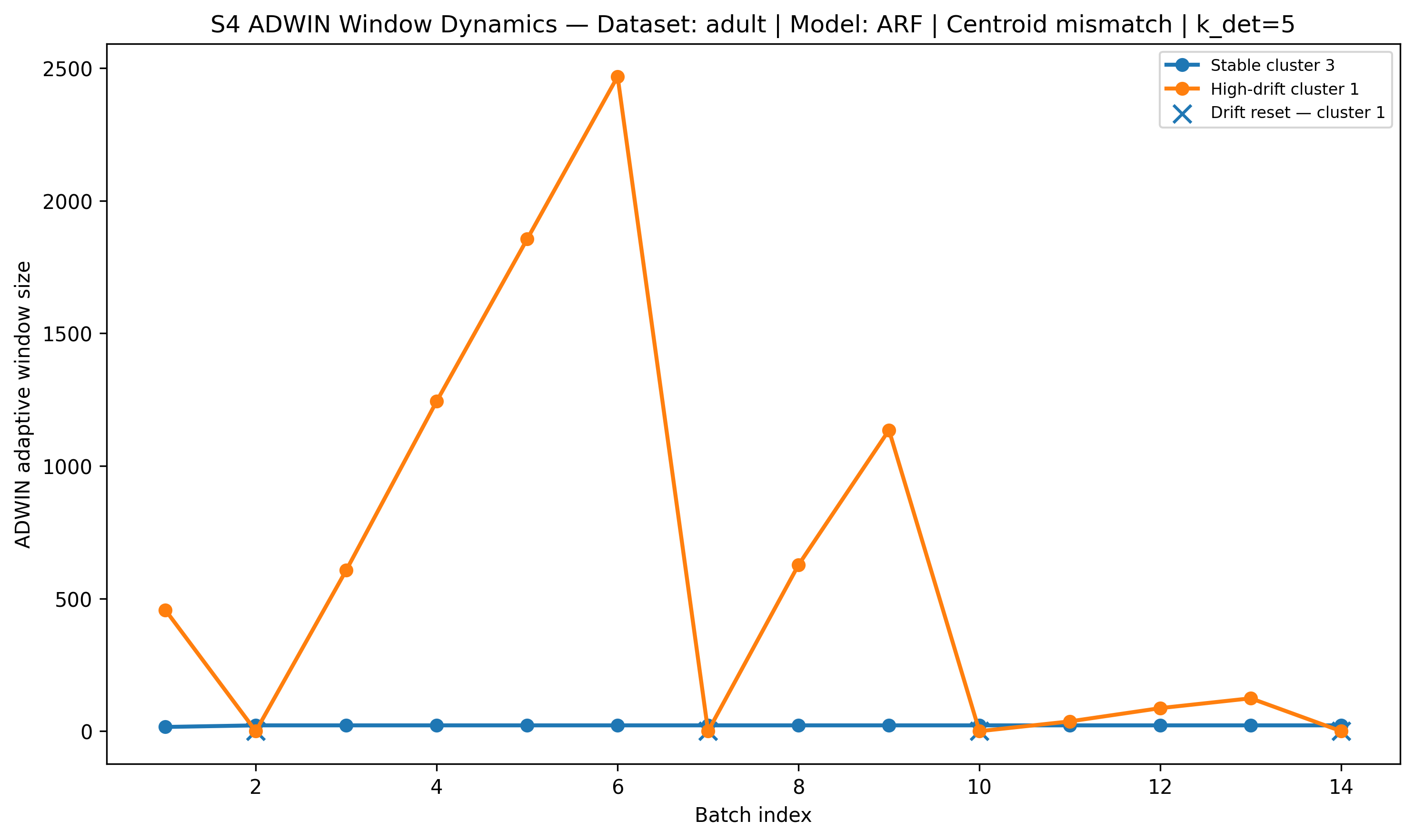}
    
    \vspace{0.2cm}
    \footnotesize
    \textbf{(b)} Adult -- ARF\\
    Centroid mismatch: $k_{gen}=5$, $k_{det}=5$
\end{minipage}
\hfill
\begin{minipage}[t]{0.32\textwidth}
    \centering
    \includegraphics[width=\linewidth]{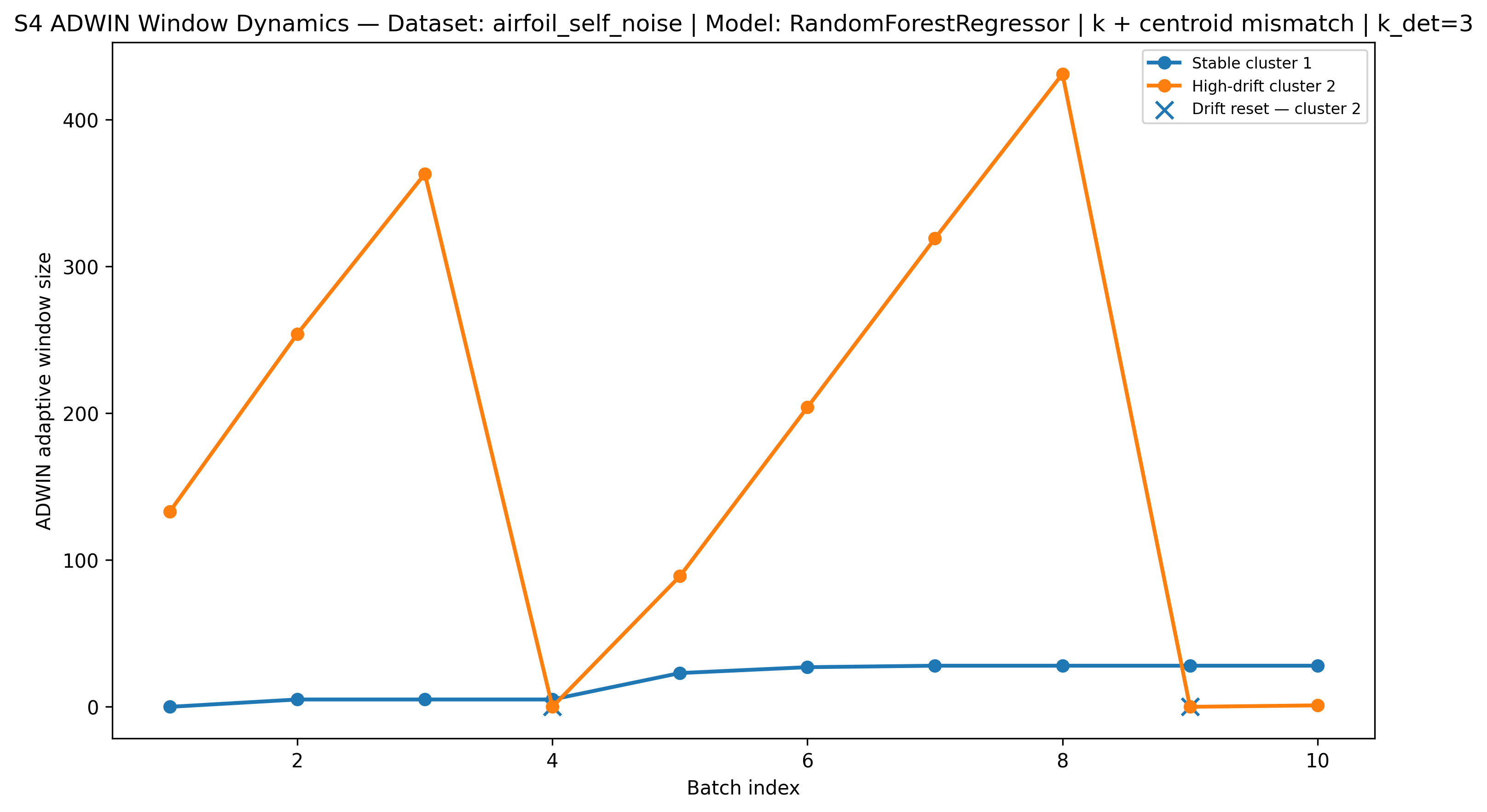}
    
    \vspace{0.2cm}
    \footnotesize
    \textbf{(c)} Airfoil Self Noise -- RFRegressor\\
    $k$ + centroid mismatch: $k_{gen}=5$, $k_{det}=3$
\end{minipage}

\vspace{-0.1cm}

\begin{minipage}[t]{0.32\textwidth}
    \centering
    \includegraphics[width=\linewidth]{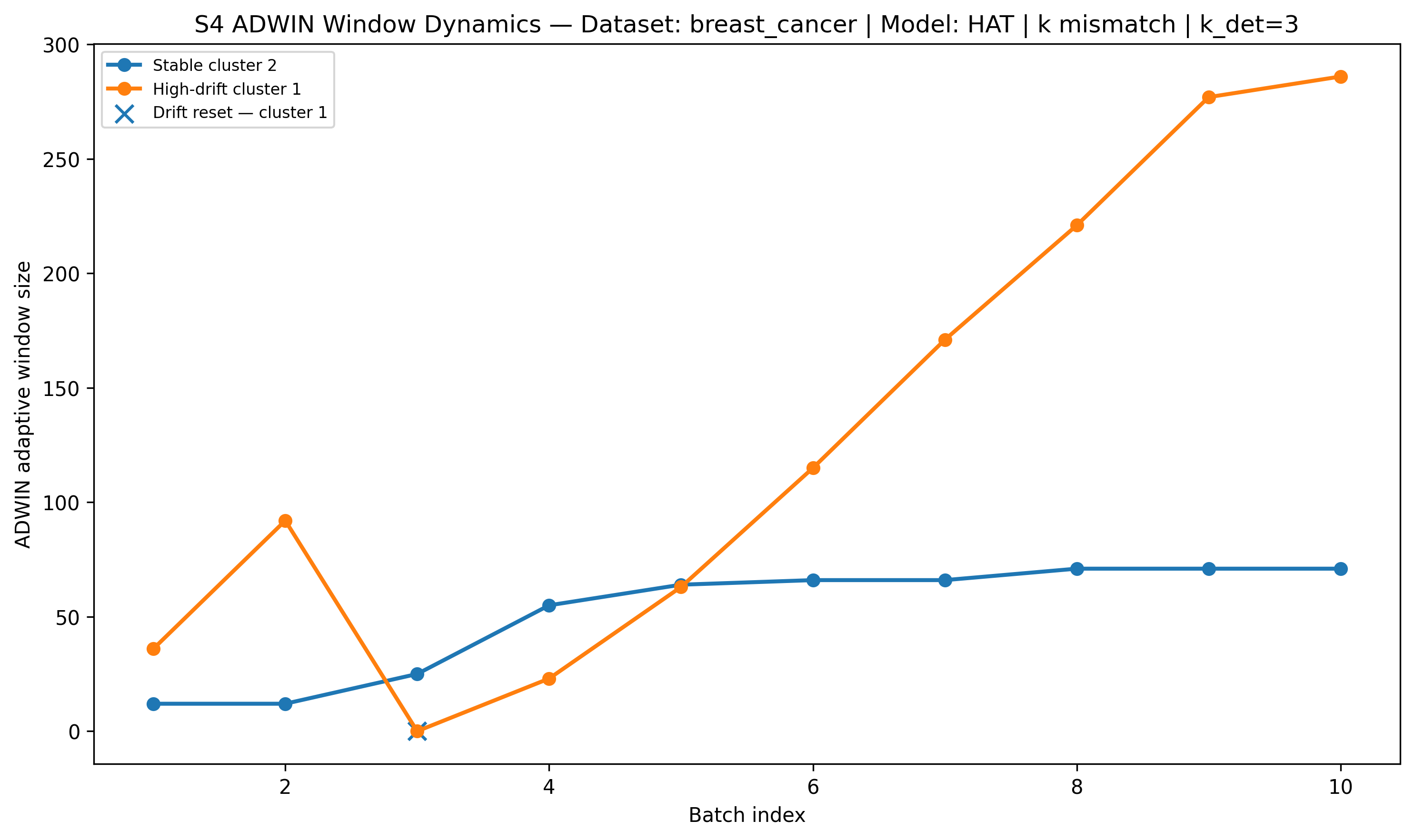}
    
    \vspace{0.2cm}
    \footnotesize
    \textbf{(d)} Breast Cancer -- HAT\\
    $k$ mismatch: $k_{gen}=5$, $k_{det}=3$
\end{minipage}
\hfill
\begin{minipage}[t]{0.32\textwidth}
    \centering
    \includegraphics[width=\linewidth]{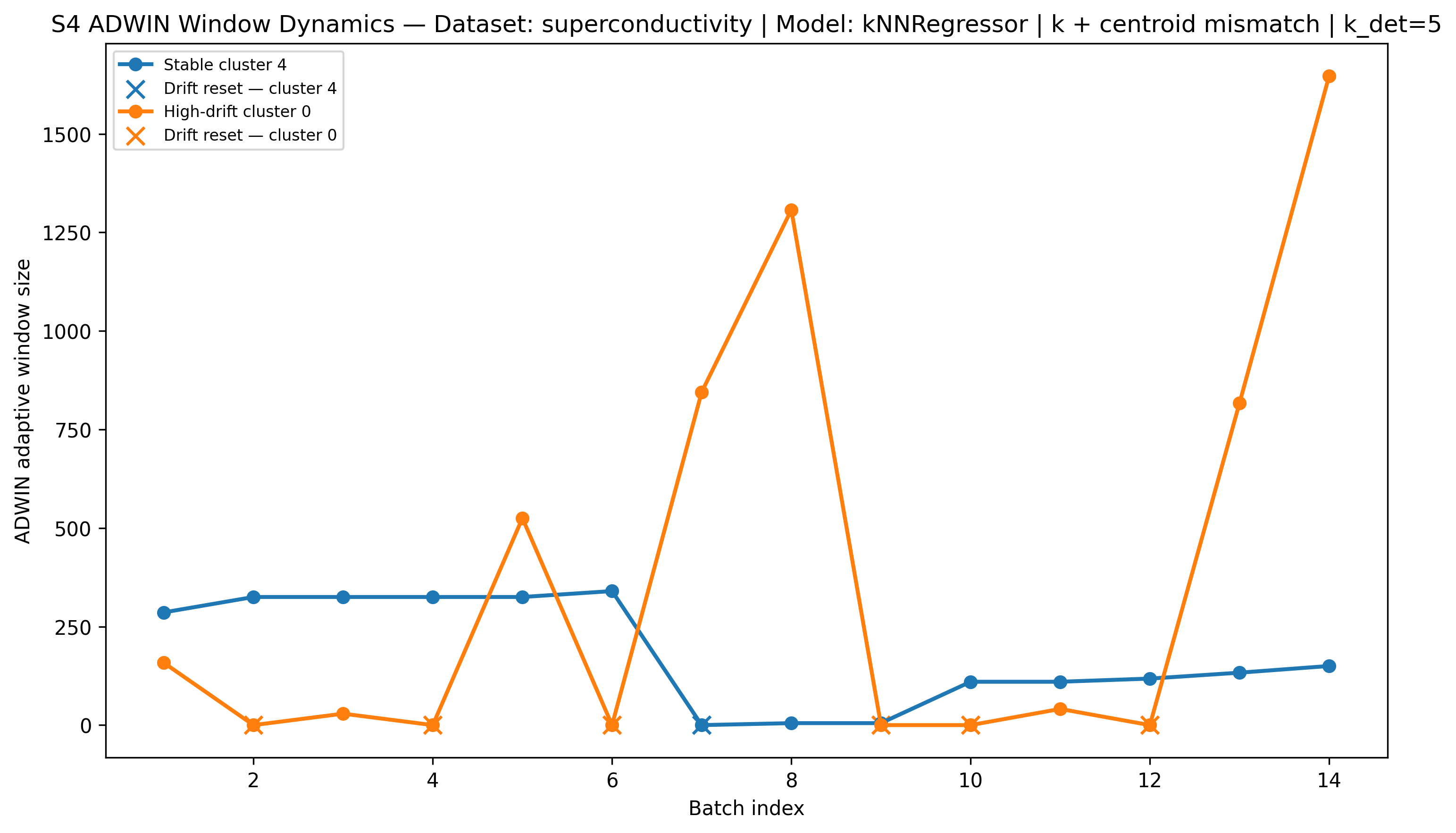}
    
    \vspace{0.2cm}
    \footnotesize
    \textbf{(e)} Superconductivty data -- kNNRegressor\\
    $k$ + centroid mismatch: $k_{gen}=5$, $k_{det}=5$
\end{minipage}

\caption{
Localized ADWIN adaptive window dynamics for stable and high-drift clusters under different robustness configurations. Sudden window collapses correspond to detected drift resets, while steadily increasing windows indicate stable feature-space regions.
}
\label{fig:s4_window_dynamics_grid}

\end{figure*}

Figure~\ref{fig:s4_window_dynamics_grid}(a) illustrates the matched configuration on the Wine Quality dataset using Random Forest with $k_{gen}=5$ and $k_{det}=5$. The stable cluster exhibits gradual and consistent window growth, indicating relatively stationary local behaviour. In contrast, the high-drift cluster shows a sharp reset near batch 5, where the adaptive window collapses from over $1000$ samples to near zero. This demonstrates S4’s ability to isolate localized drift events while preserving stable memory in unaffected regions.

Figure~\ref{fig:s4_window_dynamics_grid}(b) shows the Adult dataset using ARF under centroid mismatch conditions ($k_{gen}=5$, $k_{det}=5$ with different centroid seeds). Despite the detector clusters being spatially misaligned with the generator clusters, the high-drift region still exhibits multiple abrupt resets, whereas the stable cluster maintains a comparatively small but consistent adaptive window. This indicates that S4 remains sensitive to localized drift even when cluster boundaries are imperfectly aligned.

Figure~\ref{fig:s4_window_dynamics_grid}(c) presents a regression example on the Airfoil Self Noise dataset using RandomForestRegressor under combined centroid and cluster-count mismatch ($k_{gen}=5$, $k_{det}=3$). The high-drift cluster repeatedly expands and collapses, showing recurring shift detection cycles, while the stable cluster maintains a relatively smooth and monotonic increase. This behaviour suggests that even under under-clustered detector configurations, S4 can still distinguish stable and unstable feature-space regions.

Figure~\ref{fig:s4_window_dynamics_grid}(d) shows the Breast Cancer dataset using the Hoeffding Adaptive Tree (HAT) under cluster-count mismatch ($k_{gen}=5$, $k_{det}=3$). The high-drift cluster demonstrates progressive growth interrupted by an early reset event, after which the detector rebuilds its adaptive window. Meanwhile, the stable cluster exhibits gradual stabilization with minimal disruption. This highlights that localized adaptation remains effective even for fully online streaming tree models.

Finally, Figure~\ref{fig:s4_window_dynamics_grid}(e) illustrates the Superconductivity dataset using kNNRegressor under combined centroid mismatch conditions. The high-drift cluster undergoes multiple aggressive resets and strong oscillatory behaviour, reflecting substantial local non-stationarity. In contrast, the stable cluster accumulates substantially larger windows over time despite occasional contractions. This example demonstrates how S4 dynamically allocates sensitivity according to the local behaviour of each region rather than enforcing a single global adaptation policy.

Overall, these visualizations provide direct evidence that the proposed S4 framework performs localized drift management rather than relying on a single global detector state. Different regions of the feature space evolve independently, with stable regions preserving long adaptive memory while high-drift regions trigger frequent resets and retraining. This behaviour explains why S4 can maintain competitive predictive performance while avoiding unnecessary global adaptation across the entire stream.

\subsection{Spatial Decomposition of Drift Adaptation}\label{secRetrainingEffort}

In this section, we analyse the cumulative retraining effort attributed to individual detector clusters in the localized adaptation dynamics of the proposed S4 (Cluster-Specific ADWIN) strategy. This investigation addresses the spatial decomposition of drift, and making a critical consideration for understanding how adaptation effort is distributed across the feature space.

Unlike global strategies that trigger uniform retraining, S4 maintains independent ADWIN instances for each cluster. Consequently, only clusters experiencing statistically significant distributional shifts trigger retraining events. This mechanism induces a spatially heterogeneous retraining profile, isolating adaptation to unstable manifold regions while preserving the state of stable regions.

Figures~\ref{fig:s4_retraining_effort_adult} and~\ref{fig:s4_retraining_effort_superconductivity} present cumulative retraining-effort heatmaps for the Adult and Superconductivity datasets, respectively selected due to their complexity and covering both classification and regression tasks. In these visualizations, columns represent specific robustness configurations (predictive model and cluster count $k$) that have been encoded for better readability, and their mapping tables are displayed on Tables \ref{tab:adult_config_mapping_compact} and \ref{tab:superconductivity_config_mapping_compact}, while rows correspond to detector cluster identifiers. The intensity indicates the cumulative volume of retraining samples processed by each cluster over the stream duration.

Analysis of these heatmaps yields three primary insights:

\begin{enumerate}
    \item \textbf{Effort Sparsity:} Retraining is highly concentrated within a sparse subset of clusters. Across both datasets, large partitions of the feature space remain near-zero throughout the stream. This validates our core hypothesis: drift is often regionally bounded, and localized detection prevents redundant global updates.
    
    \item \textbf{Robustness to Mismatch:} While the "matched" setting ($k_{gen}=k_{det}$) shows the sharpest localization, the framework maintains effectiveness under centroid and $k$-mismatch configurations. Although mismatch causes adaptation effort to become more diffuse as drift "bleeds" across cluster boundaries, the profile remains non-uniform. S4 thus demonstrates structural resilience to imperfect partitioning.
    
    \item \textbf{Dataset-Specific Dynamics:} The Adult dataset exhibits extreme localization, where specific clusters dominate the adaptation budget across various model configurations. In contrast, the Superconductivity dataset displays a more distributed profile—likely due to the higher structural complexity of the regression task—yet still maintains distinct "stable zones" with minimal retraining.
\end{enumerate}

These results provide empirical evidence that S4 successfully achieves spatial drift decomposition. By selectively allocating adaptation budget to unstable regions, the framework minimizes unnecessary computational overhead, explaining the efficiency gains observed in the aggregate metrics previously reported.

\begin{figure*}[t]
    \centering
    \includegraphics[width=\textwidth]{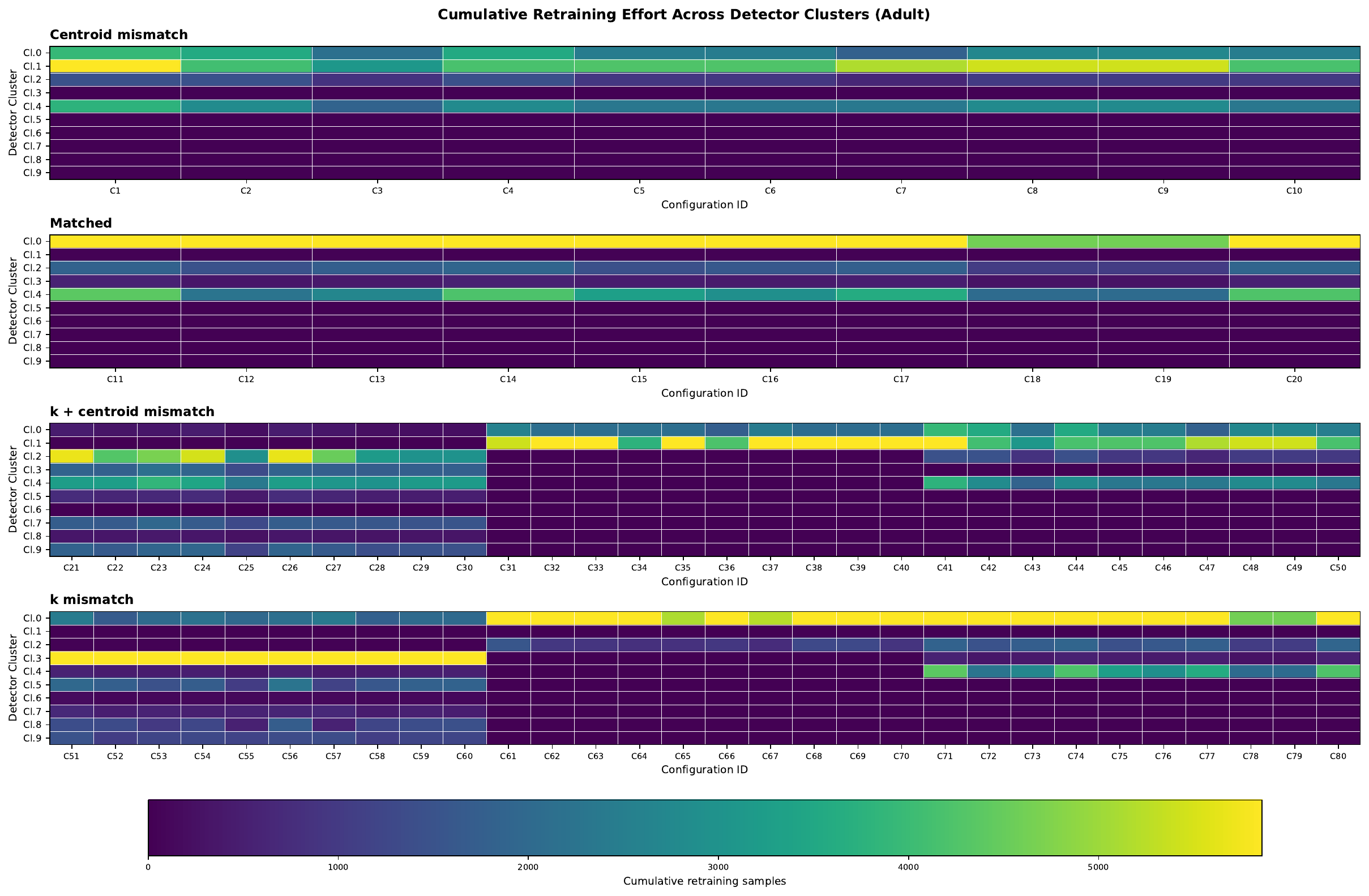}
    \caption{Cumulative retraining effort per cluster (Adult dataset). Bright regions denote clusters dominating adaptation activity; dark regions indicate stable feature-space partitions requiring minimal intervention.}
    \label{fig:s4_retraining_effort_adult}
\end{figure*}

\begin{table*}[t]
\centering
\caption{Compact configuration mapping for the Adult dataset heatmap. Within each row, configuration IDs follow the model order shown in the final column.}
\label{tab:adult_config_mapping_compact}
\footnotesize
\begin{tabular}{@{}p{0.14\textwidth}p{0.30\textwidth}p{0.08\textwidth}p{0.40\textwidth}@{}}
\toprule
\textbf{Config. IDs} & \textbf{Robustness setting} & \textbf{$k_{\mathrm{det}}$} & \textbf{Model order} \\
\midrule
C1--C10 & Centroid mismatch & 5 & ARF, HAT, LogisticRegression, OnlineKNNClassifier, OnlineLogisticRegression, OnlineNaiveBayes, OnlineSoftmaxRegression, RandomForest, XGBoost, kNN \\
C11--C20 & Matched & 5 & ARF, HAT, LogisticRegression, OnlineKNNClassifier, OnlineLogisticRegression, OnlineNaiveBayes, OnlineSoftmaxRegression, RandomForest, XGBoost, kNN \\
C21--C30 & k + centroid mismatch & 10 & ARF, HAT, LogisticRegression, OnlineKNNClassifier, OnlineLogisticRegression, OnlineNaiveBayes, OnlineSoftmaxRegression, RandomForest, XGBoost, kNN \\
C31--C40 & k + centroid mismatch & 3 & ARF, HAT, LogisticRegression, OnlineKNNClassifier, OnlineLogisticRegression, OnlineNaiveBayes, OnlineSoftmaxRegression, RandomForest, XGBoost, kNN \\
C41--C50 & k + centroid mismatch & 5 & ARF, HAT, LogisticRegression, OnlineKNNClassifier, OnlineLogisticRegression, OnlineNaiveBayes, OnlineSoftmaxRegression, RandomForest, XGBoost, kNN \\
C51--C60 & k mismatch & 10 & ARF, HAT, LogisticRegression, OnlineKNNClassifier, OnlineLogisticRegression, OnlineNaiveBayes, OnlineSoftmaxRegression, RandomForest, XGBoost, kNN \\
C61--C70 & k mismatch & 3 & ARF, HAT, LogisticRegression, OnlineKNNClassifier, OnlineLogisticRegression, OnlineNaiveBayes, OnlineSoftmaxRegression, RandomForest, XGBoost, kNN \\
C71--C80 & k mismatch & 5 & ARF, HAT, LogisticRegression, OnlineKNNClassifier, OnlineLogisticRegression, OnlineNaiveBayes, OnlineSoftmaxRegression, RandomForest, XGBoost, kNN \\
\bottomrule
\end{tabular}
\end{table*}
\FloatBarrier

\begin{figure*}[t]
    \centering
    \includegraphics[width=\textwidth]{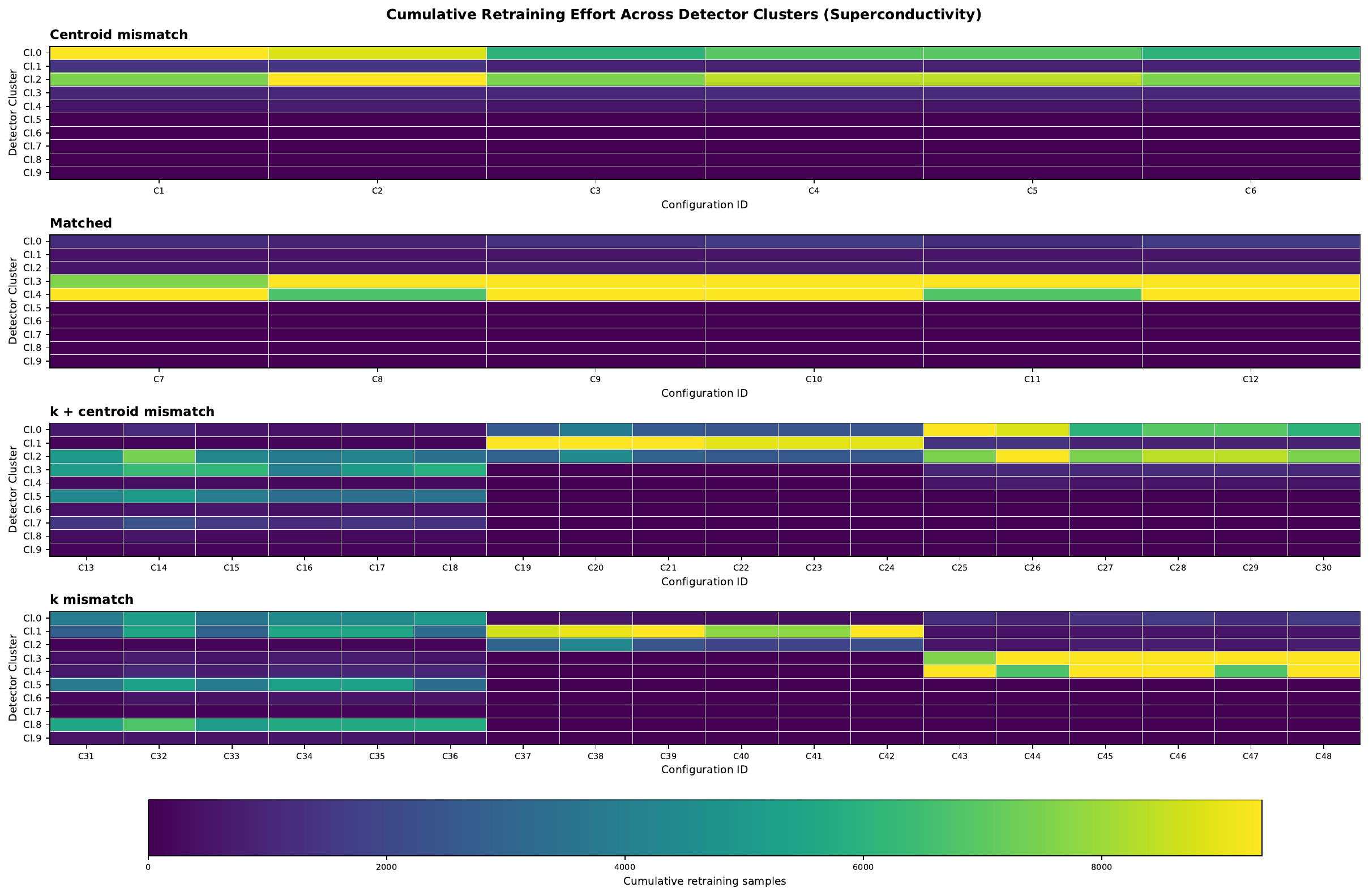}
    \caption{Cumulative retraining effort per cluster (Superconductivity dataset). The localized concentration confirms that S4 selectively allocates resources to non-stationary regions, even in high-dimensional regression tasks.}
    \label{fig:s4_retraining_effort_superconductivity}
\end{figure*}

\begin{table*}[t]
\centering
\caption{Compact configuration mapping for the Superconductivity dataset heatmap. Within each row, configuration IDs follow the model order shown in the final column.}
\label{tab:superconductivity_config_mapping_compact}
\footnotesize
\begin{tabular}{@{}p{0.14\textwidth}p{0.30\textwidth}p{0.08\textwidth}p{0.40\textwidth}@{}}
\toprule
\textbf{Config. IDs} & \textbf{Robustness setting} & \textbf{$k_{\mathrm{det}}$} & \textbf{Model order} \\
\midrule
C1--C6 & Centroid mismatch & 5 & ARFRegressor, HoeffdingTreeRegressor, LinearRegression, RandomForestRegressor, XGBRegressor, kNNRegressor \\
C7--C12 & Matched & 5 & ARFRegressor, HoeffdingTreeRegressor, LinearRegression, RandomForestRegressor, XGBRegressor, kNNRegressor \\
C13--C18 & k + centroid mismatch & 10 & ARFRegressor, HoeffdingTreeRegressor, LinearRegression, RandomForestRegressor, XGBRegressor, kNNRegressor \\
C19--C24 & k + centroid mismatch & 3 & ARFRegressor, HoeffdingTreeRegressor, LinearRegression, RandomForestRegressor, XGBRegressor, kNNRegressor \\
C25--C30 & k + centroid mismatch & 5 & ARFRegressor, HoeffdingTreeRegressor, LinearRegression, RandomForestRegressor, XGBRegressor, kNNRegressor \\
C31--C36 & k mismatch & 10 & ARFRegressor, HoeffdingTreeRegressor, LinearRegression, RandomForestRegressor, XGBRegressor, kNNRegressor \\
C37--C42 & k mismatch & 3 & ARFRegressor, HoeffdingTreeRegressor, LinearRegression, RandomForestRegressor, XGBRegressor, kNNRegressor \\
C43--C48 & k mismatch & 5 & ARFRegressor, HoeffdingTreeRegressor, LinearRegression, RandomForestRegressor, XGBRegressor, kNNRegressor \\
\bottomrule
\end{tabular}
\end{table*}
\FloatBarrier

\subsection{Training Time Analysis Across Benchmarking Strategies}
\label{subsec:training_time_analysis}

In addition to predictive performance and adaptation effectiveness, the computational overhead introduced by each retraining strategy was also evaluated by measuring the mean update training time across all experiments. Figure~\ref{fig:training_time_by_strategy} summarises the average update time required by each benchmarking method.

The results reveal substantial differences in computational cost between the evaluated strategies. As expected, the static baseline strategy (S1) incurred virtually no retraining overhead since no model updates were performed after the initial training phase. Among the adaptive approaches, S6 (Feature-Partitioned ADWIN) and S5 (Random Subspace ADWIN) achieved the lowest average update times, both requiring approximately 0.1 seconds per update. S3 (Global ADWIN) introduced a slightly higher computational cost, averaging roughly 0.16 seconds per retraining operation.

More computationally demanding strategies were observed for S4 (Cluster-Specific ADWIN) and especially S2 (Sliding Window). S4 required approximately 0.35 seconds on average per update due to the additional clustering and localized drift handling operations performed during adaptation. In contrast, S2 exhibited the highest computational overhead by a large margin, exceeding 1 second of average update time. This behaviour is expected because the sliding-window strategy performs full retraining operations using substantially larger subsets of recent data at every detected adaptation point.

These results highlight an important trade-off between predictive robustness and computational efficiency. While S2 frequently achieved the strongest predictive performance across both classification and regression tasks, it also imposed the highest computational burden. Conversely, S5 and S6 provided considerably lower update costs while still maintaining competitive adaptation performance in several datasets. Therefore, the choice of adaptation strategy should consider not only predictive quality but also the computational constraints and latency requirements of the target streaming environment.

\begin{figure}[ht]
    \centering
    \includegraphics[width=0.9\linewidth]{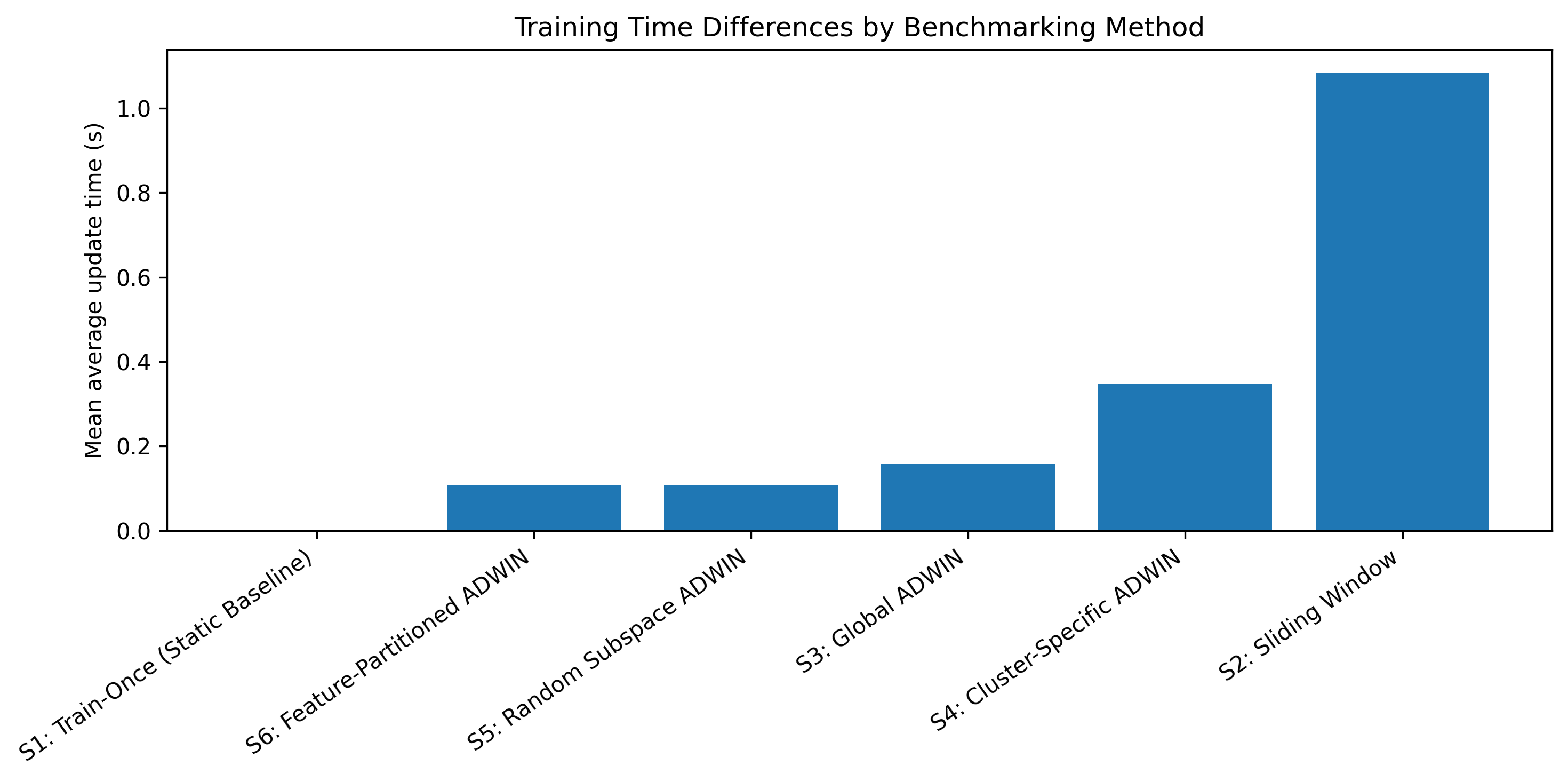}
    \caption{Mean update training time across benchmarking strategies.}
    \label{fig:training_time_by_strategy}
\end{figure}

\subsubsection{Classification Training Time Analysis}
\label{subsubsec:classification_training_time}

Figure~\ref{fig:classification_training_heatmap} presents the mean update training time for the classification experiments across all robustness settings and benchmarking strategies. The heatmap highlights both the consistency and computational variability introduced by the different adaptation mechanisms under distinct drift conditions.

The results show that the static baseline strategy (S1) maintained negligible update cost across all robustness settings, as no retraining operations were executed after the initial model fitting phase. In contrast, S2 (Sliding Window) consistently exhibited the highest computational overhead in every evaluated scenario, with average update times exceeding 1 second in most settings. The highest costs were observed under the matched and centroid mismatch configurations, indicating that repeated full-window retraining introduces substantial computational demands regardless of the drift configuration.

Among the ADWIN-based approaches, S4 (Cluster-Specific ADWIN) showed the second-highest training cost, with moderate variability across robustness settings. This behaviour reflects the additional computational complexity introduced by clustering operations and localized adaptation mechanisms. The computational overhead of S4 remained significantly lower than S2, but consistently higher than the remaining ADWIN-based approaches.

S3 (Global ADWIN) demonstrated relatively stable and moderate update times across all settings, indicating that global drift monitoring introduces only limited additional overhead. Meanwhile, S5 (Random Subspace ADWIN) and S6 (Feature-Partitioned ADWIN) consistently achieved the lowest adaptive retraining costs among the dynamic approaches. Their training times remained close to the baseline across all robustness configurations, suggesting that lightweight retraining mechanisms can substantially reduce computational expense while still supporting adaptive behaviour.

Overall, the classification training-time analysis reveals a clear trade-off between adaptation complexity and computational efficiency. Strategies relying on large-scale retraining operations, such as S2, achieved strong predictive adaptation performance but incurred the highest update costs. Conversely, S5 and S6 provided considerably lower computational overhead, making them more suitable for resource-constrained or real-time streaming environments where rapid adaptation and low latency are critical requirements.

\begin{figure}[ht]
    \centering
    \includegraphics[width=\linewidth]{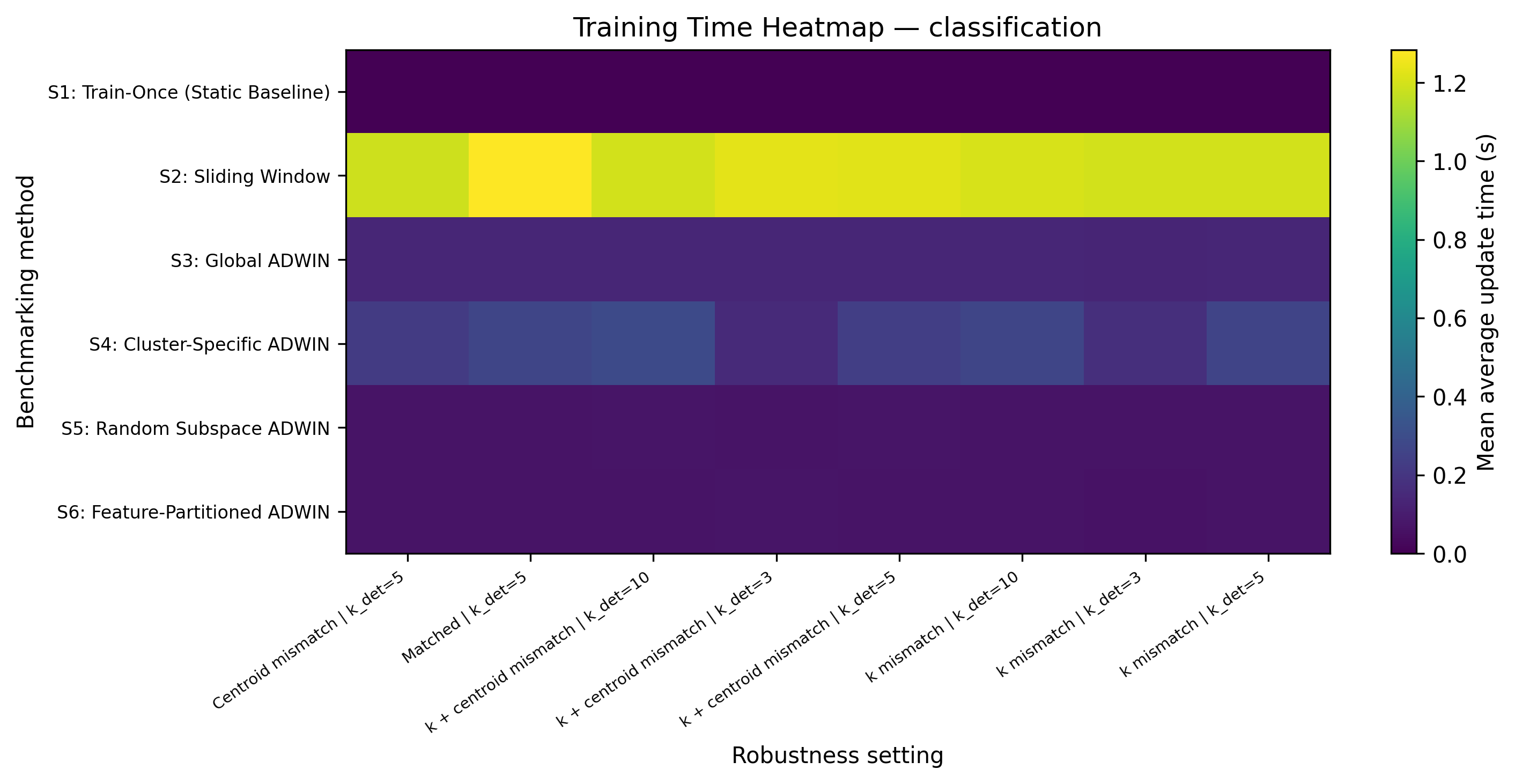}
    \caption{Heatmap of mean update training times for classification tasks across robustness settings and benchmarking strategies.}
    \label{fig:classification_training_heatmap}
\end{figure}

\subsubsection{Regression Training Time Analysis}
\label{subsubsec:regression_training_time}

Figure~\ref{fig:regression_training_heatmap} presents the mean update training times for the regression experiments across all robustness settings and benchmarking strategies. Similar to the classification scenario, the results reveal clear differences in computational overhead between lightweight adaptive approaches and strategies relying on extensive retraining operations.

The static baseline strategy (S1) again maintained negligible update cost across all robustness configurations, as no online retraining was performed after initialization. In contrast, S2 (Sliding Window) consistently produced the highest computational overhead among all evaluated methods, with update times remaining close to 0.75 seconds across nearly every robustness setting. This behaviour confirms that repeated full-window retraining imposes a substantial computational burden in continuous regression streams.

S4 (Cluster-Specific ADWIN) exhibited the second-highest training cost and also showed the greatest variability across robustness settings. In several configurations, particularly the matched and \(k\)-mismatch scenarios, S4 reached computational costs comparable to S2. However, under other settings such as \(k\)-plus-centroid mismatch, the training overhead decreased considerably, suggesting that cluster-localized retraining can adapt its computational demand depending on the drift structure encountered in the stream.

The remaining ADWIN-based strategies—S3 (Global ADWIN), S5 (Random Subspace ADWIN), and S6 (Feature-Partitioned ADWIN)—demonstrated substantially lower and highly stable update times across all robustness settings. Their computational overhead remained moderate and relatively uniform throughout the experiments, indicating that these methods achieve adaptive behaviour without incurring the large retraining costs observed in S2 and S4.

Overall, the regression training-time analysis highlights a similar trade-off to that observed in the classification experiments. Strategies emphasizing aggressive retraining, particularly S2 and partially S4, achieved stronger adaptive capabilities at the expense of significantly higher computational cost. Conversely, S3, S5, and S6 maintained considerably lower and more stable training times, making them more suitable for real-time regression stream applications where computational efficiency and scalability are essential.

\begin{figure}[ht]
    \centering
    \includegraphics[width=\linewidth]{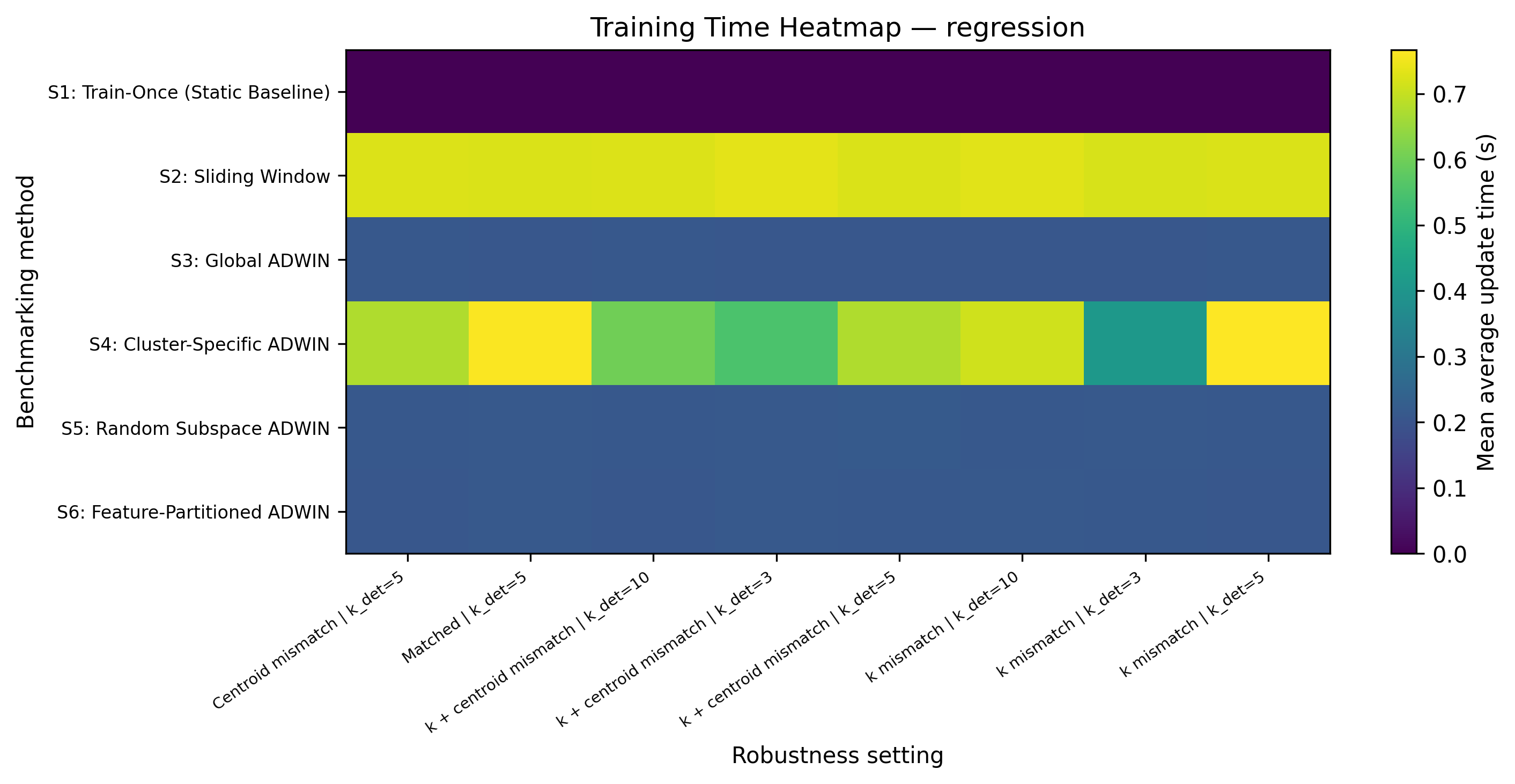}
    \caption{Heatmap of mean update training times for regression tasks across robustness settings and benchmarking strategies.}
    \label{fig:regression_training_heatmap}
\end{figure}
\par
\endgroup

\section{Discussion}\label{secDiscussion}

The experimental results obtained from the evaluated classification and regression benchmark datasets provide a broad perspective on the behaviour of cluster-local drift adaptation under heterogeneous drift conditions, varying feature spaces, and different model families. The analysed datasets include three classification tasks (Adult, Wine Quality, and Breast Cancer) together with two regression benchmarks (Airfoil Self-Noise and Superconductivty data). Across these settings, several consistent patterns emerge, particularly regarding retraining efficiency, robustness to drift, and the interaction between adaptation strategies and model inductive bias.

\subsection{Comparative Behaviour of the Evaluated Adaptation Strategies}

The experimental evaluation revealed substantial differences between the six evaluated adaptation strategies with respect to predictive robustness, computational efficiency, and adaptation responsiveness under controlled distribution shift conditions. Across both classification and regression tasks, the results consistently demonstrated that no single strategy simultaneously optimized predictive performance, update frequency, and retraining cost. Instead, the obtained outcomes highlight a clear trade-off between adaptation aggressiveness and operational efficiency.

The static baseline S1 consistently produced the lowest computational overhead because no retraining operations were performed after initialization. However, its predictive performance progressively degraded as distributional changes accumulated throughout the stream. This behaviour confirms the expected limitations of static learners under evolving environments and establishes S1 as a lower-bound reference for adaptation performance.

At the opposite extreme, S2 (Sliding-Window Retraining) consistently achieved the strongest predictive metrics across several datasets and model families. The strategy maintained high Accuracy, F1-score, and regression performance because the model was continuously refreshed using the most recent observations. However, these gains came at a substantial operational cost. S2 produced the highest cumulative retraining effort, the largest number of adaptation events, and the highest update training times across nearly all robustness settings. Consequently, the results suggest that S2 functions primarily as an upper-bound responsiveness benchmark rather than a practically efficient deployment strategy for long-running streaming systems.

The global drift-detection strategy S3 (Global ADWIN) achieved a substantially lower computational burden than S2 while still preserving competitive predictive performance in several scenarios. By relying on a single global detector, S3 reduced unnecessary retraining operations and maintained stable adaptation behaviour across most datasets. However, the results also indicate that global monitoring may dilute localized distributional changes, particularly under heterogeneous drift conditions where only specific regions of the feature space are affected.

The proposed S4 (Cluster-Specific ADWIN) consistently achieved the strongest overall balance between predictive robustness and adaptation efficiency. Across both classification and regression experiments, S4 maintained competitive predictive performance while substantially reducing retraining frequency and cumulative training effort relative to S2. The strategy effectively combined localized drift sensitivity with selective adaptation, allowing the framework to respond to region-specific degradation without continuously retraining the entire model.

The alternative localized baselines S5 (Random-Subspace ADWIN) and S6 (Feature-Partition ADWIN) further clarified the importance of semantically meaningful detector partitioning. Both strategies consistently achieved lower computational overhead and lower adaptation frequencies than S4. However, their predictive performance generally remained below that of S4 across both classification and regression tasks. These findings suggest that arbitrary detector decomposition alone is insufficient to fully capture heterogeneous drift dynamics. Instead, the results support the hypothesis that cluster-aware localization provides a more coherent and effective representation of evolving feature-space regions.

Collectively, the obtained results indicate that adaptation selectivity constitutes a more effective optimization objective than adaptation frequency alone. While highly reactive approaches such as S2 maximize responsiveness through continuous retraining, the localized statistically-triggered adaptation mechanism employed by S4 achieves a more balanced compromise between predictive stability and computational efficiency.

\subsection{Influence of Predictive Model Families}
The experiments additionally demonstrated that the effectiveness of the evaluated adaptation strategies depends strongly on the underlying predictive model architecture. Across both classification and regression tasks, different model families exhibited distinct levels of robustness and responsiveness under evolving distributions.

Among the classification models, ensemble-based learners such as RandomForest, XGBoost, and Adaptive Random Forest (ARF) consistently achieved the strongest overall predictive performance across most adaptation strategies. In particular, RandomForest produced some of the highest Accuracy and F1-score values under S2 and S4, while also benefiting substantially from localized adaptation. These results suggest that ensemble learners can effectively exploit localized retraining by preserving nonlinear feature interactions and maintaining region-specific decision boundaries under drift.

The online and incremental linear models, including LogisticRegression, OnlineLogisticRegression, and OnlineSoftmaxRegression, also demonstrated strong adaptation behaviour. LogisticRegression consistently remained among the top-performing classification methods across nearly all strategies, while OnlineLogisticRegression achieved particularly strong improvements under S4. These findings indicate that localized corrective updates can be efficiently integrated into incremental optimization procedures, allowing linear discriminative models to maintain competitive performance under structured distributional change.

Distance-based methods such as kNN and OnlineKNNClassifier exhibited moderate but stable improvements under adaptive strategies. Their sensitivity to localized feature-space structure likely contributed to the benefits observed under S4, where cluster-specific monitoring enabled more targeted adaptation behaviour. Similarly, OnlineNaiveBayes achieved stable but comparatively modest performance improvements, reflecting the limitations of simpler probabilistic assumptions under heterogeneous drift conditions.

The tree-based online learners HAT and AMF generally exhibited weaker predictive performance and smaller adaptation gains across several strategies. Although both methods are inherently adaptive, the results suggest that their representational flexibility may be insufficient to fully exploit localized drift information under the controlled cluster-induced shift scenarios evaluated in this study.

The regression experiments revealed similar trends. Ensemble regression models such as RandomForestRegressor, ARFRegressor, and XGBRegressor consistently achieved strong predictive robustness across the evaluated adaptation strategies. LinearRegression demonstrated stable behaviour under moderate drift conditions, while kNNRegressor benefited from localized adaptation under several mismatch configurations. HoeffdingTreeRegressor, although adaptive by design, generally produced more limited gains compared with stronger ensemble-based approaches.

These results indicate that the success of localized adaptation depends jointly on the detector mechanism and the representational capacity of the predictive learner itself. Models capable of preserving nonlinear structure and efficiently integrating localized updates benefited most substantially from the proposed framework.

\subsection{Effectiveness of Localized Drift Monitoring}

A central contribution of this work is the demonstration that localized drift monitoring provides substantial benefits over purely global adaptation mechanisms under heterogeneous non-stationary conditions. Unlike S3, which relies on a single global ADWIN detector, S4 maintains independent localized detectors across distinct regions of the feature space. This enables the framework to identify regional drift patterns that may otherwise be diluted within global error statistics.

The adaptive-window analyses presented in Figure 7 clearly illustrate this behaviour. Stable clusters exhibited steadily expanding ADWIN windows corresponding to relatively stationary local behaviour, whereas drift-prone regions experienced repeated abrupt collapses associated with localized detector resets. These observations demonstrate that the proposed framework successfully isolates drift at the regional level rather than reacting indiscriminately to global fluctuations.

Importantly, the comparison against S5 and S6 further demonstrates that the effectiveness of S4 does not arise merely from maintaining multiple detectors. While S5 and S6 also employed localized detector decomposition, their random or deterministic feature partitions lacked the semantically coherent structure provided by cluster-aware partitioning. Consequently, the results suggest that meaningful feature-space localization is necessary to fully exploit heterogeneous drift dynamics.

\subsection{Robustness under Cluster Mismatch Conditions}

An important concern for cluster-based adaptation frameworks is whether their effectiveness depends on precise alignment between detector clusters and the latent drift-generating structure. To address this issue, the experimental protocol intentionally decoupled stream-generation clustering from detector clustering through centroid perturbations and cluster-count mismatch configurations.

The obtained results demonstrate that S4 remained effective even under substantial detector-generation misalignment. Localized ADWIN detectors continued to identify region-specific instability despite imperfect cluster boundaries, and the adaptive-window analyses revealed consistent drift-reset behaviour across both classification and regression tasks under centroid mismatch and combined mismatch conditions.

These findings indicate that the proposed framework does not require perfect recovery of the latent drift-generating structure in order to maintain effective adaptation behaviour. Instead, the localized detectors appear capable of capturing approximate regional instability even when the detector partitions only partially align with the true underlying drift regions.

Furthermore, the robustness experiments demonstrate that the benefits of localized adaptation arise primarily from regional statistical sensitivity rather than exact geometric partitioning. This substantially strengthens the practical applicability of the proposed framework in real-world streaming environments where the true structure of evolving distributions is generally unknown.

\subsection{Practical Implications for Adaptive Streaming Systems}

The proposed framework has several important implications for real-world adaptive streaming applications operating under constrained computational budgets. Many practical systems, including industrial monitoring pipelines, financial forecasting platforms, healthcare decision-support systems, cybersecurity monitoring infrastructures, and cyber-physical environments, require continuous adaptation while maintaining bounded computational overhead.

Under such conditions, continuously adaptive strategies such as S2 may become operationally impractical due to their large retraining costs and high update frequencies. Conversely, fully static approaches such as S1 cannot maintain predictive robustness under evolving distributions. The obtained results indicate that detector-triggered strategies such as S3 and particularly S4 provide a more practical compromise between responsiveness and efficiency.

The moderate update frequency and reduced retraining effort achieved by S4 suggest that cluster-local adaptation constitutes a viable deployment strategy for long-running streaming systems requiring continuous online monitoring. Moreover, the ability of S4 to preserve stable behaviour in unaffected regions of the feature space may improve model reliability in applications where maintaining stable predictions is as important as adapting to new concepts.

\subsection{Limitations and Future Work}

Despite the promising results obtained in this study, several limitations remain. First, the proposed framework relies on synthetic cluster-induced drift rather than naturally occurring temporal evolution. Although this design enables precise control and reproducibility, real-world drift processes may exhibit more complex temporal dependencies and latent causal interactions.

Second, the effectiveness of localized monitoring remains partially dependent on the quality of the clustering structure used during detector initialization. While the robustness experiments demonstrated resilience to imperfect alignment, highly unstable or poorly separable feature spaces may reduce localization fidelity and detector sensitivity.

Third, the detector clusters remain static throughout the stream and therefore cannot evolve alongside the underlying distribution. Future work may investigate dynamic clustering mechanisms, online partition evolution, density-based region discovery, or representation-learning-driven localization capable of adapting the monitoring structure itself over time.

Additional future directions include adaptive detector granularity, delayed-label adaptation, multivariate drift monitoring, continual representation learning, and hybrid frameworks combining localized statistical detectors with modern online learning architectures. Such extensions may further improve robustness under highly complex non-stationary environments while preserving the efficiency advantages demonstrated by the proposed framework.

Overall, the experimental evidence demonstrates that localized statistically-triggered adaptation constitutes a substantially more efficient adaptation paradigm than globally reactive retraining under heterogeneous distributional shift conditions. By combining cluster-aware monitoring with selective retraining, the proposed S4 framework achieves a strong balance between predictive robustness, adaptation responsiveness, and computational efficiency across diverse datasets, robustness settings, and predictive model families.
\par
\endgroup

\section{Conclusion}\label{secConclusion}
This study introduced a cluster-induced distribution shift simulation framework that transforms static tabular datasets into controlled evolving streams, enabling systematic and reproducible evaluation of adaptive learning strategies under heterogeneous non-stationary conditions. Using five benchmark datasets spanning both classification and regression tasks, namely Adult, Wine Quality, Breast Cancer Wisconsin, Airfoil Self-Noise, and Superconductivty data, the proposed framework enabled comparative analysis of six adaptation strategies across diverse predictive model families, including ensemble methods, linear models, distance-based learners, and adaptive online approaches.

The experimental results demonstrate that no single adaptation strategy universally optimizes predictive performance, computational efficiency, and adaptation responsiveness simultaneously. Instead, the effectiveness of adaptive learning depends jointly on the structure of the underlying drift process, the complexity of the dataset, the predictive task, and the representational capacity of the learner itself. Nevertheless, several consistent conclusions emerge from the obtained results.

First, the experiments confirm that static training is generally insufficient under meaningful distributional evolution. Across both classification and regression tasks, the non-adaptive baseline S1 consistently exhibited predictive degradation as drift accumulated throughout the stream, demonstrating the limitations of fixed learners in evolving environments.

Second, the sliding-window retraining strategy S2 consistently achieved the strongest predictive responsiveness across most datasets and model families. Continuous retraining allowed S2 to maintain high Accuracy, F1-score, and regression performance under recurring and sustained drift conditions. However, this robustness came at a substantial operational cost. S2 produced the highest cumulative retraining effort, the largest number of update events, and the highest computational overhead across nearly all robustness settings. Consequently, the results position S2 primarily as an upper-bound responsiveness benchmark rather than a practically efficient deployment strategy for long-running adaptive systems.

Third, the detector-triggered strategies S3, S4, S5, and S6 demonstrated that selective adaptation can substantially reduce retraining cost while preserving competitive predictive performance. In particular, the proposed Cluster-Specific ADWIN strategy S4 consistently achieved the strongest balance between predictive robustness and computational efficiency across both classification and regression tasks. By maintaining independent localized drift detectors across semantically meaningful feature-space regions, S4 effectively identified heterogeneous drift patterns while avoiding unnecessary global retraining operations.

Across multiple datasets and model families, S4 maintained near-parity predictive performance relative to the strongest adaptive baselines while substantially reducing retraining effort and update frequency. Ensemble learners such as RandomForest, ARF, XGBoost, RandomForestRegressor, and XGBRegressor benefited particularly strongly from localized adaptation, while online linear methods such as OnlineLogisticRegression also demonstrated substantial gains. These findings indicate that localized statistically-triggered adaptation can preserve predictive stability while substantially limiting unnecessary adaptation operations under evolving distributions.

The comparative analysis of S5 (Random-Subspace ADWIN) and S6 (Feature-Partition ADWIN) further demonstrated that the effectiveness of localized adaptation does not arise merely from maintaining multiple detectors. Although both strategies achieved strong computational efficiency and conservative retraining behaviour, their predictive performance generally remained below that of S4 across the more challenging drift scenarios. These results suggest that semantically meaningful feature-space partitioning plays a critical role in effective localized adaptation. In particular, the findings indicate that cluster-aware regional monitoring provides a more coherent representation of heterogeneous drift dynamics than arbitrary feature decomposition alone.

An additional contribution of this work is the robustness evaluation under imperfect clustering assumptions. The mismatch experiments demonstrated that S4 remained effective even when the detector clusters were intentionally misaligned with the underlying drift-generating structure through centroid perturbations and cluster-count mismatch configurations. The localized detectors continued to identify region-specific instability despite imperfect partition alignment, indicating that the proposed framework does not require exact recovery of the latent drift structure in order to maintain effective adaptation behaviour. This robustness substantially strengthens the practical applicability of the proposed framework in real-world non-stationary environments where the true distributional structure is typically unknown.

From a broader methodological perspective, the proposed framework provides a reproducible bridge between traditional static supervised learning benchmarks and adaptive streaming evaluation. By transforming static tabular datasets into controlled evolving streams with known drift boundaries, the framework enables systematic analysis of adaptation behaviour under precisely controlled heterogeneous distributional changes. This allows direct comparison of adaptation mechanisms while preserving the structural complexity of real-world tabular datasets.

Overall, the obtained results demonstrate that localized statistically-triggered adaptation constitutes a substantially more efficient adaptation paradigm than globally reactive retraining under heterogeneous distributional shift conditions. By combining cluster-aware monitoring with selective retraining, the proposed S4 framework achieves a strong balance between predictive robustness, adaptation responsiveness, and computational efficiency across diverse datasets, robustness settings, and predictive model families.

Future work should investigate dynamic clustering mechanisms, online partition evolution, representation-learning-driven localization, adaptive detector granularity, and delayed-label adaptation settings. Additional evaluation on naturally evolving real-world streams would further clarify how localized drift decomposition interacts with different forms of abrupt, gradual, recurring, and heterogeneous controlled distribution shift under operational deployment conditions.

\sloppy
\section*{Acknowledgements}
\label{secAcknowledgements}

Funded by the European Union Horizon Europe programme CL2\allowbreak-2024\allowbreak-TRANSFORMATIONS\allowbreak-01\allowbreak-06 through ALFIE under Grant Agreement 101177912. Views and opinions expressed are however those of the author(s) only and do not necessarily reflect those of the European Union or the Agency. Neither the European Union nor the granting authority can be held responsible for them.

For the purpose of open access, the author(s) has (have) applied a Creative Commons Attribution (CC BY) licence to any Author Accepted Manuscript version arising from this submission.

\printcredits

\bibliographystyle{cas-model2-names}

\bibliography{cas-refs}

@misc{barry_becker_adult_1996,
  author       = {Becker, Barry and Kohavi, Ronny},
  title        = {Adult},
  howpublished = {UCI Machine Learning Repository},
  year         = {1996},
  doi          = {10.24432/C5XW20}
}

@article{cortez_modeling_2009,
  author  = {Cortez, Paulo and Cerdeira, Ant{\'o}nio and Almeida, Fernando and Matos, Telmo and Reis, Jos{\'e}},
  title   = {Modeling Wine Preferences by Data Mining from Physicochemical Properties},
  journal = {Decision Support Systems},
  volume  = {47},
  number  = {4},
  pages   = {547--553},
  year    = {2009},
  month   = nov,
  doi     = {10.1016/j.dss.2009.05.016}
}

@inproceedings{street_nuclear_1993,
  author    = {Street, W. Nick and Wolberg, W. H. and Mangasarian, O. L.},
  title     = {Nuclear Feature Extraction for Breast Tumor Diagnosis},
  booktitle = {Biomedical Image Processing and Biomedical Visualization},
  pages     = {861--870},
  publisher = {SPIE},
  year      = {1993},
  month     = jul,
  doi       = {10.1117/12.148698}
}

@article{gama_survey_2014,
  author  = {Gama, Jo{\~a}o and {\v{Z}}liobait{\.e}, Indr{\.e} and Bifet, Albert and Pechenizkiy, Mykola and Bouchachia, Abdelhamid},
  title   = {A Survey on Concept Drift Adaptation},
  journal = {ACM Computing Surveys},
  volume  = {46},
  number  = {4},
  pages   = {44:1--44:37},
  year    = {2014},
  month   = mar,
  doi     = {10.1145/2523813}
}

@article{lu_learning_2019,
  author  = {Lu, Jie and Liu, Anjin and Dong, Fan and Gu, Feng and Gama, Jo{\~a}o and Zhang, Guangquan},
  title   = {Learning under Concept Drift: A Review},
  journal = {IEEE Transactions on Knowledge and Data Engineering},
  volume  = {31},
  number  = {12},
  pages   = {2346--2363},
  year    = {2019},
  month   = dec,
  doi     = {10.1109/TKDE.2018.2876857}
}

@misc{aguiar2023,
  author        = {Aguiar, Gabriel J. and Cano, Alberto},
  title         = {A Comprehensive Analysis of Concept Drift Locality in Data Streams},
  year          = {2023},
  eprint        = {2311.06396},
  archiveprefix = {arXiv},
  primaryclass  = {cs.LG},
  url           = {https://arxiv.org/abs/2311.06396}
}

@article{bayram_concept_2022,
  author  = {Bayram, Firas and Ahmed, Bestoun S. and Kassler, Andreas},
  title   = {From Concept Drift to Model Degradation: An Overview on Performance-Aware Drift Detectors},
  journal = {Knowledge-Based Systems},
  volume  = {245},
  pages   = {108632},
  year    = {2022},
  month   = jun,
  doi     = {10.1016/j.knosys.2022.108632}
}

@article{agrahari_concept_2022,
  author  = {Agrahari, Supriya and Singh, Anil Kumar},
  title   = {Concept Drift Detection in Data Stream Mining: A Literature Review},
  journal = {Journal of King Saud University -- Computer and Information Sciences},
  volume  = {34},
  number  = {10},
  pages   = {9523--9540},
  year    = {2022},
  month   = nov,
  doi     = {10.1016/j.jksuci.2021.11.006}
}

@article{webb_characterizing_2016,
  author  = {Webb, Geoffrey I. and Hyde, Roy and Cao, Hong and Nguyen, Hai Long and Petitjean, Francois},
  title   = {Characterizing Concept Drift},
  journal = {Data Mining and Knowledge Discovery},
  volume  = {30},
  number  = {4},
  pages   = {964--994},
  year    = {2016},
  month   = jul,
  doi     = {10.1007/s10618-015-0448-4}
}

@article{ditzler_learning_2015,
  author  = {Ditzler, Gregory and Roveri, Manuel and Alippi, Cesare and Polikar, Robi},
  title   = {Learning in Nonstationary Environments: A Survey},
  journal = {IEEE Computational Intelligence Magazine},
  volume  = {10},
  number  = {4},
  pages   = {12--25},
  year    = {2015},
  month   = nov,
  doi     = {10.1109/MCI.2015.2471196}
}

@inproceedings{bifet_learning_2007,
  author    = {Bifet, Albert and Gavald{\`a}, Ricard},
  title     = {Learning from Time-Changing Data with Adaptive Windowing},
  booktitle = {Proceedings of the 2007 SIAM International Conference on Data Mining},
  pages     = {443--448},
  publisher = {SIAM},
  year      = {2007},
  month     = apr,
  doi       = {10.1137/1.9781611972771.42}
}

@inproceedings{gama_learning_2004,
  author    = {Gama, Jo{\~a}o and Medas, Pedro and Castillo, Gladys and Rodrigues, Pedro},
  title     = {Learning with Drift Detection},
  booktitle = {Advances in Artificial Intelligence -- SBIA 2004},
  pages     = {286--295},
  publisher = {Springer},
  year      = {2004},
  doi       = {10.1007/978-3-540-28645-5\_29}
}

@inproceedings{baena-garcia_eddm_2006,
  author    = {Baena-Garc{\'i}a, Manuel and del Campo-{\'A}vila, Jos{\'e} and Fidalgo, Ra{\'u}l and Bifet, Albert and Gavald{\`a}, Ricard and Morales-Bueno, Rafael},
  title     = {Early Drift Detection Method},
  booktitle = {Fourth International Workshop on Knowledge Discovery from Data Streams},
  pages     = {77--86},
  year      = {2006}
}

@article{ross_ewma_2012,
  author  = {Ross, Gordon J. and Adams, Niall M. and Tasoulis, Dimitris K. and Hand, David J.},
  title   = {Exponentially Weighted Moving Average Charts for Detecting Concept Drift},
  journal = {Pattern Recognition Letters},
  volume  = {33},
  number  = {2},
  pages   = {191--198},
  year    = {2012},
  month   = jan,
  doi     = {10.1016/j.patrec.2011.08.019}
}

@inproceedings{street_sea_2001,
  author    = {Street, W. Nick and Kim, YongSeog},
  title     = {A Streaming Ensemble Algorithm ({SEA}) for Large-Scale Classification},
  booktitle = {Proceedings of the Seventh ACM SIGKDD International Conference on Knowledge Discovery and Data Mining},
  pages     = {377--382},
  publisher = {ACM},
  year      = {2001},
  month     = aug,
  doi       = {10.1145/502512.502568}
}

@inproceedings{wang_ensemble_2003,
  author    = {Wang, Haixun and Fan, Wei and Yu, Philip S. and Han, Jiawei},
  title     = {Mining Concept-Drifting Data Streams Using Ensemble Classifiers},
  booktitle = {Proceedings of the Ninth ACM SIGKDD International Conference on Knowledge Discovery and Data Mining},
  pages     = {226--235},
  publisher = {ACM},
  year      = {2003},
  month     = aug,
  doi       = {10.1145/956750.956778}
}

@inproceedings{kifer_detecting_2004,
  author    = {Daniel Kifer and Shai Ben-David and Johannes Gehrke},
  title     = {Detecting Change in Data Streams},
  booktitle = {Proceedings of the Thirtieth International Conference on Very Large Data Bases (VLDB)},
  year      = {2004},
  pages     = {180--191},
  isbn      = {0-08-053979-3, 978-0-08-053979-9}
}

@article{khannouz_benchmark_2020,
  author    = {Khannouz, M. and Glatard, T.},
  title     = {A Benchmark of Data Stream Classification for Human Activity Recognition on Connected Objects},
  journal   = {Sensors},
  year      = {2020},
  volume    = {20},
  number    = {22},
  pages     = {6486},
  doi       = {10.3390/s20226486},
  publisher = {MDPI}
}

@inproceedings{bifet_adaptive_2009,
  author    = {Bifet, Albert and Gavald{\`a}, Ricard},
  title     = {Adaptive Learning from Evolving Data Streams},
  booktitle = {Advances in Intelligent Data Analysis VIII},
  pages     = {249--260},
  publisher = {Springer},
  year      = {2009},
  doi       = {10.1007/978-3-642-03915-7\_22}
}

@article{gomes_arf_2017,
  author  = {Gomes, Heitor M. and Bifet, Albert and Read, Jesse and Barddal, Jean Paul and Enembreck, Fabr{\'i}cio and Pfahringer, Bernhard and Holmes, Geoff and Abdessalem, Talel},
  title   = {Adaptive Random Forests for Evolving Data Stream Classification},
  journal = {Machine Learning},
  volume  = {106},
  number  = {9},
  pages   = {1469--1495},
  year    = {2017},
  month   = oct,
  doi     = {10.1007/s10994-017-5642-8}
}

@article{schlimmer_incremental_1986,
  author  = {Schlimmer, Jeffrey C. and Granger, Richard H.},
  title   = {Incremental Learning from Noisy Data},
  journal = {Machine Learning},
  volume  = {1},
  number  = {3},
  pages   = {317--354},
  year    = {1986},
  month   = sep,
  doi     = {10.1007/BF00116895}
}

@incollection{aggarwal_clustream_2003,
  author    = {Aggarwal, Charu C. and Han, Jiawei and Wang, Jianyong and Yu, Philip S.},
  title     = {A Framework for Clustering Evolving Data Streams},
  booktitle = {Data Streams: Models and Algorithms},
  pages     = {81--92},
  publisher = {Springer},
  year      = {2003},
  doi       = {10.1016/B978-012722442-8/50016-1}
}

@inproceedings{spinosa_olindda_2007,
  author    = {Spinosa, Eduardo J. and de Leon F. de Carvalho, Andr{\'e} Ponce and Gama, Jo{\~a}o},
  title     = {{OLINDDA}: A Cluster-Based Approach for Detecting Novelty and Concept Drift in Data Streams},
  booktitle = {Proceedings of the 2007 ACM Symposium on Applied Computing},
  pages     = {448--452},
  publisher = {ACM},
  year      = {2007},
  month     = mar,
  doi       = {10.1145/1244002.1244107}
}

@inproceedings{bifet_moa_2010,
  author    = {Bifet, Albert and Holmes, Geoff and Pfahringer, Bernhard and Kranen, Philipp and Kremer, Hardy and Jansen, Timm and Seidl, Thomas},
  title     = {{MOA}: Massive Online Analysis, a Framework for Stream Classification and Clustering},
  booktitle = {Proceedings of the First Workshop on Applications of Pattern Analysis},
  pages     = {44--50},
  publisher = {PMLR},
  year      = {2010},
  month     = sep
}

@inproceedings{pesaranghader_mcdiarmid_ijcnn_2018,
  author    = {Pesaranghader, Ali and Viktor, Herna L. and Paquet, Eric},
  title     = {{McDiarmid} Drift Detection Methods for Evolving Data Streams},
  booktitle = {2018 International Joint Conference on Neural Networks ({IJCNN})},
  pages     = {1--9},
  year      = {2018},
  month     = jul,
  doi       = {10.1109/IJCNN.2018.8489260}
}

@article{liu_concept_2021,
  author  = {Liu, Anjin and Lu, Jie and Zhang, Guangquan},
  title   = {Concept Drift Detection via Equal Intensity K-Means Space Partitioning},
  journal = {IEEE Transactions on Cybernetics},
  volume  = {51},
  number  = {6},
  pages   = {3198--3211},
  year    = {2021},
  month   = jun,
  doi     = {10.1109/TCYB.2020.2983962}
}

@inproceedings{domingos_mining_2000,
  author    = {Domingos, Pedro and Hulten, Geoff},
  title     = {Mining High-Speed Data Streams},
  booktitle = {Proceedings of the Sixth ACM SIGKDD International Conference on Knowledge Discovery and Data Mining},
  pages     = {71--80},
  publisher = {ACM},
  year      = {2000},
  month     = aug,
  doi       = {10.1145/347090.347107}
}

@inproceedings{oza_online_2005,
  author    = {Oza, N. C.},
  title     = {Online Bagging and Boosting},
  booktitle = {Proceedings of the IEEE International Conference on Systems, Man and Cybernetics},
  pages     = {2340--2345},
  year      = {2005},
  month     = oct,
  doi       = {10.1109/ICSMC.2005.1571498}
}

@inproceedings{read_batch-incremental_2012,
  author    = {Read, Jesse and Bifet, Albert and Pfahringer, Bernhard and Holmes, Geoff},
  title     = {Batch-Incremental versus Instance-Incremental Learning in Dynamic and Evolving Data},
  booktitle = {Advances in Intelligent Data Analysis XI},
  pages     = {313--323},
  publisher = {Springer},
  year      = {2012},
  doi       = {10.1007/978-3-642-34156-4\_29}
}

@article{morales_samoa_2015,
  author  = {Morales, Gianmarco De Francisci and Bifet, Albert},
  title   = {{SAMOA}: Scalable Advanced Massive Online Analysis},
  journal = {Journal of Machine Learning Research},
  volume  = {16},
  number  = {5},
  pages   = {149--153},
  year    = {2015}
}

@article{losing_incremental_2018,
  author  = {Losing, Viktor and Hammer, Barbara and Wersing, Heiko},
  title   = {Incremental On-Line Learning: A Review and Comparison of the State of the Art Algorithms},
  journal = {Neurocomputing},
  volume  = {275},
  pages   = {1261--1274},
  year    = {2018},
  month   = jan,
  doi     = {10.1016/j.neucom.2017.06.084}
}

@article{souza_challenges_2020,
  author  = {Souza, Vinicius M. A. and dos Reis, Denis M. and Maletzke, Andr{\'e} G. and Batista, Gustavo E. A. P. A.},
  title   = {Challenges in Benchmarking Stream Learning Algorithms with Real-World Data},
  journal = {Data Mining and Knowledge Discovery},
  volume  = {34},
  number  = {6},
  pages   = {1805--1858},
  year    = {2020},
  month   = nov,
  doi     = {10.1007/s10618-020-00698-5}
}

@misc{zliobaite_learning_2010,
  author       = {{\v{Z}}liobait{\.e}, Indr{\.e}},
  title        = {Learning under Concept Drift: An Overview},
  howpublished = {arXiv preprint arXiv:1010.4784},
  year         = {2010},
  month        = oct,
  doi          = {10.48550/arXiv.1010.4784}
}

@article{haque_sand_2016,
  author  = {Haque, Ahsanul and Khan, Latifur and Baron, Michael},
  title   = {{SAND}: Semi-Supervised Adaptive Novel Class Detection and Classification over Data Stream},
  journal = {Proceedings of the AAAI Conference on Artificial Intelligence},
  volume  = {30},
  number  = {1},
  year    = {2016},
  doi     = {10.1609/aaai.v30i1.10283}
}

@article{breiman2001random,
  author  = {Breiman, Leo},
  title   = {Random Forests},
  journal = {Machine Learning},
  volume  = {45},
  number  = {1},
  pages   = {5--32},
  year    = {2001},
  doi     = {10.1023/A:1010933404324}
}

@book{hosmer2000applied,
  author    = {Hosmer, David W. and Lemeshow, Stanley},
  title     = {Applied Logistic Regression},
  publisher = {Wiley},
  year      = {2000},
  isbn      = {9780471356325},
  doi       = {10.1002/0471722146}
}

@article{cover1967nearest,
  author  = {Cover, T. M. and Hart, P. E.},
  title   = {Nearest Neighbor Pattern Classification},
  journal = {IEEE Transactions on Information Theory},
  volume  = {13},
  number  = {1},
  pages   = {21--27},
  year    = {1967},
  doi     = {10.1109/TIT.1967.1053964}
}

@inproceedings{chen2016xgboost,
  author    = {Chen, T. and Guestrin, C.},
  title     = {XGBoost: A Scalable Tree Boosting System},
  booktitle = {Proceedings of the 22nd ACM SIGKDD International Conference on Knowledge Discovery and Data Mining},
  pages     = {785--794},
  year      = {2016},
  doi       = {10.1145/2939672.2939785}
}

@book{hastie2009elements,
  author    = {Hastie, T. and Tibshirani, R. and Friedman, J.},
  title     = {The Elements of Statistical Learning: Data Mining, Inference, and Prediction},
  edition   = {2},
  publisher = {Springer},
  year      = {2009},
  doi       = {10.1007/978-0-387-21606-5}
}

@book{bishop2006pattern,
  author    = {Bishop, C. M.},
  title     = {Pattern Recognition and Machine Learning},
  publisher = {Springer},
  year      = {2006},
  address   = {New York},
  series    = {Information Science and Statistics},
  isbn      = {978-0-387-31073-2}
}

@book{geron2019hands,
  author    = {G{\'e}ron, Aur{\'e}lien},
  title     = {Hands-On Machine Learning with Scikit-Learn, Keras, and TensorFlow},
  publisher = {O'Reilly Media},
  year      = {2019},
  edition   = {2},
  isbn      = {978-1492032649}
}

@book{james2013introduction,
  author    = {James, G. and Witten, D. and Hastie, T. and Tibshirani, R.},
  title     = {An Introduction to Statistical Learning: with Applications in R},
  publisher = {Springer},
  year      = {2013},
  address   = {New York},
  doi       = {10.1007/978-1-4614-7138-7}
}

@book{little2002statistical,
  author    = {Little, R. J. A. and Rubin, D. B.},
  title     = {Statistical Analysis with Missing Data},
  publisher = {Wiley},
  year      = {2002},
  edition   = {2},
  doi       = {10.1002/9781119013563},
  isbn      = {9780471183860}
}

@article{Hancock2020CategoricalNN,
  author    = {Hancock, John T. and Khoshgoftaar, Taghi M.},
  title     = {Survey on categorical data for neural networks},
  journal   = {Journal of Big Data},
  volume    = {7},
  number    = {1},
  pages     = {28},
  year      = {2020},
  doi       = {10.1186/s40537-020-00305-w},
  url       = {https://doi.org/10.1186/s40537-020-00305-w}
}

@inproceedings{Vinagre2014,
  author    = {Vinagre, Jo{\~a}o and Jorge, Al{\'i}pio M. and Gama, Jo{\~a}o},
  title     = {Evaluation of recommender systems in streaming environments},
  booktitle = {Proceedings of the 9th Workshop on Real-Time Business Intelligence and Analytics},
  series    = {BIRTE '15},
  year      = {2014},
  pages     = {1--8},
  doi       = {10.1145/2611286.2611299},
}

@article{suarez2023recurring,
  title   = {A survey on machine learning for recurring concept drifting data streams},
  author  = {Su{\'a}rez-Cetrulo, Andr{\'e}s L. and Quintana, Diego and Cervantes, Alejandro},
  journal = {Expert Systems with Applications},
  volume  = {213},
  pages   = {118934},
  year    = {2023},
  publisher = {Elsevier},
  doi     = {10.1016/j.eswa.2022.118934},
  url     = {https://www.sciencedirect.com/science/article/pii/S0957417422019522}
}

@book{gama2010knowledge,
  title     = {Knowledge Discovery from Data Streams},
  author    = {Gama, Jo{\~a}o},
  edition   = {1},
  year      = {2010},
  publisher = {Chapman and Hall/CRC},
  address   = {Boca Raton, FL},
  doi       = {10.1201/EBK1439826119},
  url       = {https://doi.org/10.1201/EBK1439826119}
}

@article{Mehmoodetal2021,
  author  = {Mehmood, Hassan and Kostakos, Panos and Cortes, Marta and Anagnostopoulos, Theodoros and Pirttikangas, Susanna and Gilman, Ekaterina},
  title   = {Concept Drift Adaptation Techniques in Distributed Environment for Real-World Data Streams},
  journal = {Smart Cities},
  year    = {2021},
  volume  = {4},
  number  = {1},
  pages   = {349--371},
  doi     = {10.3390/smartcities4010021}
}

@inproceedings{CabelloLopezetal2022,
  author    = {Cabello-López, Tomás and Cañizares-Juan, Manuel and Carranza-García, Manuel and Garcia-Gutiérrez, Jorge and Riquelme, José C.},
  title     = {Concept Drift Detection to Improve Time Series Forecasting of Wind Energy Generation},
  booktitle = {Lecture Notes in Computer Science},
  year      = {2022},
  pages     = {133--140},
  doi       = {10.1007/978-3-031-15471-3\_12}
}

@inproceedings{KebirTabia2024,
  author    = {Kebir, Sara and Tabia, Karim},
  title     = {On Handling Concept Drift, Calibration and Explainability in Non-Stationary Environments and Resources Limited Contexts},
  booktitle = {Proceedings of the 16th International Conference on Agents and Artificial Intelligence},
  year      = {2024},
  pages     = {336--346},
  doi       = {10.5220/0012382200003636}
}

@article{Liuetal2017,
  author  = {Liu, Anjin and Song, Yiliao and Zhang, Guangquan and Lu, Jie},
  title   = {Regional Concept Drift Detection and Density Synchronized Drift Adaptation},
  journal = {Proceedings of the Twenty-Sixth International Joint Conference on Artificial Intelligence},
  year    = {2017},
  pages   = {2280--2286},
  doi     = {10.24963/ijcai.2017/317}
}

@article{Hinderetal2024,
  author  = {Hinder, Fabian and Vaquet, Valerie and Hammer, Barbara},
  title   = {One or two things we know about concept drift—a survey on monitoring in evolving environments. Part B: locating and explaining concept drift},
  journal = {Frontiers in Artificial Intelligence},
  year    = {2024},
  volume  = {7},
  doi     = {10.3389/frai.2024.1330258}
}

@article{sethi2017reliable,
  title   = {On the reliable detection of concept drift from streaming unlabeled data},
  author  = {Sethi, Tegjyot Singh and Kantardzic, Mehmed},
  journal = {Expert Systems with Applications},
  volume  = {82},
  pages   = {77--99},
  year    = {2017},
  doi     = {10.1016/j.eswa.2017.04.008},
  publisher = {Elsevier}
}

@article{zliobaite2014active,
  title   = {Active Learning with Drifting Streaming Data},
  author  = {{\v{Z}}liobait{\.e}, Indr{\.e} and Bifet, Albert and Pfahringer, Bernhard and Holmes, Geoffrey},
  journal = {IEEE Transactions on Neural Networks and Learning Systems},
  volume  = {25},
  number  = {1},
  pages   = {27--39},
  year    = {2014},
  doi     = {10.1109/TNNLS.2012.2236570}
}

@misc{kam_hamidieh_data-driven_2018,
	title = {A data-driven statistical model for predicting the critical temperature of a superconductor - {ScienceDirect}},
	url = {https://www.sciencedirect.com/science/article/pii/S0927025618304877?via%3Dihub},
	urldate = {2026-04-24},
	author = {{Kam Hamidieh}},
	month = mar,
	year = {2018},
}

@misc{thomas_brooks_airfoil_1989,
	title = {Airfoil {Self}-{Noise}},
	url = {https://archive.ics.uci.edu/dataset/291},
	doi = {10.24432/C5VW2C},
	urldate = {2026-04-24},
	publisher = {UCI Machine Learning Repository},
	author = {Thomas Brooks, D. Pope},
	year = {1989},
}

@article{ho1998random,
  title={The Random Subspace Method for Constructing Decision Forests},
  author={Ho, Tin Kam},
  journal={IEEE Transactions on Pattern Analysis and Machine Intelligence},
  volume={20},
  number={8},
  pages={832--844},
  year={1998}
}

@article{kuncheva2003measures,
  title={Measures of Diversity in Classifier Ensembles and Their Relationship with the Ensemble Accuracy},
  author={Kuncheva, Ludmila I. and Whitaker, Christopher J.},
  journal={Machine Learning},
  volume={51},
  number={2},
  pages={181--207},
  year={2003}
}



\end{document}